\documentclass[letterpaper, 10 pt, conference]{ieeeconf}
\IEEEoverridecommandlockouts 
\usepackage{amsmath}
\usepackage{amssymb}
\usepackage{subfigure}
\usepackage{soul}
\usepackage{booktabs}
\usepackage{graphicx}
\usepackage[dvipsnames]{xcolor}

\title{\LARGE \bf
ALPHA: Attention-based Long-horizon Pathfinding in Highly-structured Areas
}

\author{He Chengyang$^{1}$, Yang Tianze$^{1}$, Tanishq Duhan$^{1}$, Wang Yutong$^{1}$, Guillaume Sartoretti$^{1}$
\thanks{$^{1}$Authors are with department of Mechanical Engineering, College of Design and Engineering, National University of Singapore, 21 Lower Kent Ridge Rd, Singapore
        {\tt\small \{chengyanghe, yangtianze, e1280621, e0576114\}@u.nus.edu, guillaume.sartoretti@nus.edu.sg}}
}

\begin{document}

\maketitle
\thispagestyle{empty}
\pagestyle{empty}

\begin{abstract}

The multi-agent pathfinding (MAPF) problem seeks collision-free paths for a team of agents from their current positions to their pre-set goals in a known environment, and is an essential problem found at the core of many logistics, transportation, and general robotics applications. 
Existing learning-based MAPF approaches typically only let each agent make decisions based on a limited field-of-view (FOV) around its position, as a natural means to fix the input dimensions of its policy network. 
However, this often makes policies short-sighted, since agents lack the ability to perceive and plan for obstacles/agents beyond their FOV. 
To address this challenge, we propose ALPHA, a new framework combining the use of ground truth proximal (local) information and fuzzy distal (global) information to let agents sequence local decisions based on the full current state of the system, and avoid such myopicity. 
We further allow agents to make short-term predictions about each others' paths, as a means to reason about each others' path intentions, thereby enhancing the level of cooperation among agents at the whole system level.
Our neural structure relies on a Graph Transformer architecture to allow agents to selectively combine these different sources of information and reason about their inter-dependencies at different spatial scales.
Our simulation experiments demonstrate that ALPHA outperforms both globally-guided MAPF solvers and communication-learning based ones, showcasing its potential towards scalability in realistic deployments.

\end{abstract}

\section{INTRODUCTION}

As artificial intelligence (AI) improves by leaps and bounds, robots/agents, now more than ever, can be deployed in man-made structures such as warehouses, seaports, and airports~\cite{honig2019persistent,li2021lifelong,rekik2019multi,polydorou2021learning}, to assist with the transportation of goods and personnel. 
In such cases involving multiple agents within a known, static environment, an essential sub-task is to plan collision-free paths for all agents from their current positions to their pre-set goals~\cite{stern2019multi}.
This problem is known as Multi-Agent Pathfinding (MAPF). 

The evolution of neural networks has impacted MAPF by introducing learning-based approaches, leading to a growing trend within the community to develop such methods.
However, unlike traditional planners that have access to global states and give complete solutions~\cite{ferner2013odrm,sharon2015conflict,ma2019searching,li2020new,li2022mapf}, most existing learning-based planners rely on limited field-of-view (FoV) to make local plans.
Although this reliance can lead to myopicity, particularly in highly structured environments where distant obstacles/agents should be taken into account for the planning, existing learning-based solutions still heavily depend on limited FoV due to the constraints posed by the input dimensions of Convolutional Neural Networks (CNNs)~\cite{li2021survey}.
In order to provide agents with more information beyond the confines of their FoV, current methods either use part of expert paths within the FoV or let agents communicate with each other~\cite{li2020graph,li2021message}.
However, these methods implicitly exploit partial slices of the global map rather than encoding the whole map.
Consequently, the challenge at hand is how to encode and distill the vast global information into abstract forms that can explicitly assist agents in long-horizon planning.
To address this problem, we introduce ALPHA, a novel MAPF planner which uses our proposed augmented graph representation to capture global map states and employ an attention-based network to achieve context learning by reasoning about nodes' inter-dependencies.

\begin{figure}
\centering
\includegraphics[width=2.5in]{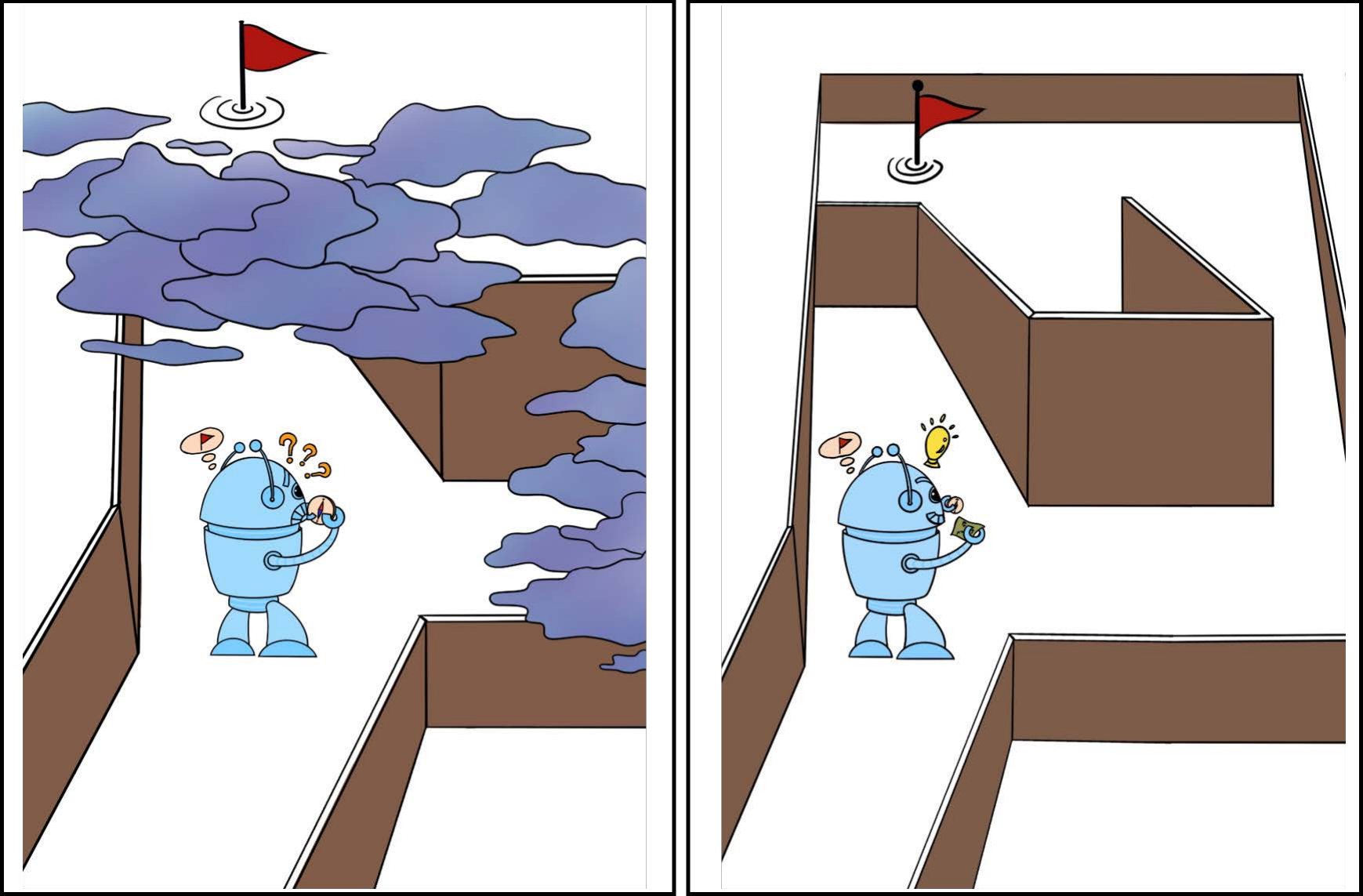}
\vspace{-0.3cm}
\caption{Most of the existing learning-based MAPF methods rely on limited FOV, which makes the agent lack the ability to perceive and consider the environment beyond the FOV and thus leads to short-sighted policies (left example), while ALPHA enables the agent to perceive global information and consequently to make the right decisions (right figure).}
\label{compass_and_map}
\vspace{-0.4cm}
\end{figure}

Our key insight revolves around creating a low-dimensional representation of the global map to grasp the map structure and other agents' intentions, which prevents the raw data from overwhelming and confusing the network due to its dense features.
We split the global information into more learnable static and dynamic channels, improving the ability of agents to understand the structure of the map and infer the intentions of other agents.
To tackle static obstacles, we design a graph representation of the global map that incorporates handcrafted features for extracting structure formed by interconnected static obstacles in highly structured maps. 
For the dynamic side, we assume that the dynamic uncertainty in the environment solely arises from the movement of fellow agents, without the presence of unmodeled dynamic obstacles.
Therefore, for dealing with other agents, we develop another graph that encapsulates the short-term intention of each agent.
To process these graphs, ALPHA utilizes attention mechanisms to help the network prioritize critical nodes, connections, and contextual information, which enables agents to better reason about more important regions to aid decision-making.
Experimental results show that with the help of these additional global graphs, ALPHA outperforms state-of-the-art non-communication learning-based MAPF solvers in both random, warehouse, and highly-structured maps (e.g. room-like environments with space-limited doors and narrow corridors).
Even in congested traffic scenarios, ALPHA demonstrates comparable performance to the latest communication-based MAPF solvers.

\section{RELATED WORK}

In recent years, there has been growing interest in using machine learning techniques to solve MAPF. 
These learning-based methods can be broadly categorized into three types: relying solely on local information, combining with global guidance, and incorporating agents' communication.

Methods relying solely on local observations, such as our previous work PRIMAL~\cite{sartoretti2019primal}, utilize reinforcement learning (RL) and imitation learning (IL) to provide fully decentralized solutions.
PRIMAL2~\cite{damani2021primal} extends this characteristic to the life-long MAPF problem, allowing agents to learn conventions that structure and coordinate their paths.
The local observations proposed by PRIMAL have demonstrated outstanding performance, inspiring many subsequent approaches to adopt this representation method.
Building upon local observations, some researchers have proposed to incorporate global guidance to enhance agent performance, as seen in MAPPER~\cite{liu2020mapper} and G2RL~\cite{wang2020mobile}. 
However, these methods often drive agents to follow the path generated by expert single-agent algorithms to various extents, which may reduce coordination and lead to more rigid decision-making.
The third class of approaches uses communication learning to exchange information with each other and achieve better individual decision-making and group coordination~\cite{ma2021distributed,wang2023scrimp}.
Although communication learning enhances the accuracy of agent dynamics, the static map structure is still indirectly accessed through other agents' FoV.
Thereby, due to the challenge of directly integrating whole map information with different spatial scales into neural network frameworks, these methods still use field-of-view-based representations to capture slices of global information locally.
While this means might entail information loss beyond the selected slice, it is still a general method in state-of-the-art learn-based MAPF solvers because it is a natural means to fix the neural network input dimension.

Unlike previous methods, ALPHA proposes to encode global information and combines it with local observations directly without any slicing or reduction of global information. 
Moreover, ALPHA does not explicitly require agents to follow expert paths, which potentially enhances the flexibility and coordination of agent decisions. 
This paradigm provides agents with more comprehensive and long-term information, enabling agents to make better planning decisions.

\section{PROBLEM STATEMENT}

\subsection{MAPF Problem Formulation}

    In the MAPF problem, we have $n$ agents $A=\{a_1, ..a_n\}$ and an undirected graph $G=(V, E)$ where $E$ is the set of edges connecting the set of vertices $V$.
    Each agent $a_i$ is assigned a unique start vertex ($s_i \in V$) and a unique destination/goal vertex ($d_i \in V$).
    We call the set of all start vertices as $S$ and the set of all destination vertices as $D$.
    We assume time to be discretized into uniform steps. At every time step, each agent executes one of the two options: i) move to one of its adjacent vertices in the graph, or ii) wait at its current vertex. 
    The set of actions that all agents perform at a time step $t_j$ is referred to as a joint set $J_{t_j}$. 
    A joint set of actions is considered valid if no two agents occupy the same vertex at any time step: $a_{i,{t_j}} \neq a_{k,{t_j}}$, and if no two agents swap vertices in a single time step ($a_{i,{t_{j+1}}} = a_{k,{t_j}} \iff a_{k,{t_{j+1}}} \neq a_{i,{t_j}}$), where $a_{i,{t_j}}$ denotes the position of agent $a_i$ at time $t_j$
    Our objective is to determine a series of valid joint sets that guide agents from their source vertices $S$ to their goal vertices $D$ in the least amount of time steps.
    
\subsection{Environment Setup}

    Remaining consistent with the standard MAPF problem, we use the following setup: the map is a 2D discrete 4-connectivity grid world, and agents can move to the free cell adjacent to their location or stay idle at each time step. 
    An episode terminates when all agents are on their goals at the end of a time step (success) or when the number of time steps reaches the pre-defined limit (failure).
    Unlike previous rigid and simple room environments~\cite{stern2019multi}, we also look at a new room-like map generator that offers greater flexibility in terms of the number, size, and shape of rooms in the generated maps.
    In particular, the new room map contains corridors of varying widths, which are highly similar to real offices, warehouses, and other environments.
    Unlike the timeliness impact of loose obstacles in random maps, continuous obstacles in such highly structured maps may have a significant impact on current decision-making even across substantial distances.

\section{MAPF \footnotesize{AS} \normalsize{AN RL PROBLEM}}

In this section, we elaborate on our approach to processing static and dynamic information in the global map to enable agents to learn informed policies from it.

\subsection{Observations}

    In order for the agent to develop long-horizon planning capability beyond its FoV, we believe that the agent's observations should be composed of ground truth local observations and fuzzy global observations.
    For local observations, we follow our previous work~\cite{sartoretti2019primal} to provide grid-based local information in four separate channels, giving agents the potential ability of local obstacle avoidance and coordination.

\subsubsection{Global Static Observations}
    
    The process of obtaining global static observations can be divided into two phases: firstly, the extraction of a graph from the global map, and secondly, augmenting the graph with high-level features to enhance its expressiveness.

    \paragraph{Graph extraction}

    To prevent agents from being overwhelmed by dense grid-based global information, we extract a concise graph representation from the map.
    We process the maps in three steps: skeletonization, neighborhood analysis, and edge generation.
    The skeletonization of the map is inspired by binary image thinning in machine vision for feature extraction and topological representation.
    We tried the Zhang-Suen algorithm and the Medial Axis Transform method.
    The skeleton generated by the former has fewer branches.
    More branches mean more information, but also more computation.
    We then perform neighborhood analysis on the resulting skeleton to identify \textit{nodes} capable of representing the map structure.
    Concretely, we select all branch pixels (connected to at least three other pixels) and leaf pixels (connected to at most one other pixel) of the thinning map to construct our set of nodes.
    Finally, we obtain the edges based on the skeleton and the nodes, which use CV methods or the A* algorithm.
    We employed the 8-connectivity A* algorithm to identify edges between two nodes in the skeleton by checking the presence of other nodes along their connecting paths.

    After graph extraction, we obtain a two-dimensional graph representation of the map with $N$ nodes at time step $t$, represented by a 2D coordinate set:
    \begin{equation}
    \begin{aligned}
    &\mathcal{V}^2_t=\{v^2_{1,t},v^2_{2,t},\cdots, v^2_{N-2,t}, v^2_{N-1,t},v^2_{N,t} \} \\
    &\forall v^2_{i,t}=(x_{i,t}, y_{i,t}),
    \end{aligned}
    \end{equation}
    which can only characterize some important free cells in the map.
    In addition to these map-generated nodes, we also add two extra nodes into the graph to represent the agent's current position $v^2_{N+1,t}$ and goal position $v^2_{N+2,t}$.
    
    \paragraph{Augmented Static Graph}

    Recognizing the limitations of 2D coordinates in capturing obstacle structures, we augment each node with high-level features to enhance the graph's ability to represent global structure.
    Our intuition is that a node's value to a single agent pathfinding can be reflected in three aspects:
    i) The amount of detour (if any) an agent requires to reach a specific node (owing to obstacles) in comparison to the Manhattan distance; ii) similarly, the amount of detour required to reach the goal from the specified node; iii) the amount of deviation from the optimal path when detoured through this specific node.  
    For this purpose, on top of the coordinates of the nodes, we set three additional features: node accessibility, detour-to-goal, and off-route degree.
    \begin{itemize}
        \item \textit{Node accessibility}, denoting the difficulty for the agent to get from its current position to the node, is defined as follows for node $i$:
        \begin{equation}
        \begin{aligned}
        d^{na}_{i,t} = \mathcal{A}^*_{len}(v^2_{N+1,t}, v^2_{i,t}) - \mathcal{M}_{dis}(v^2_{N+1,t}, v^2_{i,t})
        \end{aligned}
        \end{equation}
        \item \textit{Detour-to-goal} is used to evaluate the difficulty of reaching the agent's goal from the node $i$:
        \begin{equation}
        \begin{aligned}
        d^{dg}_{i,t} = \mathcal{A}^*_{len}(v^2_{N+2,t}, v^2_{i,t}) - \mathcal{M}_{dis}(v^2_{N+2,t}, v^2_{i,t})
        \end{aligned}
        \end{equation}
        \item \textit{Off-route degree} quantifies the extent of the node's deviation from the agent's potentially optimal path to its goal.
        \begin{equation}
        \begin{aligned}
            d^{od}_{i,t} =~ &\mathcal{A}^*_{len}(v^2_{N+1,t}, v^2_{i,t})+\mathcal{A}^*_{len}(v^2_{N+2,t}, v^2_{i,t}) \\
                      &-\mathcal{A}^*_{len}(v^2_{N+1,t}, v^2_{N+2,t})
        \end{aligned}
        \end{equation}
    \end{itemize}
    where $\mathcal{A}^*_{len}(\cdot, \cdot)$ is the length of a path generated by the A* algorithm between two coordinates, and $\mathcal{M}_{dis}(\cdot, \cdot)$ the Manhattan distance between two coordinates. 
    Ultimately, we obtain an augmented static graph with high-level features $\mathcal{V}^5_t$:
    \begin{align}
        &\mathcal{V}^5_t = \{v^5_{1,t}, v^5_{2,t}, \cdots, v^5_{N,t}\} \quad with \notag\\
        &v^5_{i,t} = (x_{i,t}, y_{i,t}, d^{na}_{i,t}, d^{dg}_{i,t}, d^{od}_{i,t}),~~i=1,\cdots,N
    \end{align}
    where $(x_t,y_t)$ is the relative coordinates of the node in the map at time $t$, and the number of vectors in the set depends on the number of nodes in the augmented graph. 
    We believe that, by focusing on these 5-dimensional nodes, the agent can learn to identify valuable areas in the map and develop long-horizon planning capability in single-agent scenarios.
    In doing so, the additional advantage of such representation is that the agent is not required to consider any specific nodes as mandatory waypoints, thus enhancing planning flexibility.
    A more flexible policy could enhance the prospects of agent cooperation, as it might prioritize team coordination over its own optimal policy.

\subsubsection{Global Dynamic Observations}

    After obtaining the augmented graph used to characterize the map structure, the next step is to acquire the intentions of other agents to facilitate higher-level coordination.
    As mentioned earlier, given that the dynamic uncertainty in the environment derives from the movements of other agents, we interpret the intentions of other agents as their (short-term) individual-A* paths to goal.
    To this end, we construct a second global graph whose nodes consist of agents in the map, whose features are used to represent the corresponding agent's intention.
    Specifically, the features of each node consist of three components: 1) the current positions of that agent, 2) its predicted future positions, and 3) its direction of travel.

    \renewcommand{\dblfloatpagefraction}{.58}
    \begin{figure*}
    \centering
    \includegraphics[width=6in]{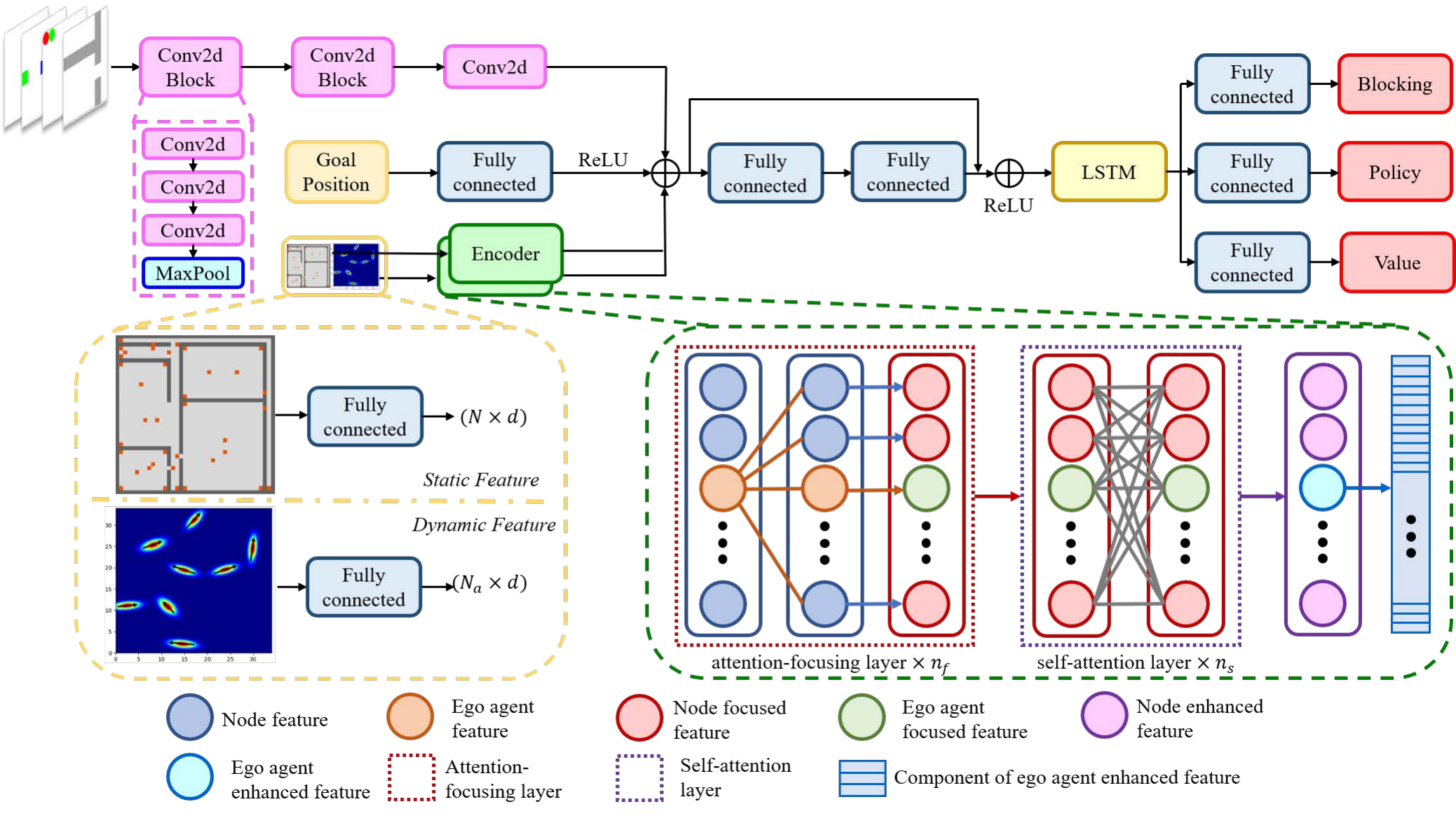}
    \caption{ALPHA's attention-based neural network. 
    Note that for global observation, the static features and dynamic features are passed into different encoders through different channels, and then the output features are concatenated together to form the final features of the global observation.
    Feature embedding: During the embedding process of nodes in the map, where the static features of any node $\forall v_{i,t}^5\in\mathcal{V}^5_t$ need to be transformed into $d$-dimensional embedding vectors through a linear layer;
    similarly, the 9-dimensional dynamic features of all agents $\forall a^i_{f,t}\in\mathcal{D}_{f,t}$ also need to undergo a linear layer transformation to obtain $d$-dimensional representations.
    $N$ and $N_a$ represent the number of nodes and the number of agents in the map, respectively.
    }
    \label{net}
    \end{figure*}
    
    The current position of agent $i$ is denoted by $(x_{curr}^i, y_{curr}^i)$.
    For the predicted future positions of other agents, we fit a 2D Gaussian distribution with the means and variances of the short-term A* predicted paths, indicating the areas where other agents are likely to appear.
    Specifically, at time $t$, we compute the trajectories of each agent for the next $f$ steps using A*, denoted as $tr_{t,f}^i(p_1^i, p_2^i, \cdots, p_f^i)$.
    Then, we calculate the mean and variance of all points in the trajectory $tr_{t,f}^i$ along the $x$-axis and $y$-axis, respectively.
    These means $(\mu_{f,x,t}^i, \mu_{f,y,t}^i)$ and variances $(\sigma_{f,x,t}^i,\sigma_{f,y,t}^i)$ are used to represent the likelihood of potential positions that the agent may occupy within the next $f$ steps.
    Finally, we assign each agent a vector $(dx_{f,t}^i, dy_{f,t}^i, mag_{f,t}^i)$ pointing towards its predicted position after $f$ steps, where $(dx_{f,t}^i, dy_{f,t}^i)$ is its unit direction vector and $mag_{f,t}^i$ is its the magnitude.
    The global dynamic graph used to describe the agents' intention consists of a 9-dimensional vector set $\mathcal{D}_{f,t}$:
    \begin{align}
        \mathcal{D}_{f,t} = &\{a^1_{f,t}, a^2_{f,t}, \cdots, a^{N_a}_{f,t}\}\notag\\
        a^i_{f,t} =& (x_{curr}^i, y_{curr}^i, \mu_{f,x,t}^i, \sigma_{f,x,t}^i, \mu_{f,y,t}^i, \sigma_{f,y,t}^i, \notag\\ 
                          &~~dx_{f,t}^i, dy_{f,t}^i, mag_{f,t}^i)~~ i = 1,2,\cdots,N_a
    \end{align}
    where $N_a$ is the number of agents.
    It is crucial to emphasize that due to the inherent uncertainty of learning-based agents, anticipating their strict adherence to A* paths and formulating policies based on A*-informed long-term predictions can have adverse effects on the performance of the MAPF planner.
    But in the short term, A* paths can provide valuable insights into the agent's immediate actions, particularly in highly structured scenarios involving narrow doors and corridors.

\subsection{Action Space and Reward}
    \begin{table}[!t]
    \renewcommand{\arraystretch}{1.3}
    \caption{Reward structure}
    \vspace{-0.1cm}
    \label{reward}
    \centering
    \scalebox{0.7}{
    \begin{tabular}{c||c}
    \hline
    Action & Reward\\
    \hline
    Move to cardinal directions & -0.3\\
    \hline
    Stay idle (not on goal) & -0.3\\
    \hline
    Stay idle (on goal) & 0.0\\
    \hline
    Collision with others/obstacle & -2.0\\
    \hline
    Blocking other agents & -1.0 $\times$ $\eta$\\
    \hline
    \end{tabular}}
    \end{table}
    The agent's action space contains five elements, namely four cardinal directions (up, down, left, right) and stay idle.
    Our reward structure is shown in Table \ref{reward}. 
    If the ego agent blocks other agents, it incurs a penalty of $blocking~penalty \times \eta$, where $\eta$ represents the number of blocked agents~\cite{sartoretti2019primal}.

\section{Attention-Based Neural Network}

For local observations derived from precise grid states, we employ convolutional layers and pooling layers (inspired by VGGnet~\cite{simonyan2014very}) to condense the information; meanwhile, for global observations structured as graphs, we utilize encoders based on the graph transformer~\cite{yun2019graph} to integrate the information.
The most interesting part of this network architecture is its novel encoder, utilized to infer the inter-dependencies among nodes in both the static and dynamic augmented graphs generated from global observations.
These inter-dependencies are commonly referred to as \textit{context}~\cite{cao2023catnipp}.
Through context learning, the agent infers which of the global context-aware nodes is more important to its decision-making and constructs its policy based on this.
Next, all features from different channels are concatenated and then passed through a residual block, consisting of two linear layers and a residual shortcut~\cite{he2016deep}, and fed into a long-short-term memory (LSTM) cell.
Finally, the agent can obtain the policy, value, and blocking through three linear layers. 

    \subsubsection{Attention-focusing Layer}
    Our motivation for designing the attention-focusing layer is to enable the agent to discern the varying importance of different regions on a map, which can improve the agent's policy.
    In this layer, the agent infers the dependencies between itself and all other nodes, further augmenting the extent of this discrimination.

    Specifically, the agent computes correlations with all other nodes in both the static and dynamic graphs, assigns attention weights, and subsequently applies them to the feature vectors.
    As an illustration, using the augmented static graph, at time step $t$, all input 5-dimensional feature vectors $v^5_{i,t}$ are transformed into $d$-dimensional embedding vectors $u_i$ via a linear layer.
    The ego agent embedding vector $u_{N+1}$ is extracted to calculate the query vector $q_a$ through the weight matrix $W^Q_{fa}$. 
    Similarly, the key vector $k_i$ is calculated by passing all embedding vectors through the weight matrix $W^K_{fa}$, allowing us to compute the ego agent's attention for all nodes using the following formulas:
    \begin{align}
        &q_a = W^Q_{fa}u_{N+1},~~k_i = W^K_{fa}u_i,~~s_{a,i} = \frac{q_a\cdot k_i^T}{\sqrt{d}},\notag\\
        &\alpha_{a,i}=\frac{e^{s_{a,i}}}{\sum_{i=1}^{N+2}e^{s_{a,i}}},~~i=1\cdots N+2
    \end{align}
    where $W_Q^{fa},W_K^{fa}\in\mathbb{R}^{d\times d}$ consist of learnable parameters of the neural network. 
    The obtained attention $\alpha_{a,i}$ is used to represent the agent's dependency for node $i$. 
    We use attention for strengthening the embedding vectors $u_i$ of interest and for weakening the embedding vectors that are not of interest:
    \begin{align}
        u'_i=\alpha_{a,i}u_i,~~i=1\cdots N+2
    \end{align}
    By stacking several attention-focusing layers ($n_f$ in Fig. \ref{net}), the agent's interest in different nodes can be significantly different.
    Agents will focus their attention on those regions that are already potentially of interest rather than viewing all nodes equally.
    Our encoder uses a self-attention layer~\cite{vaswani2017attention,kool2018attention} after the attention focusing layer to reason about the inter-dependencies of nodes in the augmented graphs and perform context learning.

We provide the computation process of the self-attention layers mentioned in this paper in the additional video, along with comprehensive training details and settings, including hardware configurations, loss function settings (RL and IL), and hyper-parameters choices\footnote{Code will be released upon paper acceptance.}.

\section{Experiments}

\begin{table*}[!ht]
  \vspace{2mm}
  \centering
  \caption{Experimental results. The notation "$\uparrow$" implies that a larger value is preferable, and vice versa.}
  \vspace{-2mm}
  \scalebox{0.7}{
    \begin{tabular}{l||cccccc||cccccc||cccccc}
    \toprule
    \textbf{Model} & \multicolumn{6}{c}{\textbf{MS}$\downarrow$} & \multicolumn{6}{c}{\textbf{AR}$\uparrow$} & \multicolumn{6}{c}{\textbf{SR}$\uparrow$}\\
    \cmidrule{2-19}

    & \multicolumn{18}{c}{\textbf{20 $\times$ 20 room-liked environment with 4, 8, 16, 32, 64, 128 agents}} \\
    
    \midrule\midrule    
    ODrM*  & 30.58  & 43.19  & 97.25  & 292.93 & 512.00 & 512.00 & 100\%   & 98.00\% & 88.00\% & 47.00\% &  0.00\% &  0.00\% & 100\% & 98\% & 88\% & 47\% &  0\% & 0\%  \\
    PRIMAL & 201.32 & 275.93 & 439.95 & 506.83 & 512.00 & 512.00 & 93.50\% & 90.88\% & 88.63\% & 81.72\% & 66.79\% & \textbf{35.74\%} &  79\% & 67\% & 30\% &  2\% &  0\% & 0\%  \\
    MAPPER & 79.81  & 101.05 & 246.69 & 427.33 & 512.00 & 512.00 & 98.75\% & 97.53\% & 95.75\% & 89.71\% & 53.92\% &  6.96\% &  97\% & 97\% & 82\% & 41\% &  0\% & 0\%  \\
    G2RL   & 42.22  & 65.46  & 159.12 & 356.94 & 511.45 & 512.00 & 98.00\% & 98.75\% & 98.00\% & 94.43\% & 68.50\% & 18.07\% &  99\% & 97\% & 87\% & 54\% &  1\% & 0\%  \\
    DHC    & 45.50  & 73.86  & 175.22 & 354.43 & 509.69 & 512.00 & 99.00\% & 98.62\% & 96.56\% & 90.69\% & 69.70\% & 20.09\% &  98\% & 93\% & 77\% & 45\% &  1\% & 0\%  \\ 
    SCRIMP & 43.42  & 61.56  & 186.34 & \textbf{214.32} & \textbf{488.98} & --     & 99.25\% & 99.37\% & 98.87\% & 97.53\% & \textbf{82.32\%} & --      &  98\% & 96\% & 93\% & 75\% & \textbf{15\%} & --   \\
    \midrule                                                                         
    ALPHA  & \textbf{37.39}  & \textbf{52.26}  & \textbf{120.65} & 310.21 & 503.87 & 512.00 & \textbf{100\%}   & \textbf{100\%}   & \textbf{99.75\%} & \textbf{97.69\%} & 70.12\% & 25.71\% & \textbf{100\%} & \textbf{100\%} & \textbf{96\%} & \textbf{78\%} &  8\% & 0\%  \\
    \midrule
    & \multicolumn{18}{c}{\textbf{40 $\times$ 40 room-like environment with 4, 8, 16, 32, 64, 128 agents}} \\
    \midrule\midrule
    ODrM*  & 56.73  & 69.34  & 91.89  & 146.88 & 375.37 & 512.00 & 100\%   & 100\%   & 97.00\% & 85.00\% & 32.00\% &  0.00\% & 100\% &100\% & 97\% & 85\% & 32\% & 0\% \\
    PRIMAL & 285.55 & 384.93 & 463.86 & 492.82 & 511.80 & 512.00 & 91.00\% & 87.62\% & 85.56\% & 82.69\% & 73.14\% & 61.71\% &  73\% & 47\% & 23\% & 11\% &  1\% & 0\% \\
    MAPPER & 104.82 & 157.12 & 218.35 & 348.95 & 491.58 & 512.00 & \textbf{100\%}   & 99.37\% & 98.00\% & 93.71\% & 76.60\% & 51.02\% & \textbf{100\%} & 96\% & 91\% & 66\% & 16\% & 0\% \\
    G2RL   & \textbf{57.60}  & 93.31  & 166.92 & 241.39 & 433.13 & 512.00 & \textbf{100\%}   & 99.75\% & 99.06\% & 98.65\% & 93.53\% & 70.50\% & \textbf{100\%} & 98\% & 89\% & 81\% & 33\% & 0\% \\
    DHC    & 104.19 & 127.78 & 188.62 & 263.81 & 427.02 & 512.00 & 97.75\% & 98.00\% & 97.88\% & 95.94\% & 91.17\% & 72.53\% &  92\% & 91\% & 80\% & 65\% & 28\% & 0\% \\
    SCRIMP & 58.53  & 91.84  & \textbf{116.05} & \textbf{183.54} & 396.93 & \textbf{484.76} & \textbf{100\%}   & 99.62\% & 99.56\% & 99.21\% & \textbf{94.10\%} & \textbf{85.09\%} & \textbf{100\%} & 97\% & 95\% & 84\% & 42\% & \textbf{12\%} \\
    \midrule
    ALPHA  & 64.04  & \textbf{88.75}  & 140.96 & 206.85 & \textbf{392.23} & 506.48 & \textbf{100\%}   & \textbf{100\%}   & \textbf{99.75\%} & \textbf{99.34\%} & 93.46\% & 73.99\% & \textbf{100\%} &\textbf{100\%} & \textbf{97\%} & \textbf{93\%} & \textbf{60\%} & 7\% \\
    \midrule
    & \multicolumn{18}{c}{\textbf{60 $\times$ 60 room-liked environment with 4, 8, 16, 32, 64, 128 agents}} \\
    \midrule\midrule
    ODrM*  & 84.71  & 98.43  & 106.46 & 163.53 & 228.95 & 457.17 & 100\%   & 100\%   & 99.00\% & 88.00\% & 72.00\% & 14.00\% & 100\% &100\% & 99\% & 88\% & 72\% & 14\% \\
    PRIMAL & 363.45 & 465.35 & 495.85 & 508.17 & 512.00 & 512.00 & 84.75\% & 78.37\% & 79.75\% & 73.62\% & 71.51\% & 62.83\% &  54\% & 25\% & 11\% &  3\% &  0\% & 0\%  \\
    MAPPER & 177.61 & 241.31 & 280.69 & 388.55 & 490.02 & 512.00 & \textbf{99.50\%} & 97.75\% & 98.31\% & 93.87\% & 85.96\% & 62.47\% &  97\% & 89\% & 90\% & 61\% & 17\% & 0\%  \\
    G2RL   & \textbf{104.40} & \textbf{140.85} & 168.70 & 280.04 & 431.21 & 512.00 & 99.00\% & 98.62\% & 97.36\% & 95.62\% & 93.76\% & 86.01\% &  96\% & 94\% & 91\% & 68\% & 34\% & 0\%  \\
    DHC    & 131.59 & 203.71 & 186.66 & 323.19 & 406.40 & 496.70 & 97.75\% & 96.75\% & 98.88\% & 95.16\% & 93.30\% & 87.79\% &  91\% & 77\% & 86\% & 54\% & 35\% & 7\%  \\
    SCRIMP & 106.79 & 166.37 & \textbf{125.50} & \textbf{211.03} & 421.65 & 498.72 & \textbf{99.50\%} & \textbf{99.25\%} & 99.61\% & 98.73\% & 96.79\% & 88.08\% &  \textbf{98\%} & 95\% & \textbf{97\%} & 81\% & 31\% & 8\%  \\
    \midrule
    ALPHA  & 110.82 & 158.59 & 173.02 & 263.74 & \textbf{357.17} & \textbf{485.23} & \textbf{99.50\%} & \textbf{99.25\%} & \textbf{99.75\%} & \textbf{99.03\%} & \textbf{97.91\%} & \textbf{89.16\%} &  \textbf{98\%} & \textbf{97\%} & \textbf{97\%} & \textbf{86\%} & \textbf{67\%} & \textbf{25\%} \\
    \bottomrule
\end{tabular}
  }
  \label{tab:results}
\end{table*}

In this section, we extensively test ALPHA through simulation experiments, compare its performance with state-of-the-art baselines, and conduct ablation studies to validate each proposed technique.
Additionally, we validate ALPHA by deploying a trained model in various simulation environments (Gazebo) and real-world scenarios.

\subsection{Comparison and Analysis}

During training, we randomly select the size of structured environments from a uniform distribution ranging from 10 to 40, while maintaining a consistent configuration of 8 agents.
While testing, we chose environments of sizes 20, 40, and 60, deploying 4 to 128 agents. 

We compare our method with five other state-of-the-art MAPF solutions, namely our previous work PRIMAL~\cite{sartoretti2019primal}, MAPPER~\cite{liu2020mapper} with global guidance, G2RL~\cite{wang2020mobile} with relaxed global guidance, graph neural network-based communication method DHC~\cite{ma2021distributed}, and transformer-based communication method SCRIMP~\cite{wang2023scrimp}.
We also present results of the search-based bounded-optimal centralized planner ODrM*~\cite{ferner2013odrm} (with inflation factor $\epsilon= 2.0$).
For each test, we generated a fixed set of 100 randomly-generated environments to evaluate these MAPF planners. 
In other words, all planners were tested on exactly the same set of environments, and the test results are presented in Table \ref{tab:results}.

We set three metrics to evaluate performance:
(a) Makespan (MS) which measures solution efficiency by counting the agents' actions to their goals in one episode.
(b) Success Rate (SR) which evaluates the ability of a planner to complete a task in its entirety.
(c) Arrival Rate (AR) which is the percentage of agents that reach their goals in all episodes. 
In learning-based methods, there are instances with a low SR, yet most agents reach their goals.
AR allows for a fair evaluation of nearly completed episodes, preventing them from being categorized as complete failures.

Compared to ODrM*, all learning-based planners exhibit significant advantages in terms of AR, which is crucial for life-long MAPF. 
ALPHA consistently outperforms PRIMAL in nearly all cases, likely due to PRIMAL's sole reliance on local information.
As the agent team size increases, the effectiveness of rigid global guidance provided by MAPPER and G2RL diminishes, leading to a decline in their performance.
Notably, even in our more complex tasks ($128$ agents, $60 \times 60$ map size, $~0.2$ obstacle density) demanding high agent cooperation, which MAPPER and G2RL were completely unable to solve, ALPHA achieved a $25\%$ SR.
This is likely due to the fact that MAPPER and G2RL, to some degree, force agents to follow paths/waypoints computed by A*, and this possibly reduces their effectiveness in denser environments.  

Another notable finding is that DHC and SCRIMP policies, at times, lead to certain agents getting stuck in corners because of the other agents. 
We believe that the same heuristic map used by DHC and SCRIMP causes this issue.
ALPHA, with its comprehensive grasp of the global map and policy flexibility, does not face this limitation.
Additionally, SCRIMP employs a tie-breaking strategy to enhance coordination, which is particularly effective in crowded cases.
However, this also results in SCRIMP needing much more time to compute a solution.
For example, in a MAPF problem with 64 agents in a map of size $20 \times 20$, ten episodes of SCRIMP take $12.29s$, while those of ALPHA take only $7.21s$ ($~41\%$ less time).
The test results of random and warehouse maps are shown in the additional video.

\vspace{-0.1cm}
\subsection{Ablation Study}

\begin{table}[!t]
  \centering
  \caption{Ablation Study}
  \scalebox{0.6}{
    \begin{tabular}{cccc|cc}
    \toprule              
    \textbf{Structural}   & \textbf{Intent}   & \textbf{Attention} & \textbf{Short-term}  & \textbf{Total}    & \textbf{Episode}         \\
    \textbf{Encoding}& \textbf{Prediction} & \textbf{Focusing} & \textbf{Intention} & \textbf{Reward}$\uparrow$ & \textbf{Length} $\downarrow$ \\
    \midrule\midrule                                 
    --                  & --                  & --                  &  --                  &    -249.760                  & 230.754                    \\
    \midrule
    \checkmark          & --                  & --                  &  --                  &    -104.412                  & 95.229                    \\
    \checkmark          & \checkmark          & --                  &  --                  &    -99.567                  & 78.931                    \\
    \checkmark          & \checkmark          & \checkmark          &  --                  &    -92.320                  & 77.481                    \\
    \checkmark          & \checkmark          & \checkmark          &  \checkmark          &    \textbf{-87.878}         & \textbf{70.071}           \\
    \bottomrule
\end{tabular}
  }
  \label{tab:ablation}
\end{table}
Our architecture is based upon two important ideas: global information and attention focusing, where global information includes static features for encoding the environment's structure and dynamic features for describing agent's intention. 
To analyze the importance of these three elements, we experimented with three ablation variants of ALPHA: 1) we removed the second graph and its encoder, using only static features for environment structure encoding, 2) incorporating dynamic features to predict agents' intentions, and 3) adding an attention focusing layer to enable the agent to differentiate the importance of different areas in the map. 
Furthermore, we attempted a fourth ablation variant of ALPHA by making shorter-term predictions (10 in practice) of agents' intentions.

The results of the ablation variants are shown in Table \ref{tab:ablation}. 
We observe that static features, which contribute to encoding the environmental structure, lead to significant performance improvements. 
Predicting agent intentions greatly enhances coordination and significantly reduces episode length. 
This aligns with our assertion that enabling the agent to understand global information facilitates better planning. 
Our attention focusing mechanism enhances the expressive power of the network. 
As mentioned in previous sections, short-term agent intention prediction outperforms long-term agent prediction, as it mitigates the inherent inaccuracies in A* prediction to some extent.

\vspace{-0.1cm}
\subsection{Experimental Validation}

\begin{figure}
\centering
\includegraphics[width=2.5in]{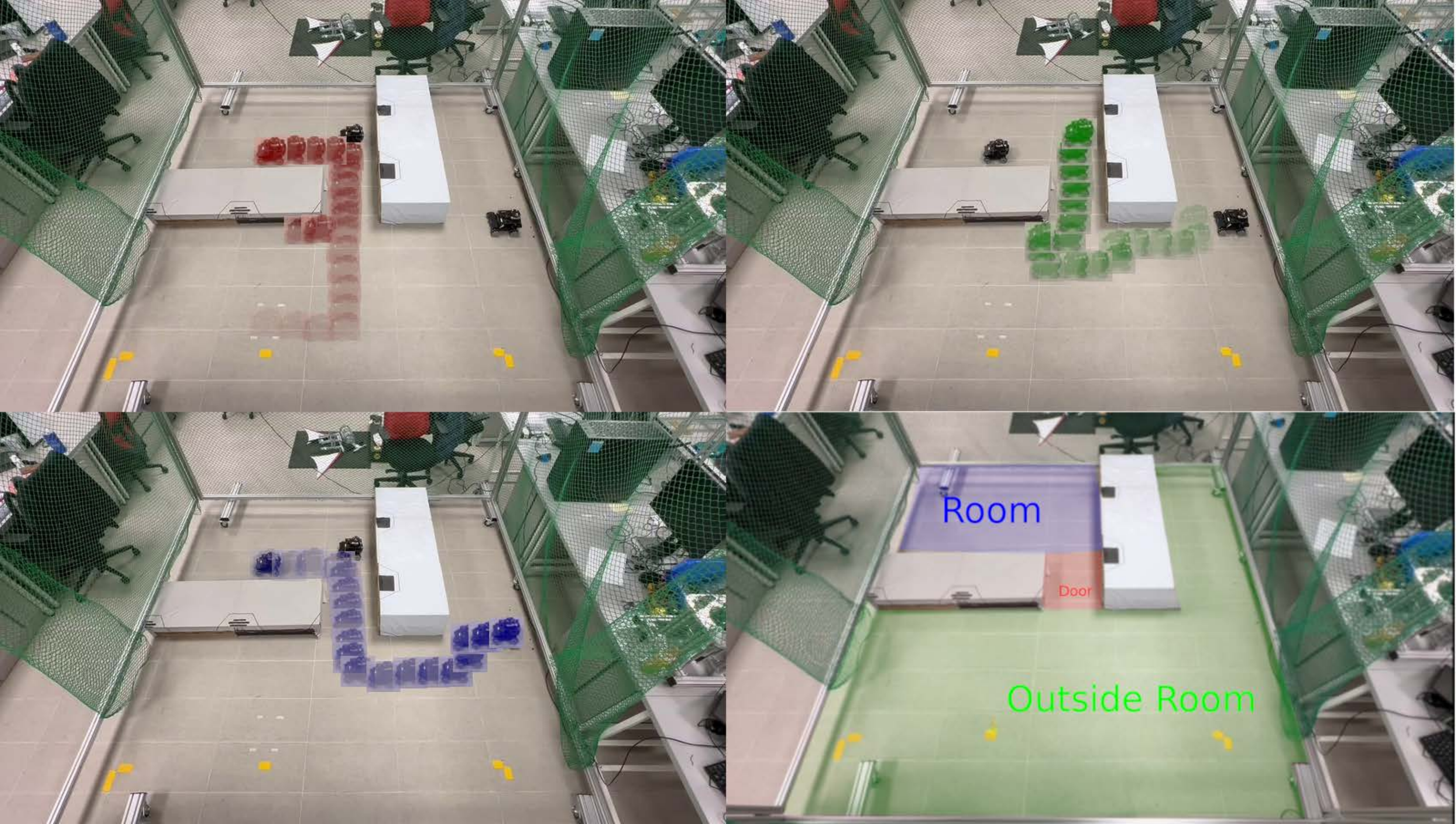}
\vspace{-0.3cm}
\caption{Illustration of the environment and the paths of the three agents.}
\label{exp}
\vspace{-0.4cm}
\end{figure}

To illustrate the ability of ALPHA to be deployed in the real world, we test it in three standard gazebo simulation environments provided by AWS robotics\footnote{https://github.com/aws-robotics}. 

Additionally, we conduct real-world experiments using three mecanum-wheeled robots ($0.2m \times 0.23m$) in an environment measuring $3.25m \times 3.25m$ containing two long, wall-like obstacles arranged to create a room with a single entrance (see Fig. \ref{exp}).
The agents follow policies generated by our trained ALPHA model and successfully reach their goals without collision.
For more simulation and real-world experiment details, refer to the additional video.

\section{Conclusion}

In this paper, we propose ALPHA, a novel MAPF planner designed to encode and utilize more comprehensive global information for long-horizon planning, to go beyond the use of limited FoVs in most existing learning-based MAPF planners.
Our proposed augmented graph-based global observations capture both obstacle structure and agent intentions, facilitating better representation learning through these abstractions.
We further utilize an attention-focusing mechanism to enhance map understanding and context learning, aiding agents in recognizing different regions' importance.
Through a comprehensive range of experiments conducted on highly structured maps of varying team and world sizes, ALPHA consistently demonstrates superior performance compared to state-of-the-art learning-based MAPF baselines and a bounded-optimal search-based planner across the majority of scenarios.
In doing so, we aim to offer a fresh perspective to the MAPF research community by highlighting that utilizing more effective methods for encoding a broader spectrum of global information can yield performance surpassing communication learning.

Future work will focus on refining representation learning to enhance the global map representation, with a particular emphasis on harnessing edge information within the graph for improved planning and coordination.

\clearpage
\bibliographystyle{IEEEtran}
\bibliography{icra24}

\clearpage
\section*{Appendix}

\subsection{Action and Reward}

    \subsubsection{Action Space}

    The action space of the agent comprises five elements, specifically the four cardinal directions (up, down, left, right) and the option to remain idle.
    According to the setting, agents can only select actions from the valid action list during training, in which action staying idle is always allowed.
    The valid action list is set to prevent the agent from colliding with obstacles or going out of the map range and returning to the previous position (preventing the generation of oscillation policies and live locks).
    Notably, we do not consider conflicts between agents as invalid actions because we want agents to learn to coordinate flexibly through training rather than forcing them to adopt a fixed paradigm through supervised learning.
    During execution, if the agent accidentally selects an invalid action, then the agent will instead execute stay idle.
    However, there is no need to worry about this possibility because the selection of a valid action is achieved by setting the loss function, which is a form of supervised learning.
    The agent can quickly learn to select the valid action from the action space.
    Finally, unlike PRIMAL, another improvement is that all the agents' actions are synchronized in the joint space rather than occurring sequentially.

    \subsubsection{Reward Structure}

    Intuitively, to encourage agents to explore the map instead of sticking to their current location, the penalty for staying still is often set slightly higher than the penalty for moving.
    However, given that room-based maps contain a large number of doors for only one agent to pass through and long corridors of varying widths, it is easy for agents to block each other.
    To increase agents' patience, keeping them from moving meaninglessly through the environment is important to avoid potential blocking that may result in coordination difficulty.
    Hence, we assign an identical penalty for both agent movement and remaining idle.
    In addition, high arrival reward might encourage the agent to make some bad decisions, so we remove this setting, and accordingly, the agent does not receive any moving penalty if it stays on its goal.
    In the event of any collision, the agent is subjected to a relatively significant penalty.

    As mentioned earlier, the room-based maps in our work are prone to blocking due to their highly structured nature.
    We set an extra penalty for all agents that choose to stay idle when they are staying in position blocks other agents.
    The way to determine if an agent is blocking other agents is to: first, use the A* algorithm to compute the path, called $\mathcal{P}_{before}^{A*}$, for the other agents to reach the goal from its current location; then, remove the ego agent from the map and again use the A* algorithm to compute the path from the current position to the goal for the other agent, called $\mathcal{P}_{after}^{A*}$.
    If $\mathcal{P}_{before}^{A*}$ does not exist and $\mathcal{P}_{after}^{A*}$ exists, or if $\mathcal{P}_{before}^{A*}>\mathcal{P}_{after}^{A*}+\tau$, then we consider that the ego agent blocks other agents, where $\tau$ is a manually set parameter, by which we can control the tolerance of blocking.
    The ego agent will be penalized by the $blocking~penalty \times \eta$, where $\eta$ is the number of blocked agents. 

    \subsection{Self-attention Layer}

    Besides the attention-focusing layer mentioned in our main text, another important part of our encoder is the self-attention layer.
    The output $u'_i$ of the attention-focusing layer is used as the input of the self-attention layer, and $W^Q_{sa}$, $W^K_{sa}$ and $W^V_{sa}$ are used to convert the embedding vector $u'_i$ into query $q_i$, key $k_i$ and value $v_i$.
    Then the score and attention are calculated from the query $q_i$ and key $k_i$, and finally, the weighted sum of attention and value is used as the output of the self-attention layer:
    \begin{align}
        &q_i = W_{sa}^Qu'_i,~~k_i=W_{sa}^Ku'_i,~~v_i=W_{sa}^Vu'_i\notag\\
        &s_{ij} = \frac{q_i\cdot k_i^T}{\sqrt{d}},~~
         \beta_{ij} = \frac{e^{s_{ij}}}{\sum_{j=1}^{N}e^{s_{ij}}},~~
         z_i = \sum_{j=1}^N \beta_{ij}v_j\notag\\
        & ~~~~~~~~~~~~~~~s.t.~~ i,j = 1,\cdots,N
    \end{align}
    where $N$ means the number of nodes in the map. 
    The output $z_i$ is used as the input of the next self-attention layer, thus stacking multiple self-attention layers.
    In the self-attention layer, the multi-head attention mechanism is used (8 heads in practice).
    Finally, the feature vector $z_{final}^a$ corresponding to the ego agent of the last layer is extracted as the output of the entire encoder.

\subsection{Training Details}

    This subsection describes some details of RL and IL training processes.
    We also show some training details and settings.

    \subsubsection{RL Training}

    As one of the most state-of-the-art reinforcement learning frameworks, Proximal Policy Optimization (PPO2) has the advantages of high stability, easy hyperparameter tuning, and good performance compared to other reinforcement learning algorithms.
    In this work, we use PPO2 as the RL algorithm instead of the A3C algorithm.
    The loss function of the network can be represented as:
    $J_{loss}(\theta)=\alpha J_{\pi}(\theta)+\beta J_V(\theta) + \iota \mathcal{H}(\pi(\cdot|o_t)) + \zeta J_{B}(\theta) + \eta J_{valid}(\theta)$, where $\alpha, \beta, \gamma, \zeta, \eta \in \mathbb{R}$ are the empirical manual-set hyperparameters.
    The policy loss is defined by the following formula:
    \begin{align}
        J_{\pi}(\theta) = \hat{\mathbb{E}}_t[-\min(r_t(\theta)\hat{A}_t,clip(r_t(\theta), 1-\epsilon,1+\epsilon))]
    \end{align}
    where $r_t(\theta)=\frac{\pi_{\theta}(a_t|o_t)}{\pi_{\theta_{pre}}(a_t|o_t)}$ denotes the ratio between the old and new policies, and $\epsilon$ is a hyperparameter.
    $\hat{A}_t=Norm(G_t-V(o_t|\theta_{pre}))$ is the estimation of the advantage function.
    The critic/value loss is defined as an L2 loss:
    \begin{align}
        J_V(\theta) = \hat{\mathbb{E}}_t[(V(o_t|\theta)-G_t)^2]
    \end{align}
    The entropy loss in our work is in line with the following form:
    \begin{align}
        \mathcal{H}(\pi_\theta(\cdot|o_t)) &= \sum_{a_t}\pi_\theta(a_t|o_t)\log(\pi_\theta(a_t|o_t))\notag\\
        &=\hat{\mathbb{E}}_{a_t\sim\pi}[\log(\pi_\theta(a_t|o_t))]
    \end{align}
    The blocking loss is used to improve the accuracy of the blocking results predicted by the network: $J_{B}(\theta) = -\hat{\mathbb{E}}_t[\rho_t\log(\varrho(b_t|o_t;\theta))+(1-\rho_t)\log(1-\varrho(b_t|o_t;\theta))]$, 
    where $\varrho(b_t|o_t;\theta)$ is the blocking prediction generated by the neural network, and $\rho_t$ is the ground truth of blocking states.
    The valid loss is a supervised learning loss item by which agents can learn to select actions from a valid action list: $J_{valid}(\theta) = -\hat{\mathbb{E}}_t[\upsilon_t\log(\overline{\pi}_\theta(a_t|o_t))+(1-\upsilon_t)\log(1-\overline{\pi}_\theta(a_t|o_t))]$. 
    $\overline{\pi}_\theta(a_t|o_t)$ obtained by letting policy $\pi_\theta(a_t|o_t)$ go through the activation function Sigmoid, and $\upsilon_t$ is an indicator vector of the same dimension as the policy $\pi_\theta(a_t|o_t)$ to indicate which actions are valid.

    \subsubsection{IL Training}
    We invoke the expert algorithm ODrM* with a certain probability to generate expert paths to guide the agents to make near-optimal choices.
    When in IL mode, the loss function of the network takes the following form:
    \begin{align}
        J_I(\theta) = -\sum_t \omega_t \log(\pi_\theta(a_t|o_t))
    \end{align}
    where $\omega_t$ represents the near-optimal policy generated by ODrM*. 
    The above equation shows that the imitation loss $J_I(\theta)$ used for IL is actually the cross entropy between the expert policy and the neural network-generated policy. 
    By minimizing the IL loss, the agent can imitate the behavior of experts to make better decisions.

    \subsubsection{Details \& Settings}

    In order for agents to learn a robust and generalized policy, environment sampling is used during training.
    Specifically, agents in different episodes during training will be in different environments.
    The size of the environment will be sampled from 10 to 40, and the density of obstacles will vary from 20\% to 40\%.
    For training, we used 8 agents, and the maximum number of steps in an episode for each agent was 256.
    If all agents reach their respective goals, then the episode ends early.
    We used a discount factor ($\gamma$) of 0.95, a clip range ($\epsilon$) of 0.2 for PPO2, and a batch size of 256.
    The manual parameters for loss are set to $\alpha=1, \beta=0.08, \iota=0.01, \zeta=0.5, \eta=0.5$.
    We use the Adam optimizer with a learning rate as $1\times 10^{-5}$.
    The embedding dimension $d=512$ is chosen in our attention network encoder.
    To speed up the data collection process, we use Ray, a distributed framework for machine learning.
    This framework can generate multiple parallel environments for data collection, and we set up 32 of them.
    Our training was performed on a workstation with 4 NVIDIA GeForce RTX 3090 GPUs, so we leverage \texttt{nn.DataParallel} to fully utilize the computational resources and accelerate the training process.

\subsection{Comparison Visualization}

In this subsection, we provide more detailed explanations of the selected baselines, highlighting the characteristics, advantages, and limitations of each approach. 
Subsequently, we visualize the experimental results and conduct a more comprehensive analysis. 

\subsubsection{Baselines Introduction}

    We have chosen five learning-based baselines, namely PRIMAL, MAPPER, G2RL, DHC, and SCRIMP, each planner possessing respective characteristics, advantages, and limitations. 

    PRIMAL is one of the most renowned learning-based MAPF planners, known for its strength in being an entirely local and decentralized planner. 
    It achieves efficient pathfinding and coordination solely through local observations and goal positions without requiring any global information or explicit communication. 
    This makes PRIMAL a cost-effective, scalable, and efficient solution for scenarios with loosely scattered obstacles.
    However, when dealing with continuous structured obstacles, PRIMAL may struggle to perform well due to the significant impact of obstacles beyond its field of view (FOV) on current decision-making. 
    During the implementation process, we made specific improvements to PRIMAL to enhance its performance.
    Firstly, we replaced the previous A3C reinforcement learning algorithm with the more powerful Proximal Policy Optimization (PPO) algorithm. 
    Additionally, in our baseline, agent movements are updated in a joint action space instead of the sequential movement of agents in the original PRIMAL implementation.

    The most significant difference between MAPPER and PRIMAL is that MAPPER is no longer an entirely local planner. 
    MAPPER incorporated global information as a guiding factor, enhancing agents' ability to navigate complex obstacles more easily.
    Specifically, MAPPER requires agents to follow their A* paths. 
    If an agent deviates from the A* path, it will receive different penalties based on the extent of the deviation. 
    Additionally, the A* path is divided into two parts: within the FOV and outside the FOV. 
    Agents can only observe the global guidance within the FOV.
    As a result, agents are trained to strictly adhere to the A* path and avoid any deviations. 
    This leads to rigid policies and reduces coordination levels among agents. 
    This planner can approach optimality in scenarios with smaller team sizes and fewer obstacles. 
    However, in more complex environments, rigid policies may cause conflicts and decrease performance.
    To compare the expressiveness of the global information provided by various MAPF solutions more fairly, we combine the global guidance of MAPPER with the local observations of PRIMAL as a baseline. 

    G2RL is an improvement over MAPPER. 
    Unlike MAPPER, which requires "strict obedience" to the A* path, G2RL introduces a modified reward structure that allows agents to explore alternative paths out of the A* path.
    In this approach, agents treat the A* path as a sequence of reward points, and reaching each point yields a positive reward.
    If an agent reaches any point on the A* path, it receives all the rewards associated with the points preceding it on the path. 
    With this flexible and dense reward structure, agents are no longer strictly bound to follow the A* path.
    However, similar to MAPPER, G2RL still projects global guidance into the FOV. 
    This limitation still restricts agents' understanding of the global environment to some extent. 
    As a result, although G2RL offers more flexibility than MAPPER in path planning, it may not fully leverage the advantages of global information for complex coordination and navigation tasks.

    \begin{figure}
    \centering
    \subfigure[success rate - world size 20 $\times$ 20]{\label{sr_20}
    \includegraphics[width=0.88\linewidth]{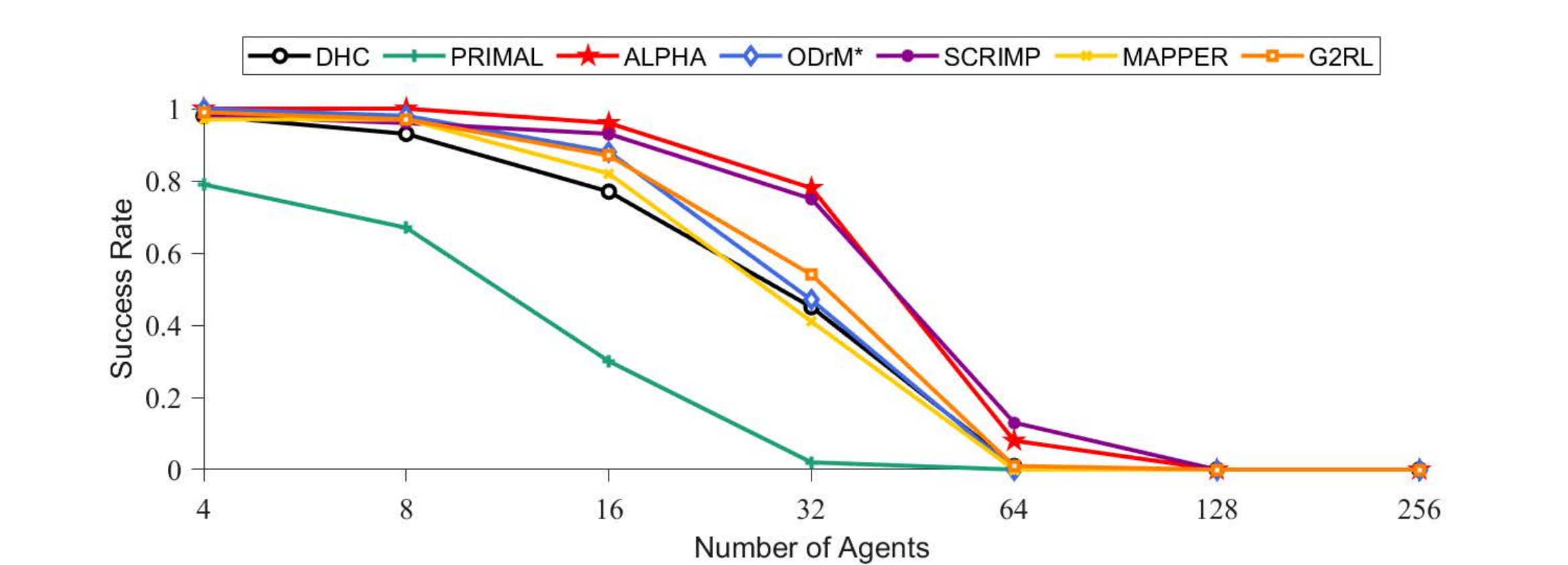}}
    \subfigure[success rate - world size 40 $\times$ 40]{\label{sr_40}
    \includegraphics[width=0.88\linewidth]{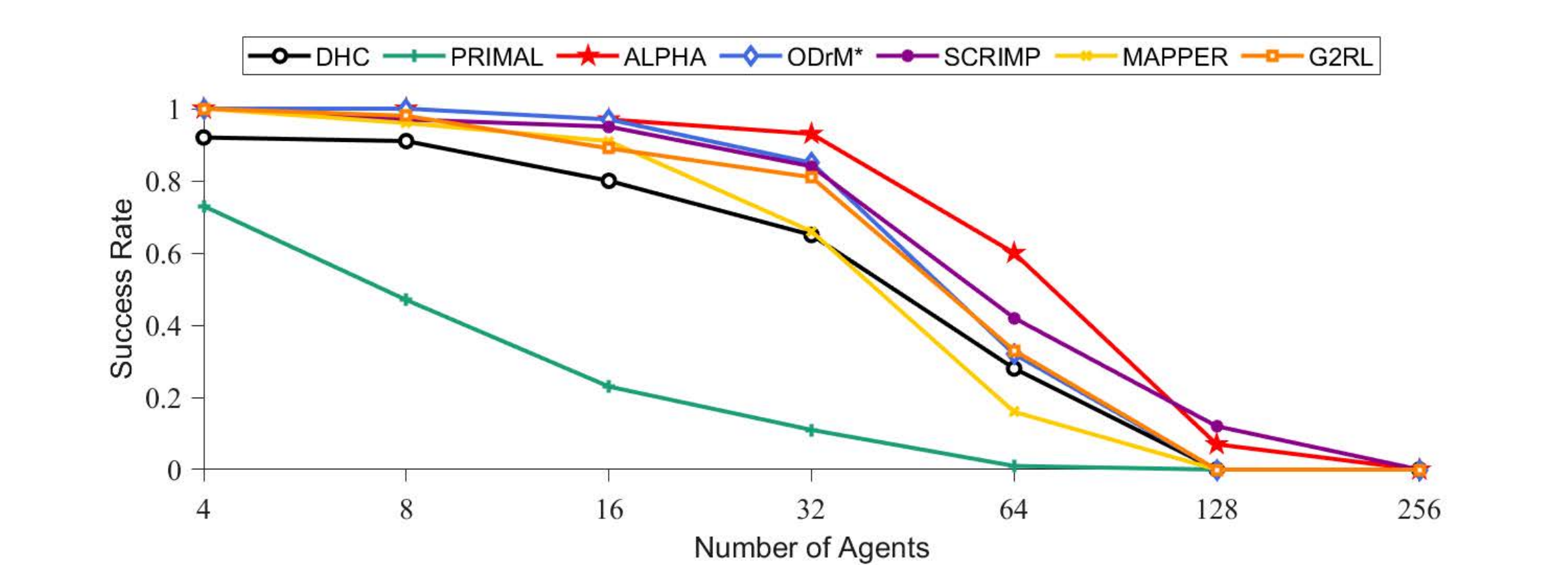}}
    \subfigure[success rate - world size 60 $\times$ 60]{\label{sr_60}
    \includegraphics[width=0.88\linewidth]{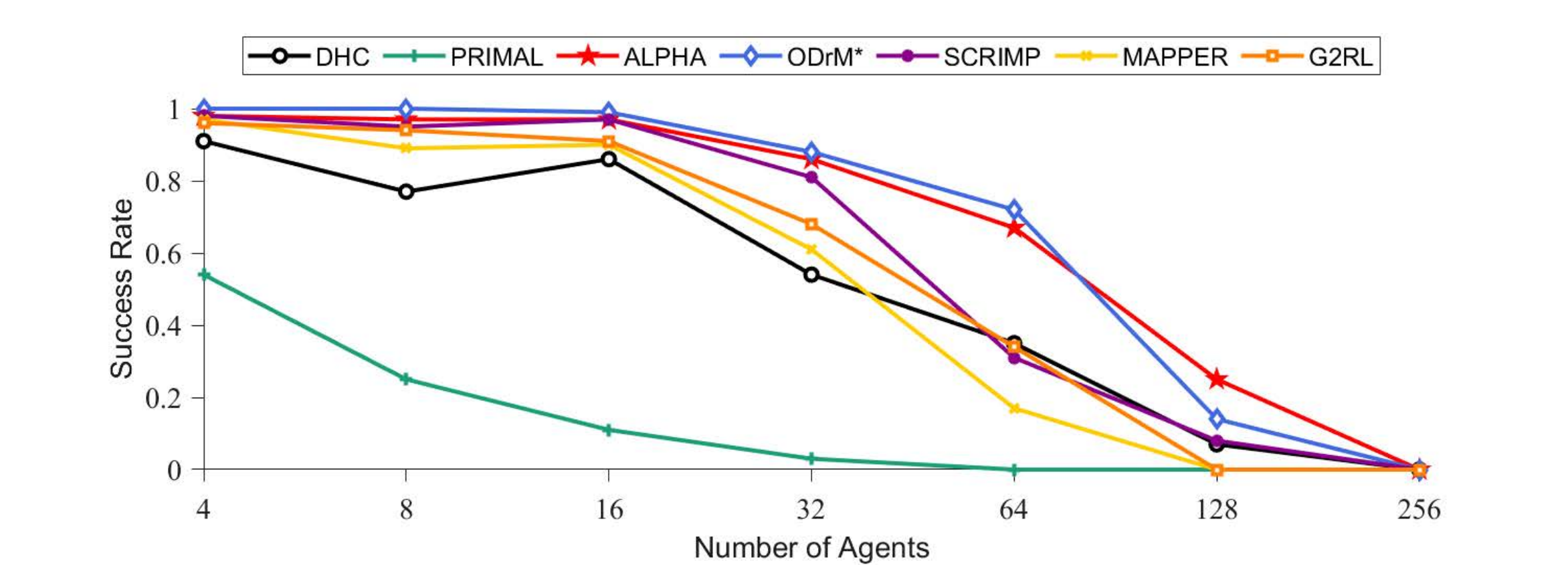}}
    \caption{
        The success rates of different planners under different team sizes and map sizes. ALPHA has demonstrated performance beyond other baselines in the vast majority of cases.
    }
    \label{sr}
    \end{figure}
    
    \begin{figure}
    \centering
    \subfigure[arrival rate - world size 20 $\times$ 20]{\label{ar_20}
    \includegraphics[width=0.88\linewidth]{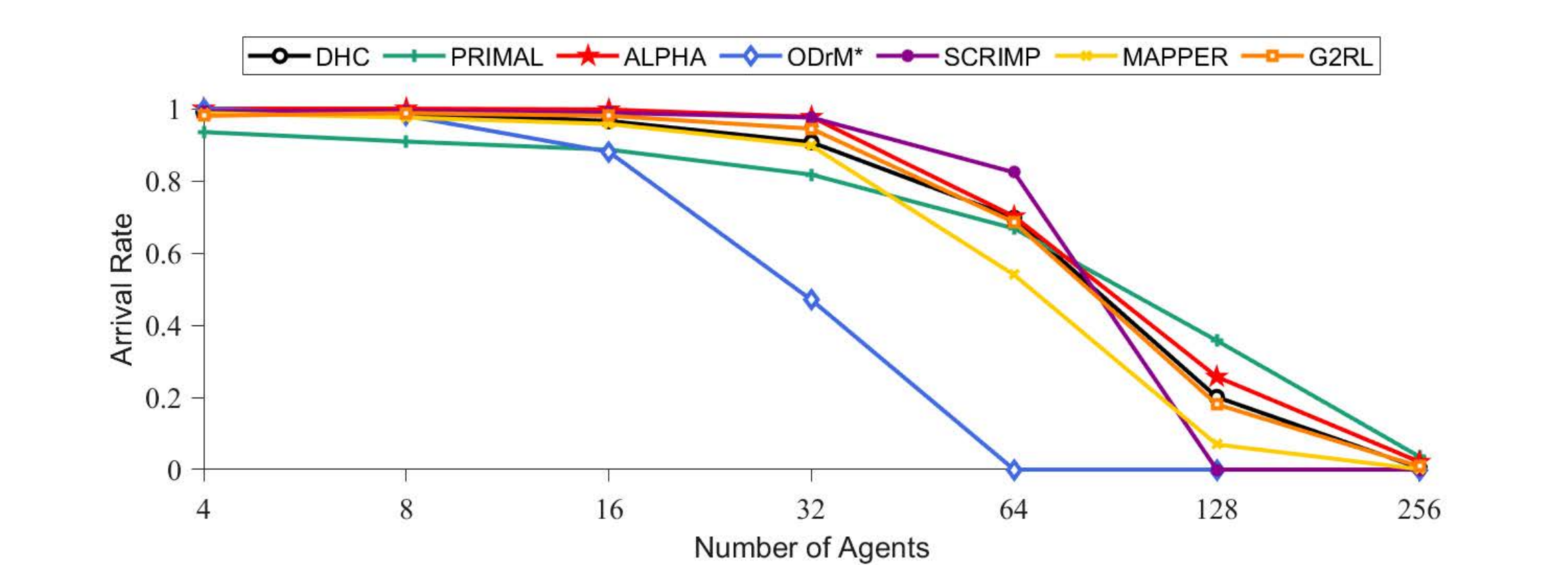}}
    \subfigure[arrival rate - world size 20 $\times$ 20]{\label{ar_40}
    \includegraphics[width=0.88\linewidth]{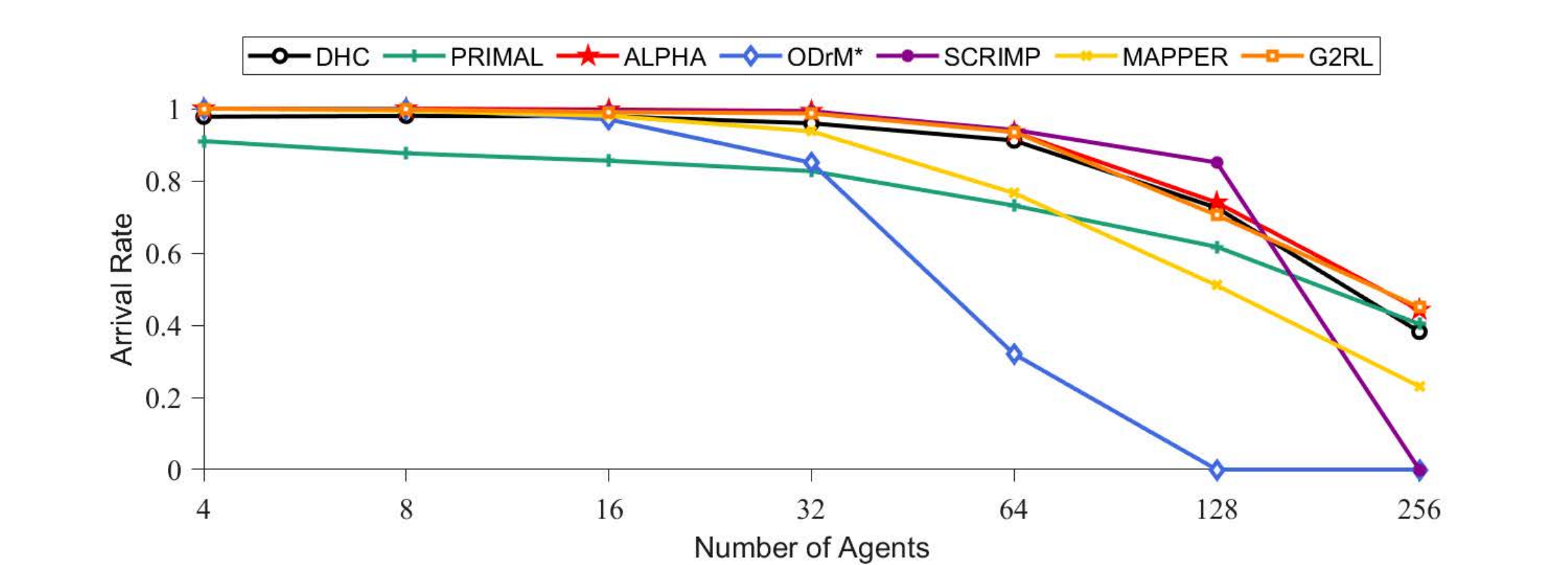}}
    \subfigure[arrival rate - world size 20 $\times$ 20]{\label{ar_60}
    \includegraphics[width=0.88\linewidth]{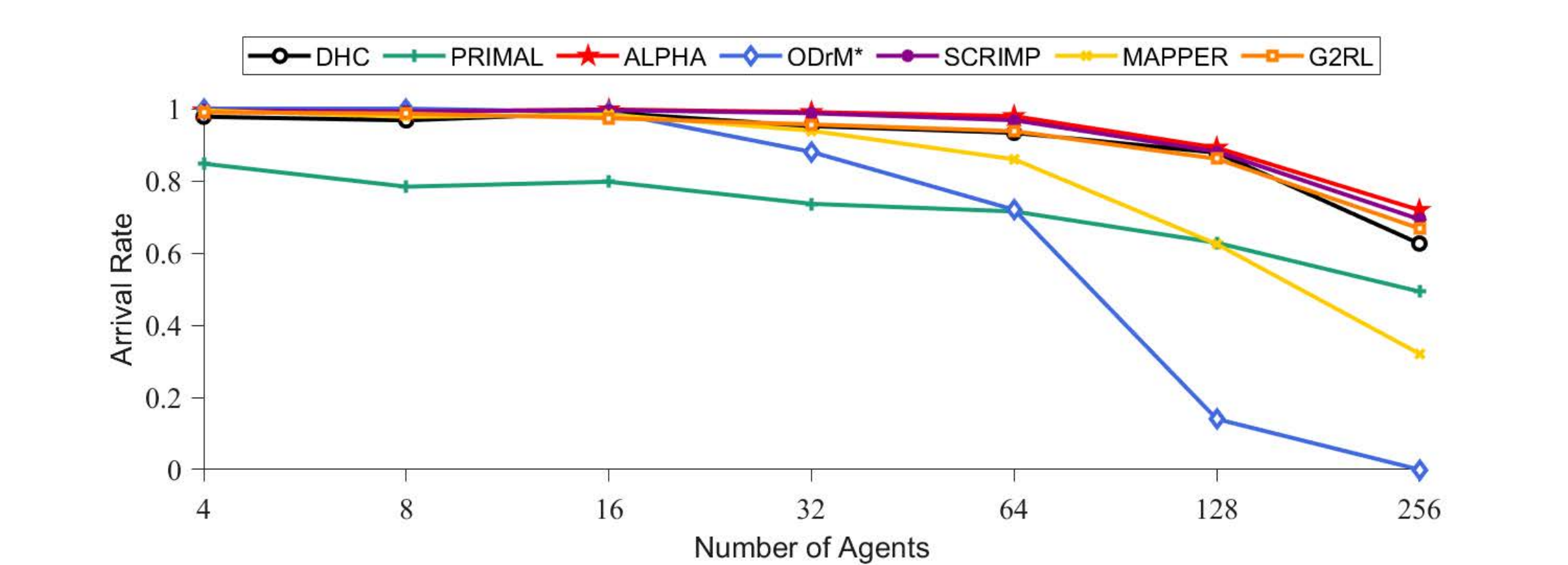}}
    \caption{
        The arrival rates of different planners under different team sizes and map sizes. ALPHA surpasses other baselines without communication modules. Even compared to baselines with communication modules, ALPHA still exhibits highly competitive performance.
    }
    \label{ar}
    \end{figure}

    DHC goes beyond just providing global guidance; it also incorporates a communication learning module. 
    The global guidance in DHC differs from MAPPER and G2RL, as it offers agents heuristic maps instead of a simple A* path.
    The heuristic map is divided into four channels, each corresponding to one action (up, down, left, right) for the agent. 
    In each channel, cells are represented by 0 and 1, indicating whether selecting the corresponding action in that cell will bring the agent closer to the goal. 
    Compared to providing a fixed A* path, this heuristic map, based on Breadth-First Search (BFS), offers agents more diverse and flexible strategies.
    However, DHC still projects the global information into the agent's FOV, resulting in certain limitations. 
    Additionally, DHC includes a GNN-based communication module, enhancing coordination among agents.

    SCRIMP utilizes the global information from DHC and further improves communication efficiency. 
    Its communication module is based on a transformer architecture, enabling more efficient communication and enhancing agent coordination.

    In summary, all the current solutions provide global guidance still based on FOV.
    This will result in the agent's decision-making being constrained by the limited FOV, making it impossible for the agent to perform long-horizon planning.
    Our proposed ALPHA overcomes this limitation by encoding global information in a graph-based representation, allowing agents to gain a global understanding of the full environment beyond the limitation of FOV.

\subsubsection{Performance Metrics}

    Success rate (SR): It is used to measure the ability of the planner to complete a MAPF task, and the following formula can express its calculation:
    \begin{align}
    sr = \frac{s}{N}
    \end{align}
    where $s$ represents the number of episodes in which all agents reach the goal within the specified number of steps, and $N$ is the total number of episodes tested.

    Arrival Rate (AR): It is used to measure the ability of the planner to complete lifelong tasks. A higher arrival rate means that the planner has better potential for lifelong MAPF tasks.
    \begin{align}
    ar = \sum_{i=1}^N\frac{arr_i}{N\times n}
    \end{align}
    Among them, $n$ represents the number of agents in each episode, and $arr_i$ represents the number of agents reaching the goal in the $i$-th episode.

    Episode Length (EL): Used to measure the optimality of the solution produced by the planner. A shorter episode length means that all agents can reach the goal faster.
    \begin{align}
    el = \sum_{i=1}^N\frac{pl_i}{N}
    \end{align}
    $pl_i$ represents the path length the last agent requires to reach the goal when all agents in the $i$-th episode reach the goal.
        
\subsubsection{Results Analysis}

    First, we visualize the most important metric in the MAPF problem (shown in Fig \ref{sr}), the success rate.
    Success rate, as the most widely used metric of MAPF, can show the ability of a planner to complete a task thoroughly in a given environment.
    Even if only one agent fails to reach the goal, the MAPF task is considered failed.
    From Fig \ref{sr}, we can observe that ALPHA outperforms other baselines in most cases.
    Because the agent must follow the A* path and does not have a communication module in MAPPER, the success rate drops very quickly as the number of agents increases.
    In contrast, despite adopting almost the same global guidance, G2RL obtains a very large advantage over MAPPER in agent dense environments due to its policy flexibility.
    However, their global information is strongly biased and limited since they still rely heavily on the A* path.
    DHC and SCRIMP adopt more comprehensive global information. But this kind of global information is still based on limited FOV, which is still short-sighted and may mislead agents.
    ALPHA uses a graph-based representation to feed global information into the agent instead of only considering the limited FOV centered on the agent.
    Only in the case of very large agent density, ALPHA may show lower performance than SCRIMP in some very extreme cases due to a lack of an explicit communication module.
    In addition, as the environment size increases, it can be seen that the performance advantages exhibited by ALPHA gradually expand.
    All in all, the results show that the global information provided by ALPHA is encoded in a more expressive way than previous methods, which is consistent with our intuition and guess.

    We also visualize the arrival rate, as shown in Fig \ref{ar}.
    We specifically add this metric because it would be unfair to consider almost completed tasks as failures.
    All learning-based baselines exhibit better performance than ODrM* as the agent team size increases.
    This is because learning-based methods update and solve the problem on a rolling way, unlike the one-time computation of traditional methods.
    This leads to learning-based planners that naturally outperform traditional methods in terms of arrival rate.
    At large team sizes and small map sizes, SCRIMP sometimes outperforms ALPHA due to the presence of a communication module.
    However, due to communication/coordination, SCRIMP consumed much more time than other baselines during the testing process.
    As the environment grows larger, ALPHA regains its performance advantage due to its ability to access information beyond the FOV, leading to a better understanding of global information and the ability to perform long-horizon planning.

\subsection{Numerical Simulations}

    \begin{figure}
    \centering
    \subfigure[slam map]{\label{slam_map}
    \includegraphics[width=0.45\linewidth]{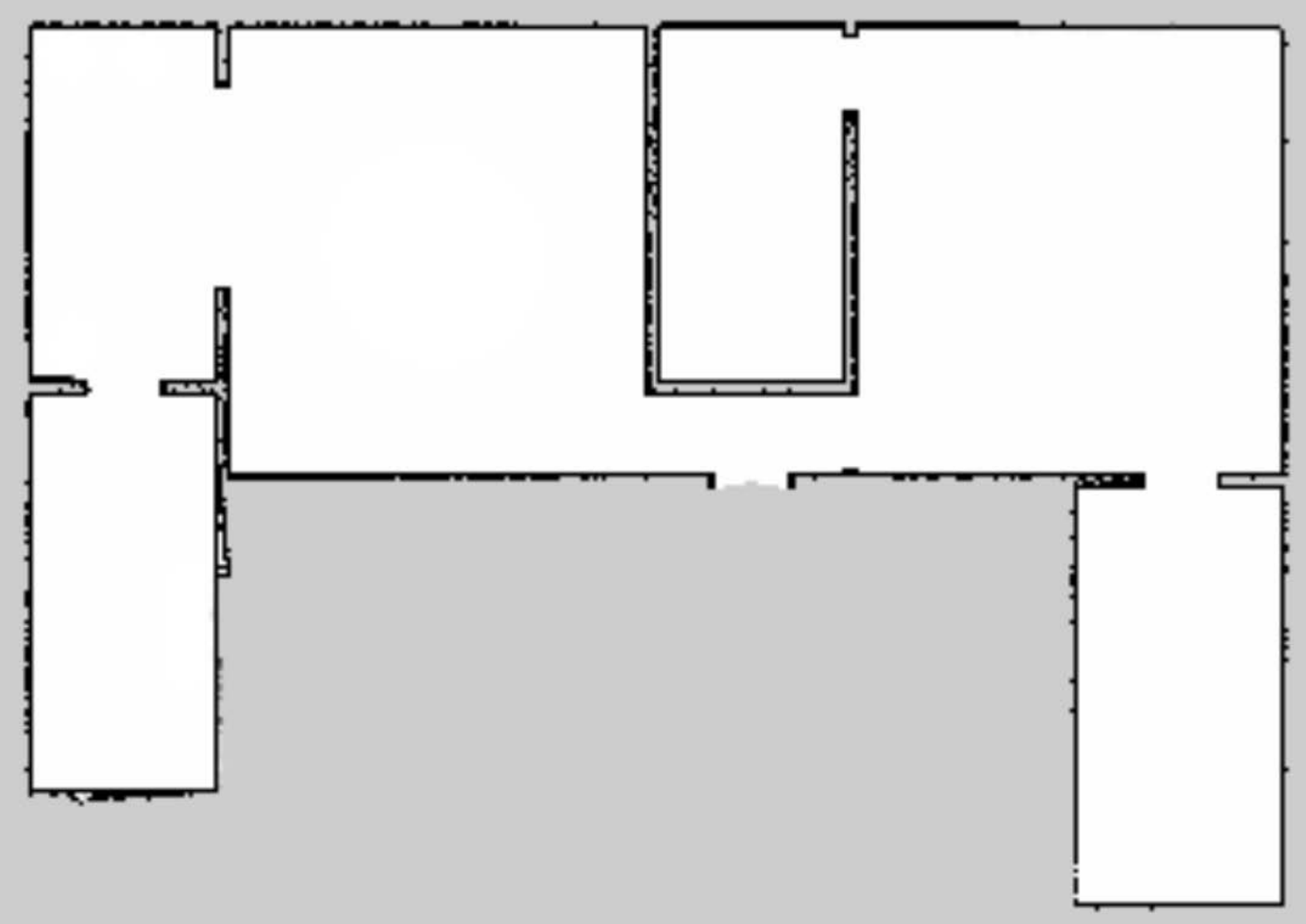}}
    \hspace{0.01\linewidth}
    \subfigure[simulation environment]{\label{simulation environment}
    \includegraphics[width=0.45\linewidth]{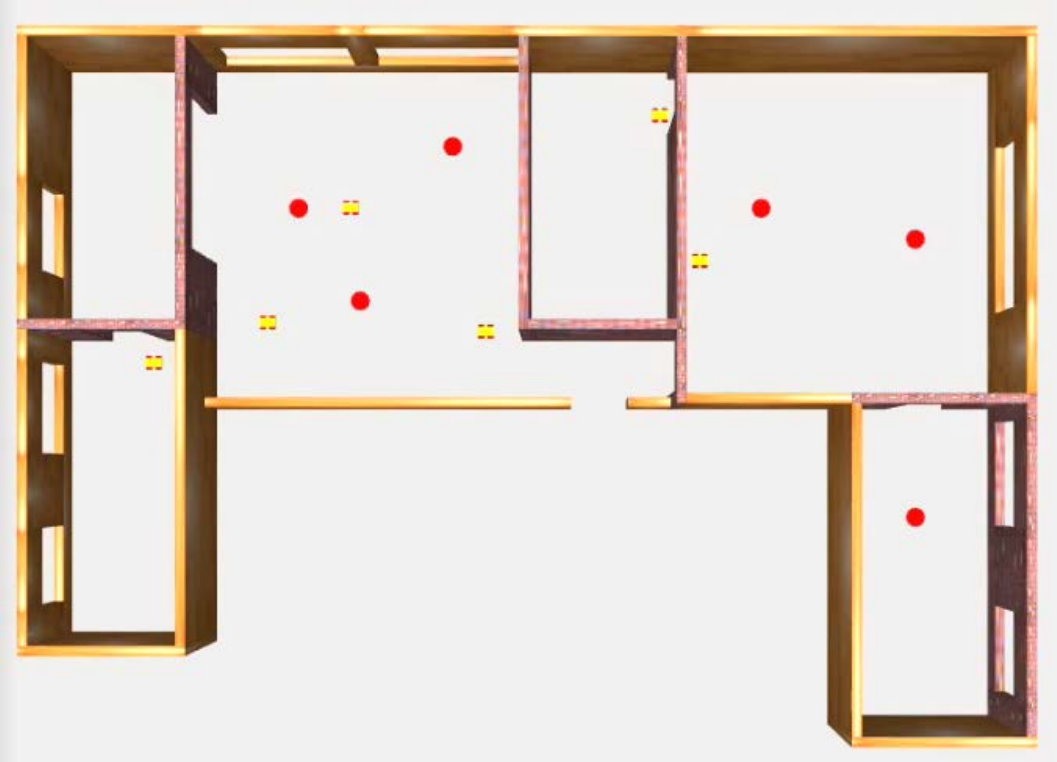}}
    \caption{
        Experimental validation. (a) Build a map through slam, and control the agent to go to the goal based on the map. (b) Simulation in Gazebo, where the red dots represent the goals of the agents and the yellow squares represent the agents equipped with mecanum wheels.
    }
    \label{sim}
    \end{figure}

We deployed six agents based on Gazebo in the room shown in Fig. \ref{sim} with a size of about $0.5m\times 0.4m$ and equipped with mecanum wheels, all of which are controlled by PID controllers.
The room size is about $11m\times 32m$, which contains complex structures such as corridors and doors with a width of only $0.7m$.
The agents move through the receding-horizon planner provided by our model.
Although the map is quite different from the grid world used in our training, the agents are still able to coordinate and reach the pre-defined goal.
This validation illustrates the ability of our provided algorithm to be deployed to the real world.

\subsection{Attention Visualization}

\begin{figure*}[t!]
\centering
\subfigure{
\begin{minipage}[t]{0.1\linewidth}
\centering
\includegraphics[width=2cm, height=2cm]{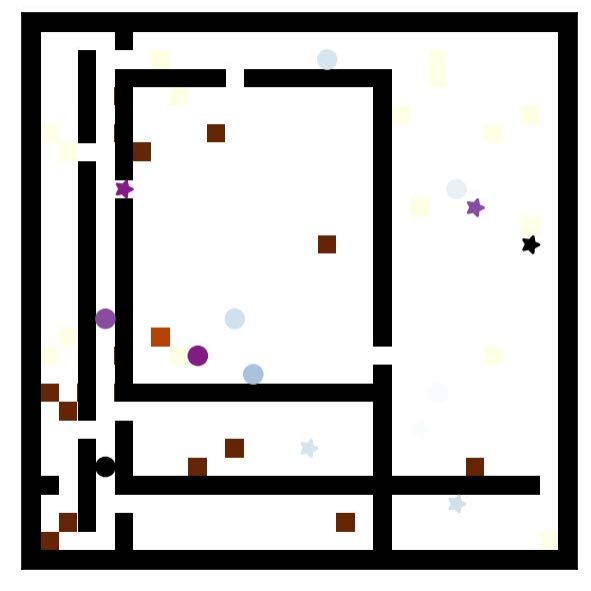}
\end{minipage}
}
\subfigure{
\begin{minipage}[t]{0.1\linewidth}
\centering
\includegraphics[width=2cm, height=2cm]{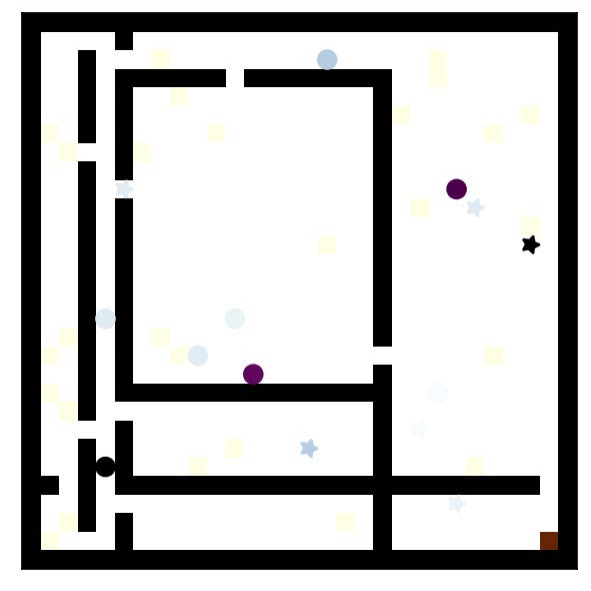}
\end{minipage}
}
\subfigure{
\begin{minipage}[t]{0.1\linewidth}
\centering
\includegraphics[width=2cm, height=2cm]{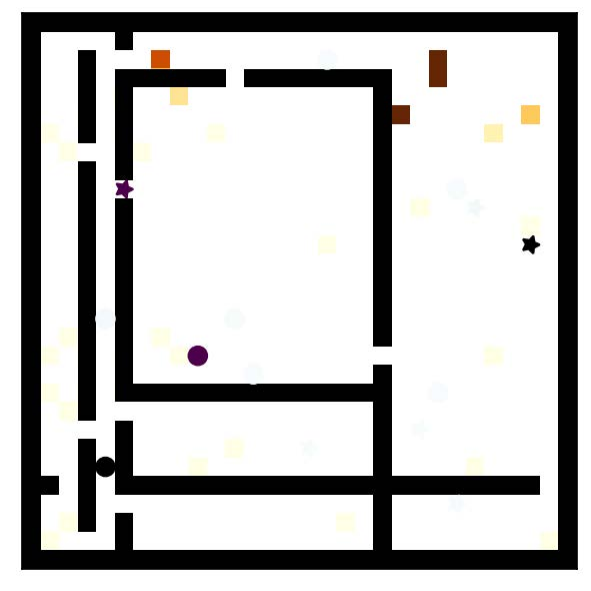}
\end{minipage}
}
\subfigure{
\begin{minipage}[t]{0.1\linewidth}
\centering
\includegraphics[width=2cm, height=2cm]{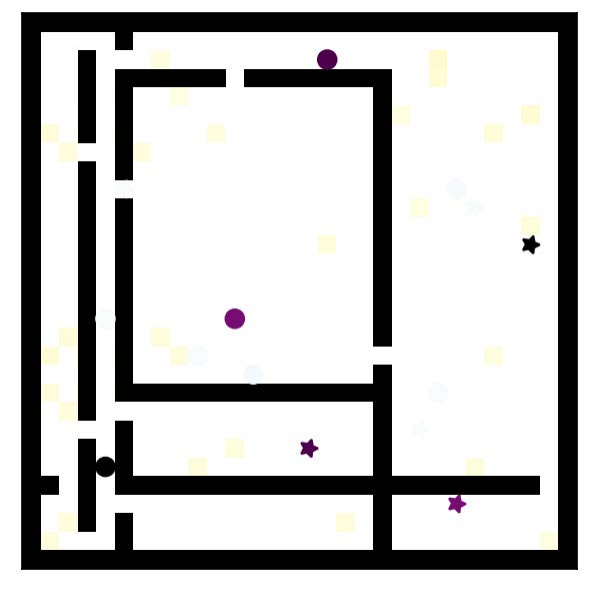}
\end{minipage}
}
\subfigure{
\begin{minipage}[t]{0.1\linewidth}
\centering
\includegraphics[width=2cm, height=2cm]{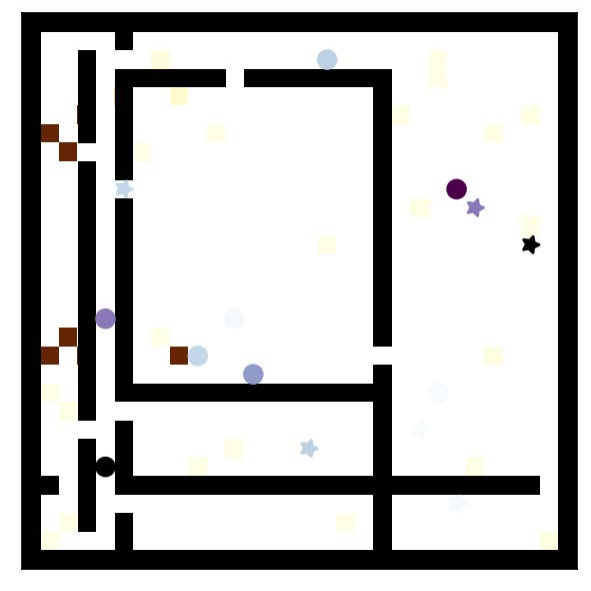}
\end{minipage}
}
\subfigure{
\begin{minipage}[t]{0.1\linewidth}
\centering
\includegraphics[width=2cm, height=2cm]{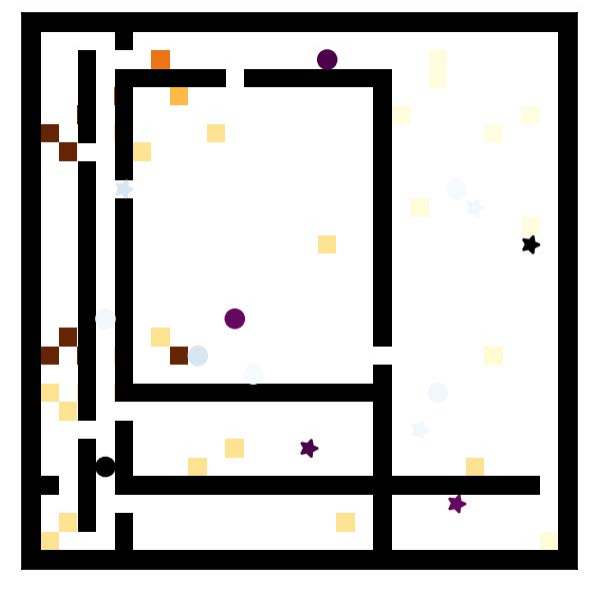}
\end{minipage}
}
\subfigure{
\begin{minipage}[t]{0.1\linewidth}
\centering
\includegraphics[width=2cm, height=2cm]{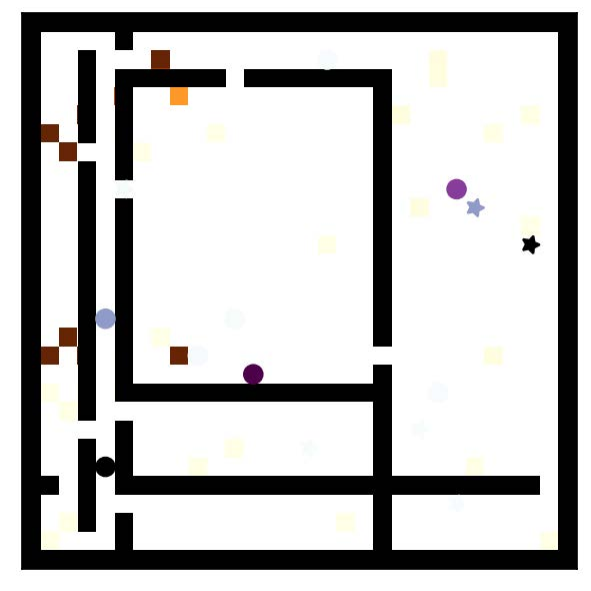}
\end{minipage}
}
\subfigure{
\begin{minipage}[t]{0.11\linewidth}
\centering
\includegraphics[width=2cm, height=2cm]{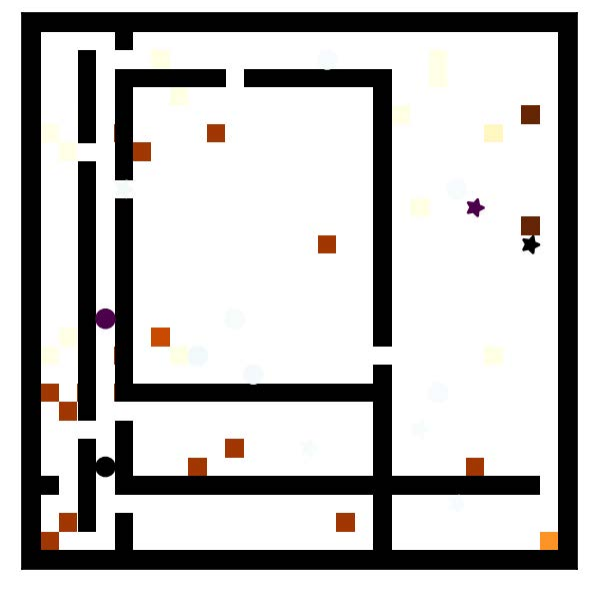}
\end{minipage}
}

\quad
\vspace{-0.5cm}

\subfigure{
\begin{minipage}[t]{0.1\linewidth}
\centering
\includegraphics[width=2cm, height=2cm]{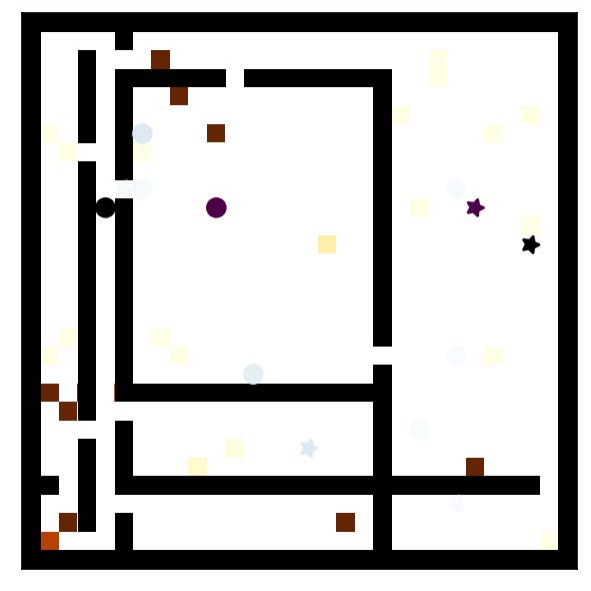}
\end{minipage}
}
\subfigure{
\begin{minipage}[t]{0.1\linewidth}
\centering
\includegraphics[width=2cm, height=2cm]{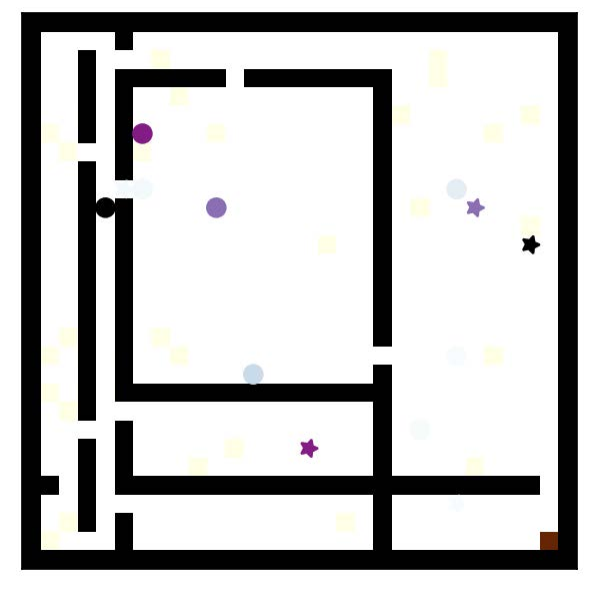}
\end{minipage}
}
\subfigure{
\begin{minipage}[t]{0.1\linewidth}
\centering
\includegraphics[width=2cm, height=2cm]{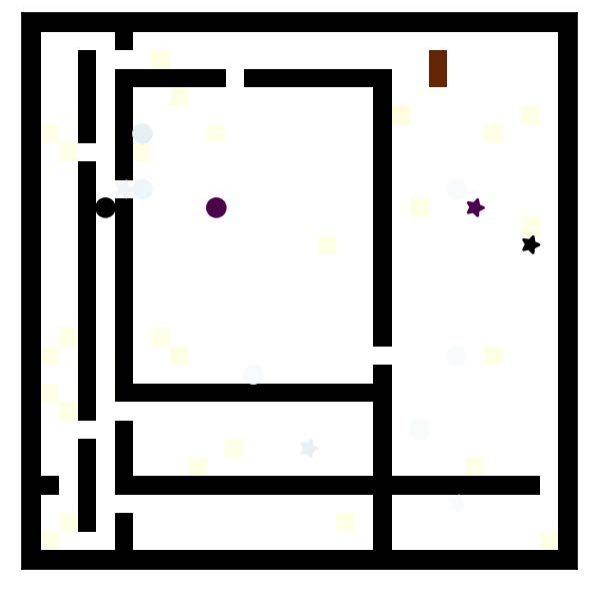}
\end{minipage}
}
\subfigure{
\begin{minipage}[t]{0.1\linewidth}
\centering
\includegraphics[width=2cm, height=2cm]{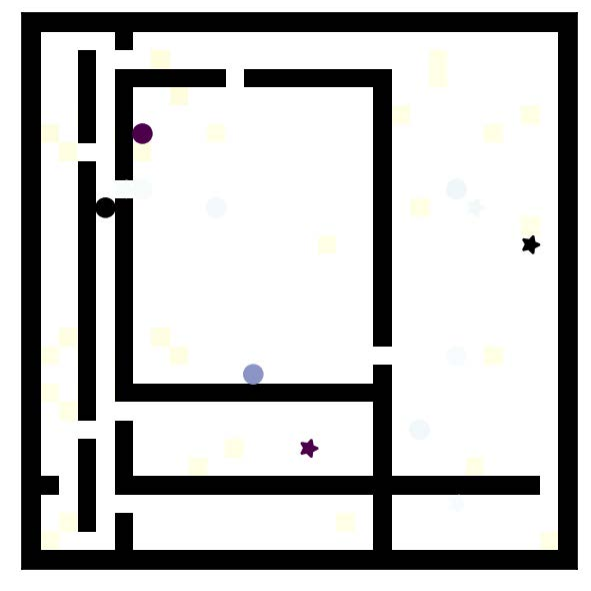}
\end{minipage}
}
\subfigure{
\begin{minipage}[t]{0.1\linewidth}
\centering
\includegraphics[width=2cm, height=2cm]{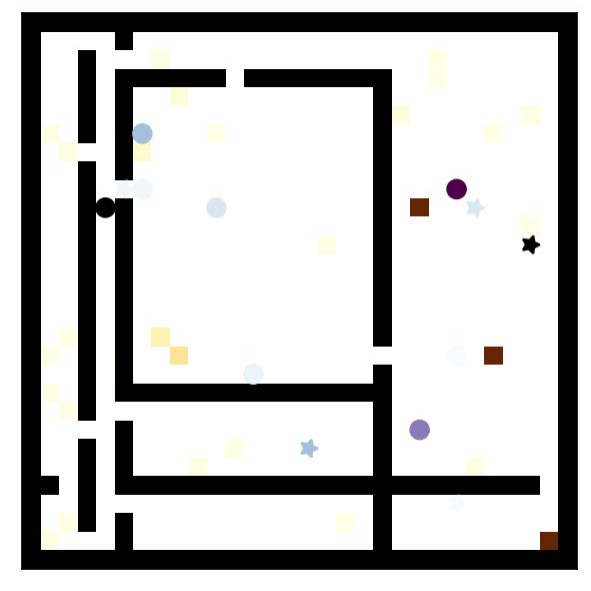}
\end{minipage}
}
\subfigure{
\begin{minipage}[t]{0.1\linewidth}
\centering
\includegraphics[width=2cm, height=2cm]{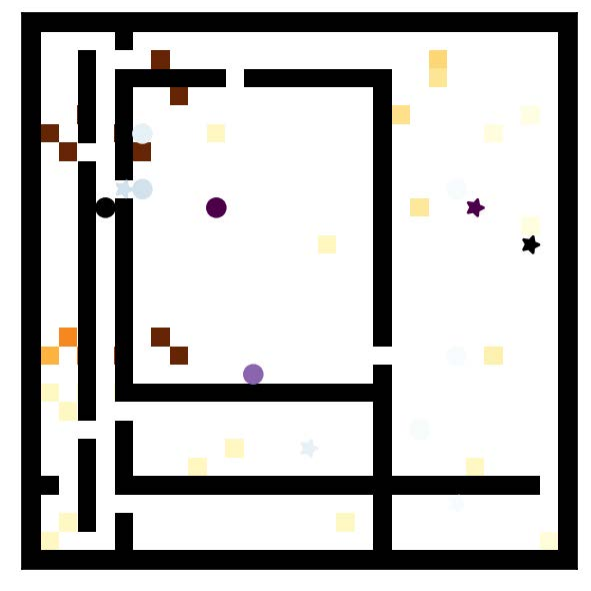}
\end{minipage}
}
\subfigure{
\begin{minipage}[t]{0.1\linewidth}
\centering
\includegraphics[width=2cm, height=2cm]{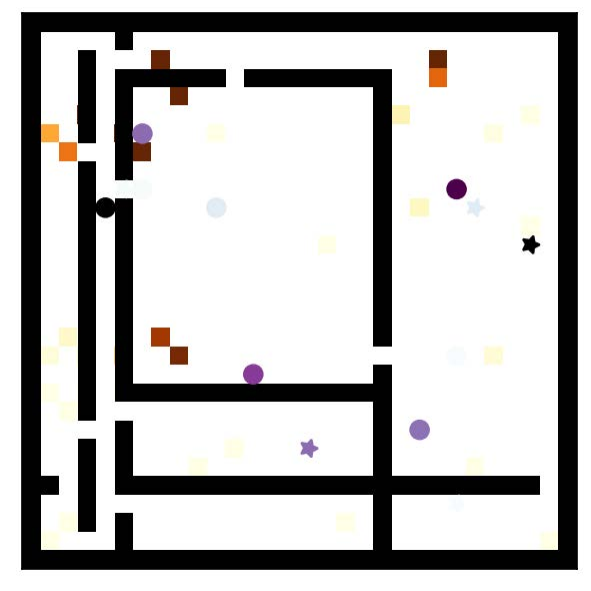}
\end{minipage}
}
\subfigure{
\begin{minipage}[t]{0.11\linewidth}
\centering
\includegraphics[width=2cm, height=2cm]{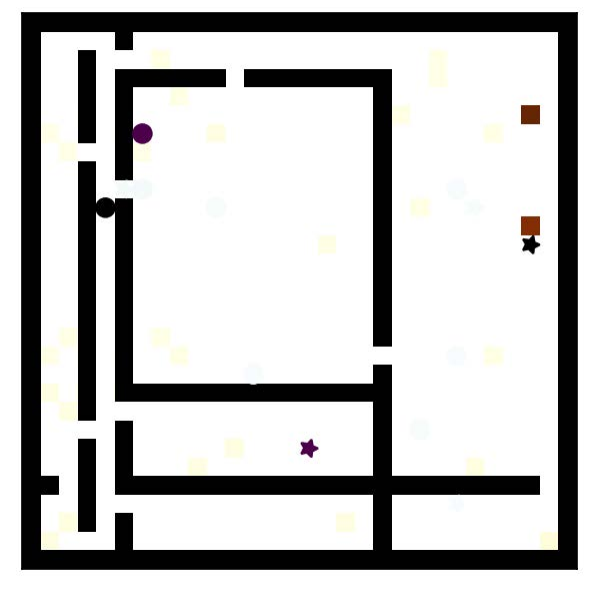}
\end{minipage}
}

\quad
\vspace{-0.5cm}

\subfigure{
\begin{minipage}[t]{0.1\linewidth}
\centering
\includegraphics[width=2cm, height=2cm]{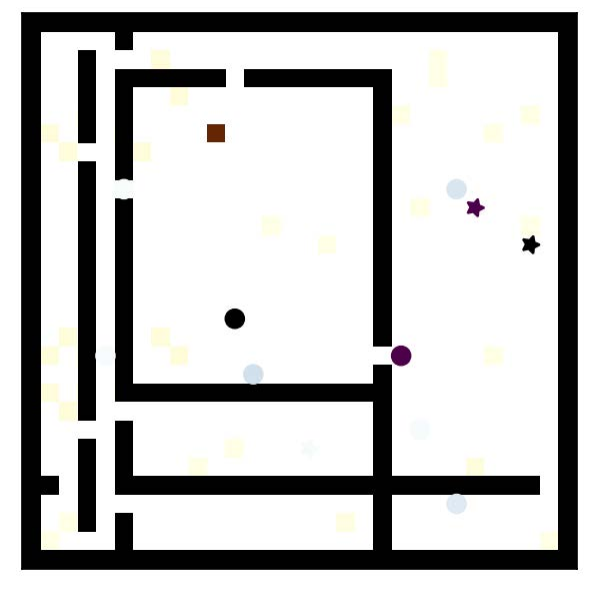}
\end{minipage}
}
\subfigure{
\begin{minipage}[t]{0.1\linewidth}
\centering
\includegraphics[width=2cm, height=2cm]{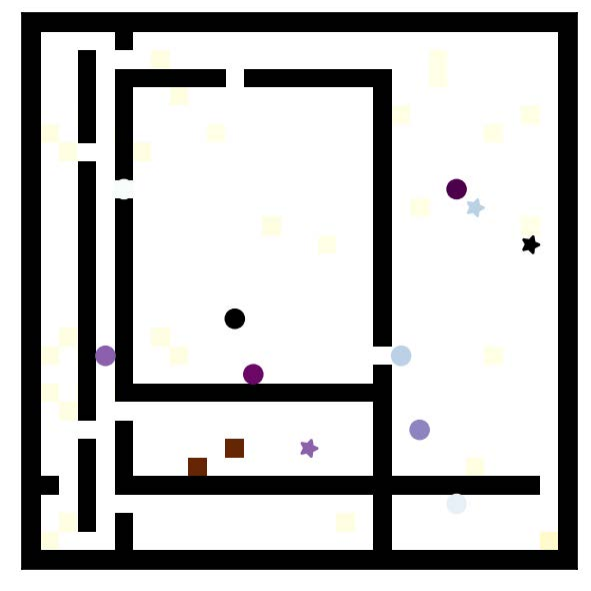}
\end{minipage}
}
\subfigure{
\begin{minipage}[t]{0.1\linewidth}
\centering
\includegraphics[width=2cm, height=2cm]{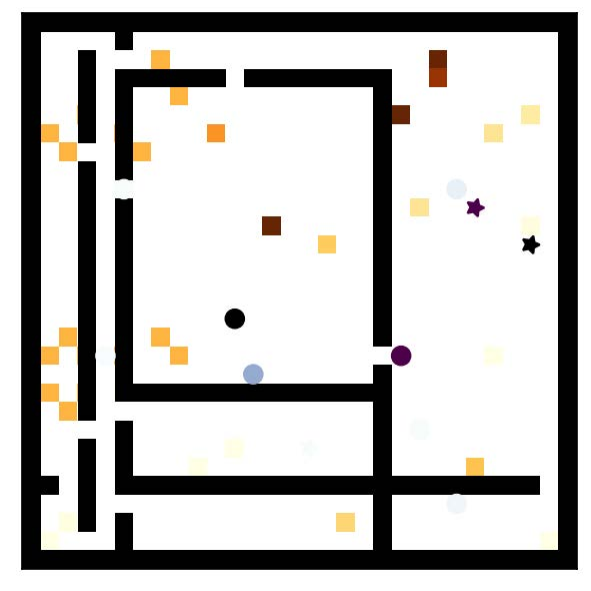}
\end{minipage}
}
\subfigure{
\begin{minipage}[t]{0.1\linewidth}
\centering
\includegraphics[width=2cm, height=2cm]{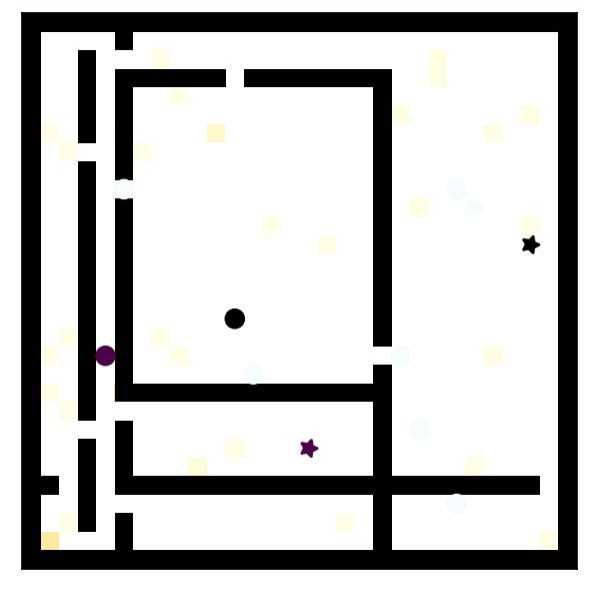}
\end{minipage}
}
\subfigure{
\begin{minipage}[t]{0.1\linewidth}
\centering
\includegraphics[width=2cm, height=2cm]{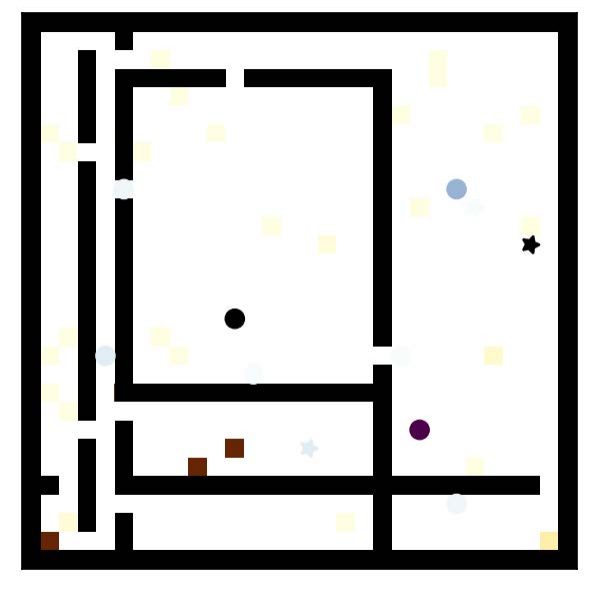}
\end{minipage}
}
\subfigure{
\begin{minipage}[t]{0.1\linewidth}
\centering
\includegraphics[width=2cm, height=2cm]{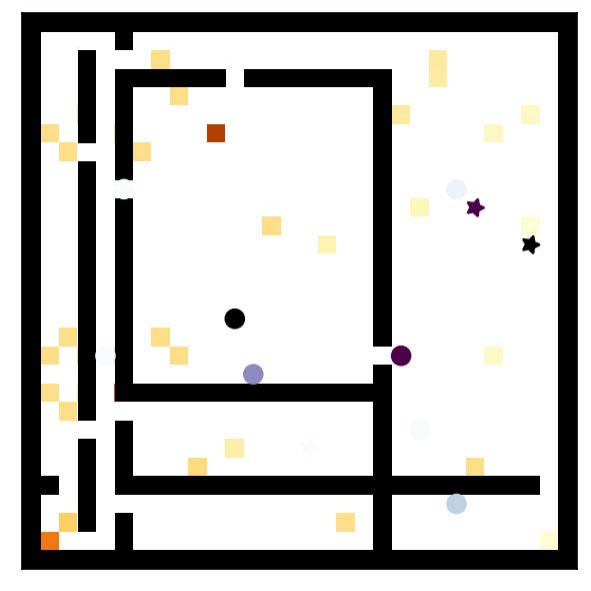}
\end{minipage}
}
\subfigure{
\begin{minipage}[t]{0.1\linewidth}
\centering
\includegraphics[width=2cm, height=2cm]{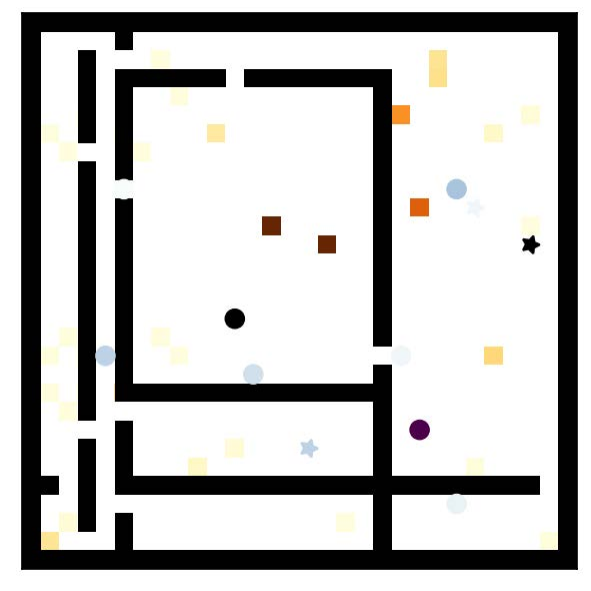}
\end{minipage}
}
\subfigure{
\begin{minipage}[t]{0.11\linewidth}
\centering
\includegraphics[width=2cm, height=2cm]{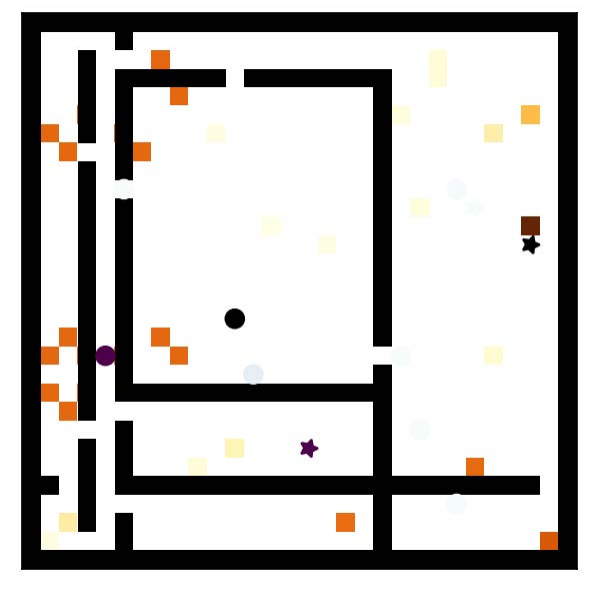}
\end{minipage}
}

\quad
\vspace{-0.5cm}

\setcounter{subfigure}{0}
\subfigure[head 1]{
\begin{minipage}[t]{0.1\linewidth}
\centering
\includegraphics[width=2cm, height=2cm]{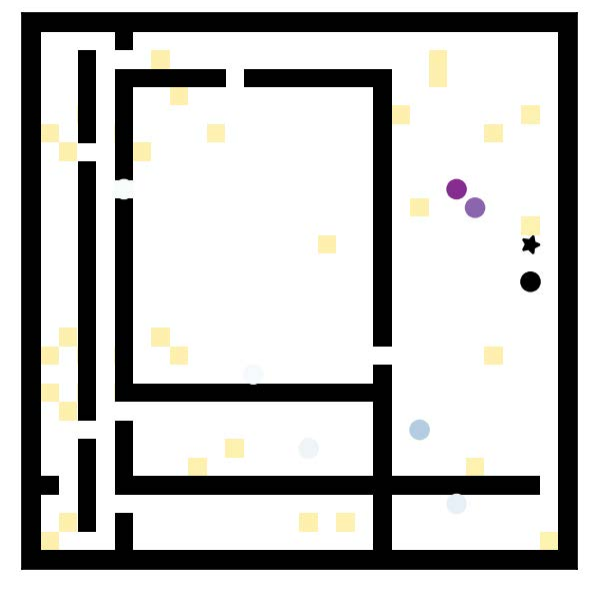}
\end{minipage}
}
\subfigure[head 2]{
\begin{minipage}[t]{0.1\linewidth}
\centering
\includegraphics[width=2cm, height=2cm]{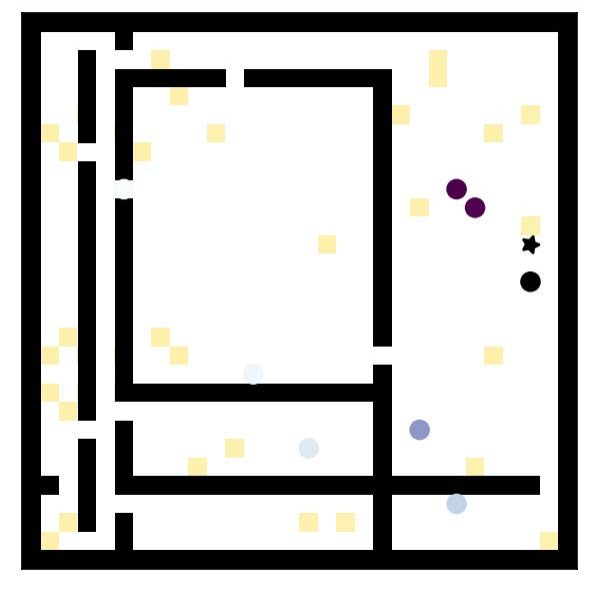}
\end{minipage}
}
\subfigure[head 3]{
\begin{minipage}[t]{0.1\linewidth}
\centering
\includegraphics[width=2cm, height=2cm]{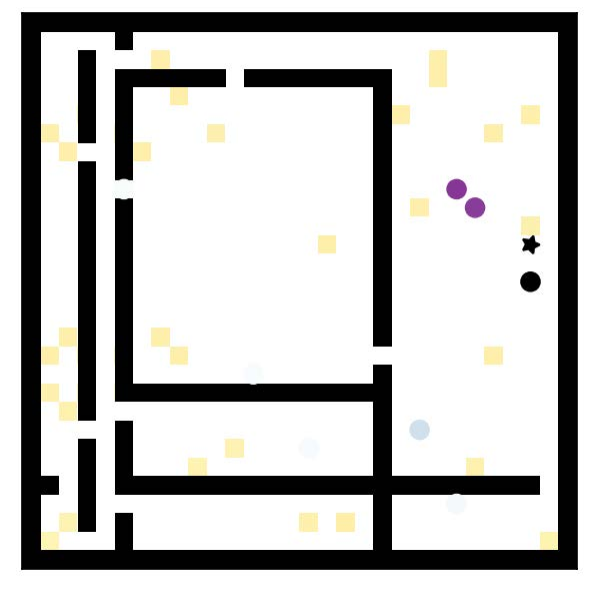}
\end{minipage}
}
\subfigure[head 4]{
\begin{minipage}[t]{0.1\linewidth}
\centering
\includegraphics[width=2cm, height=2cm]{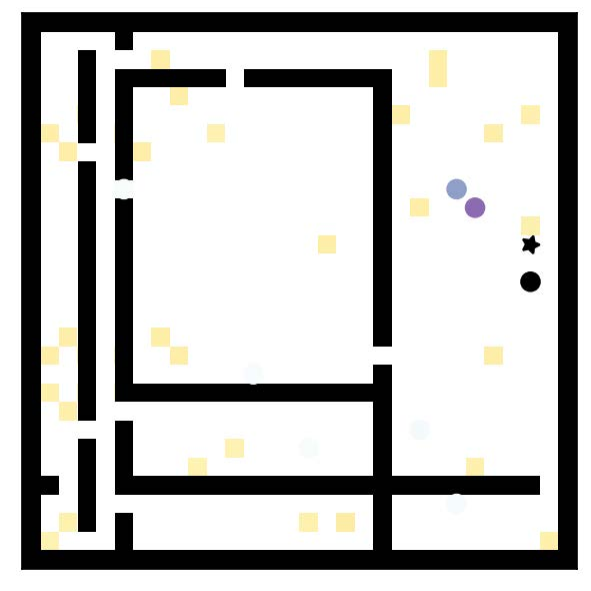}
\end{minipage}
}
\subfigure[head 5]{
\begin{minipage}[t]{0.1\linewidth}
\centering
\includegraphics[width=2cm, height=2cm]{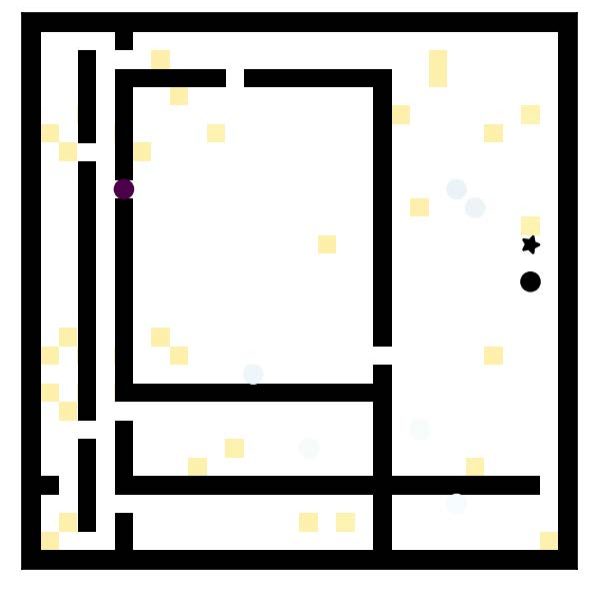}
\end{minipage}
}
\subfigure[head 6]{
\begin{minipage}[t]{0.1\linewidth}
\centering
\includegraphics[width=2cm, height=2cm]{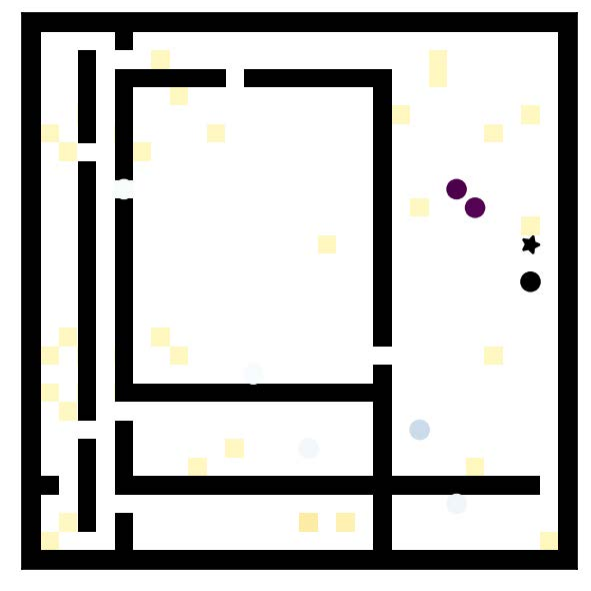}
\end{minipage}
}
\subfigure[head 7]{
\begin{minipage}[t]{0.1\linewidth}
\centering
\includegraphics[width=2cm, height=2cm]{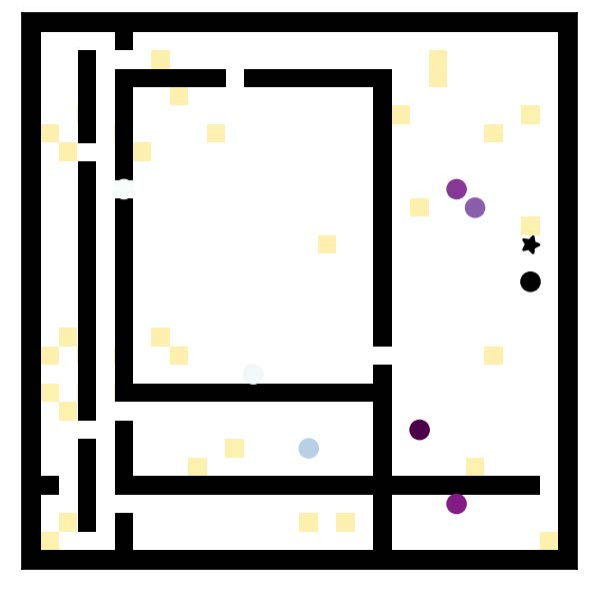}
\end{minipage}
}
\subfigure[head 8]{
\begin{minipage}[t]{0.11\linewidth}
\centering
\includegraphics[width=2cm, height=2cm]{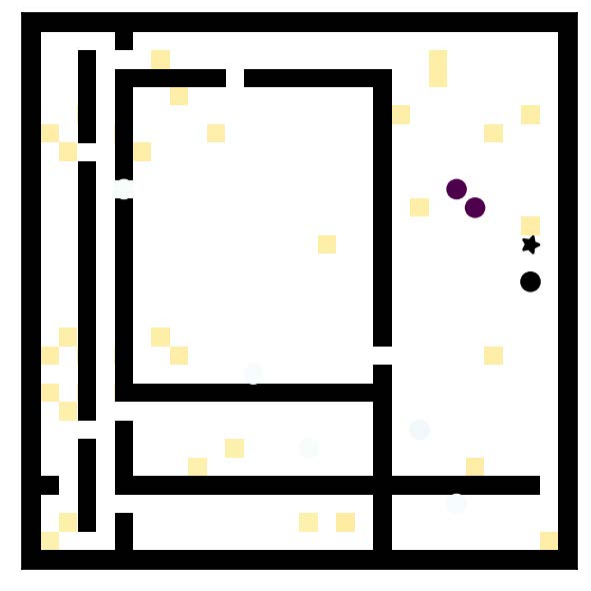}
\end{minipage}
}

\caption{\textbf{Visualization of Attention Mechanism}: Four distinct time steps are depicted, with each row representing the eight attention heads of the current agent towards nodes and other agents at the corresponding time step. Agents are represented by circles, goals by stars, and the current agent is denoted by a black circle. Darker colors on the figure indicate higher attention weights.}
\label{Visualization of attention mechanism}

\end{figure*}

\begin{figure*}[h]
\centering
\subfigure{
\begin{minipage}[t]{0.1\linewidth}
\centering
\includegraphics[width=2cm, height=2cm]{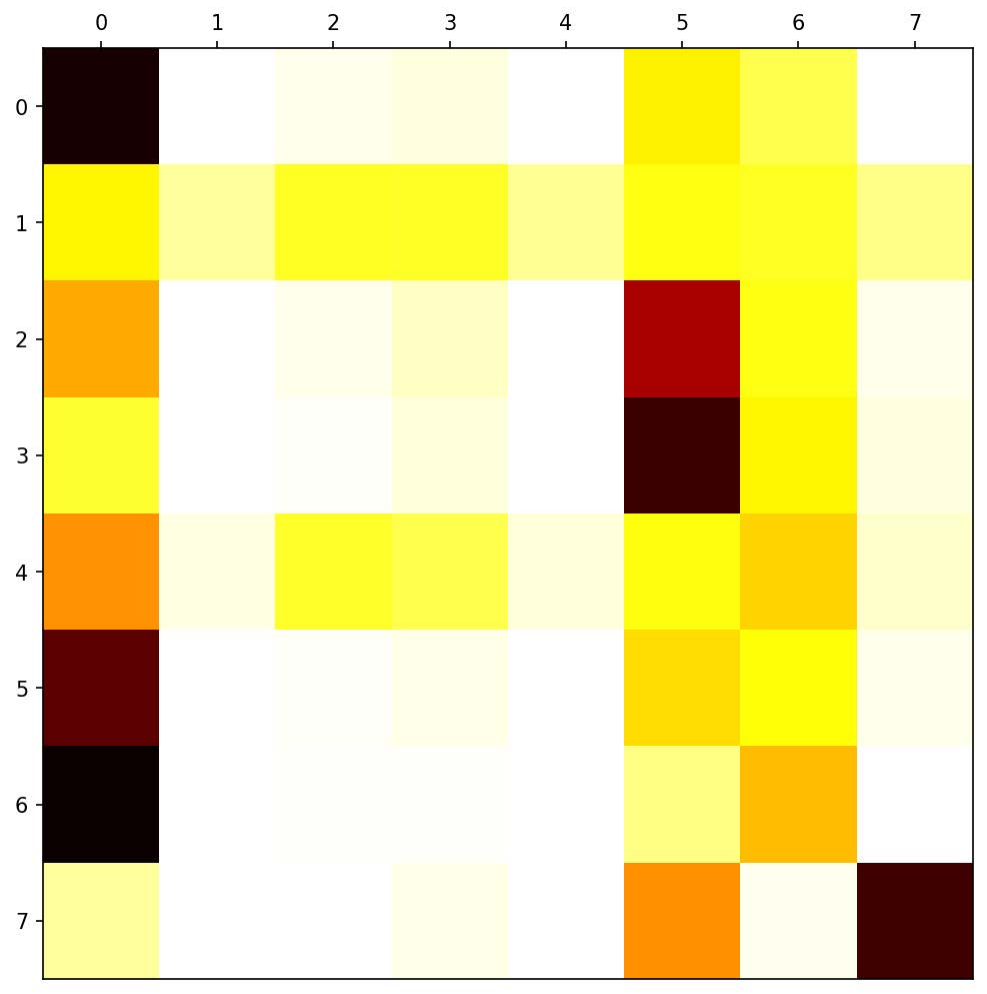}
\end{minipage}
}
\subfigure{
\begin{minipage}[t]{0.1\linewidth}
\centering
\includegraphics[width=2cm, height=2cm]{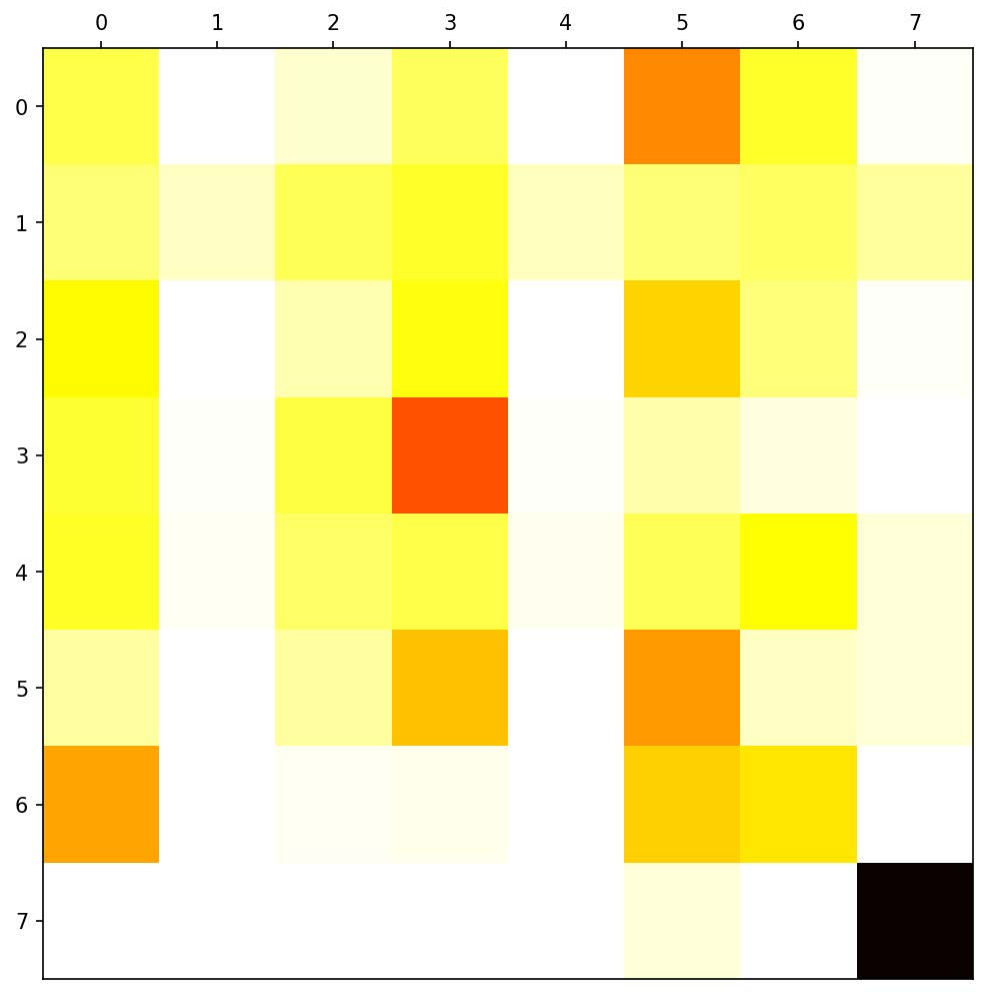}
\end{minipage}
}
\subfigure{
\begin{minipage}[t]{0.1\linewidth}
\centering
\includegraphics[width=2cm, height=2cm]{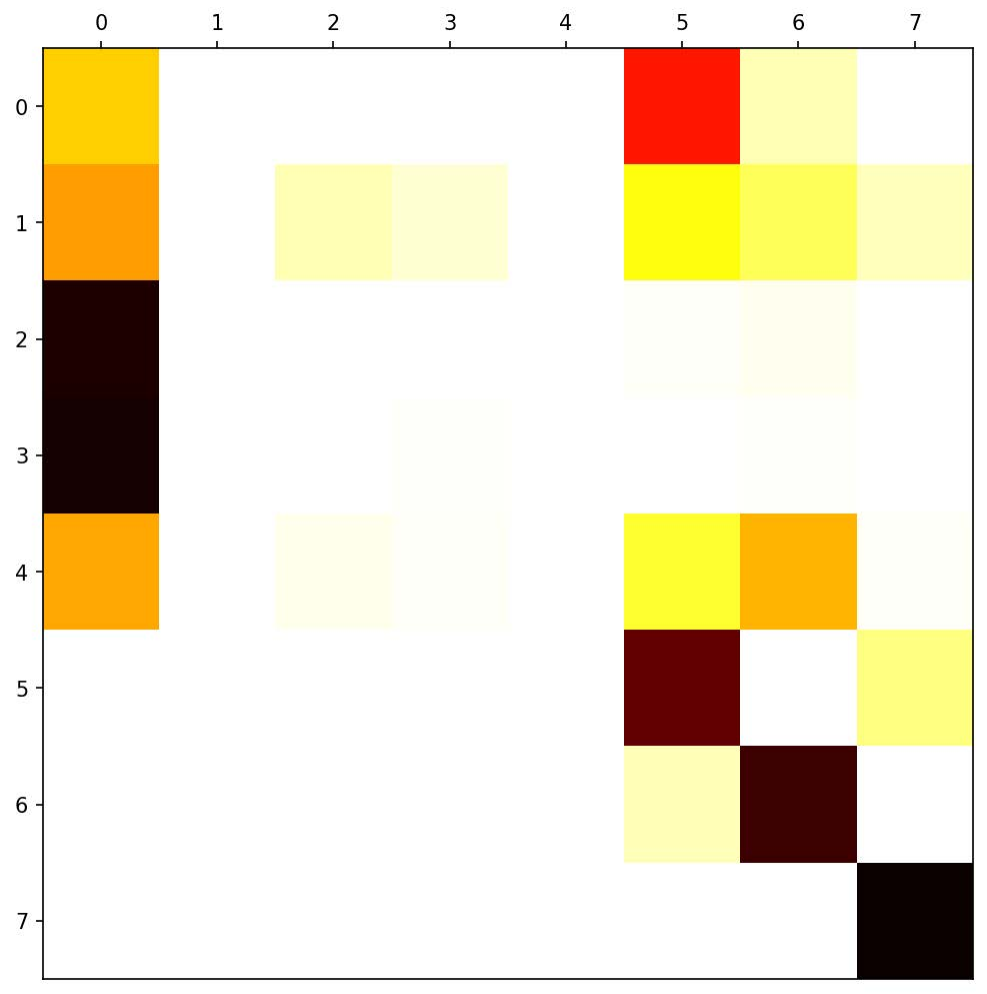}
\end{minipage}
}
\subfigure{
\begin{minipage}[t]{0.1\linewidth}
\centering
\includegraphics[width=2cm, height=2cm]{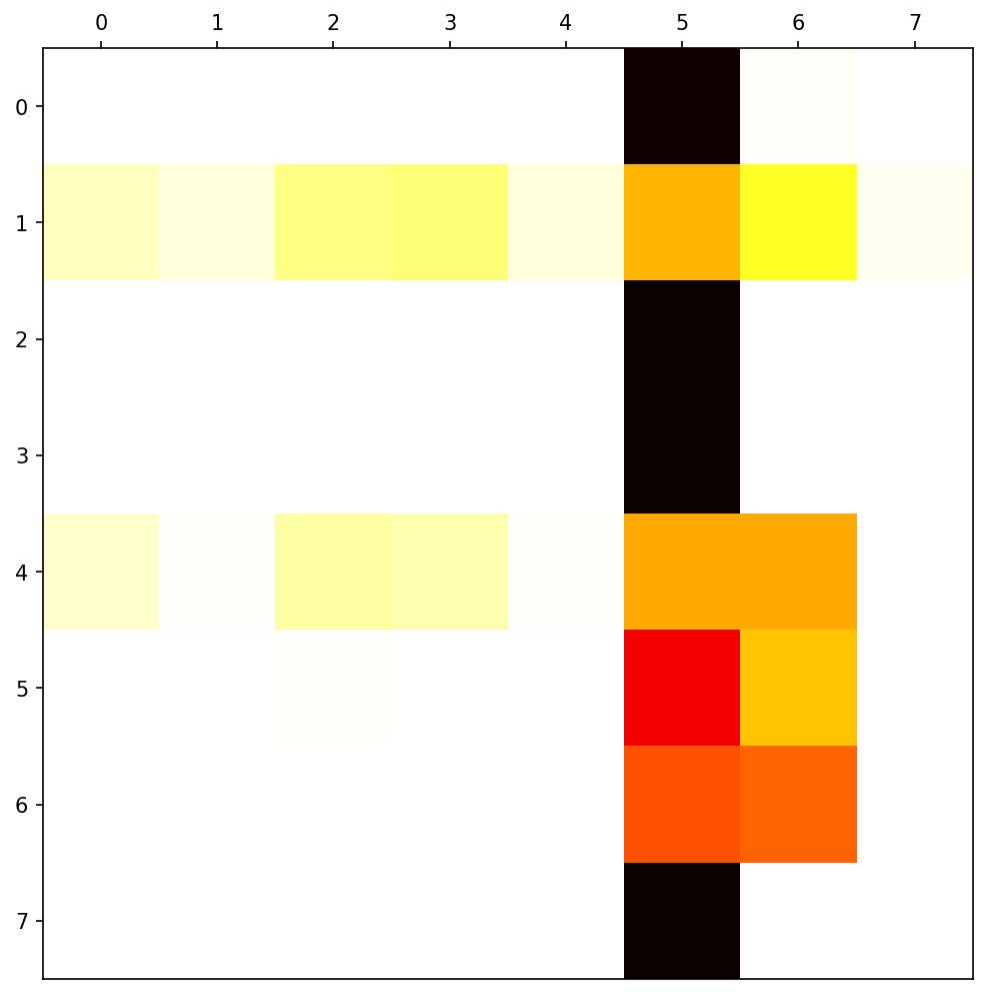}
\end{minipage}
}
\subfigure{
\begin{minipage}[t]{0.1\linewidth}
\centering
\includegraphics[width=2cm, height=2cm]{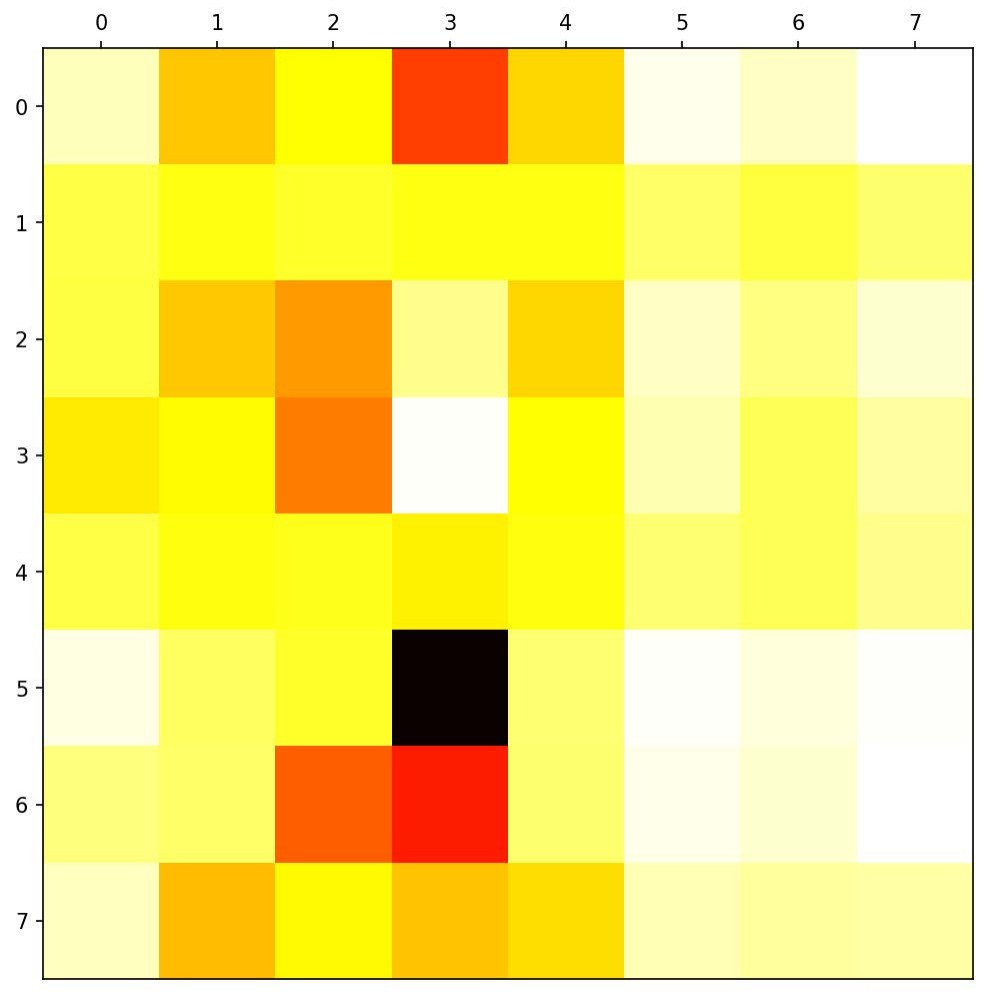}
\end{minipage}
}
\subfigure{
\begin{minipage}[t]{0.1\linewidth}
\centering
\includegraphics[width=2cm, height=2cm]{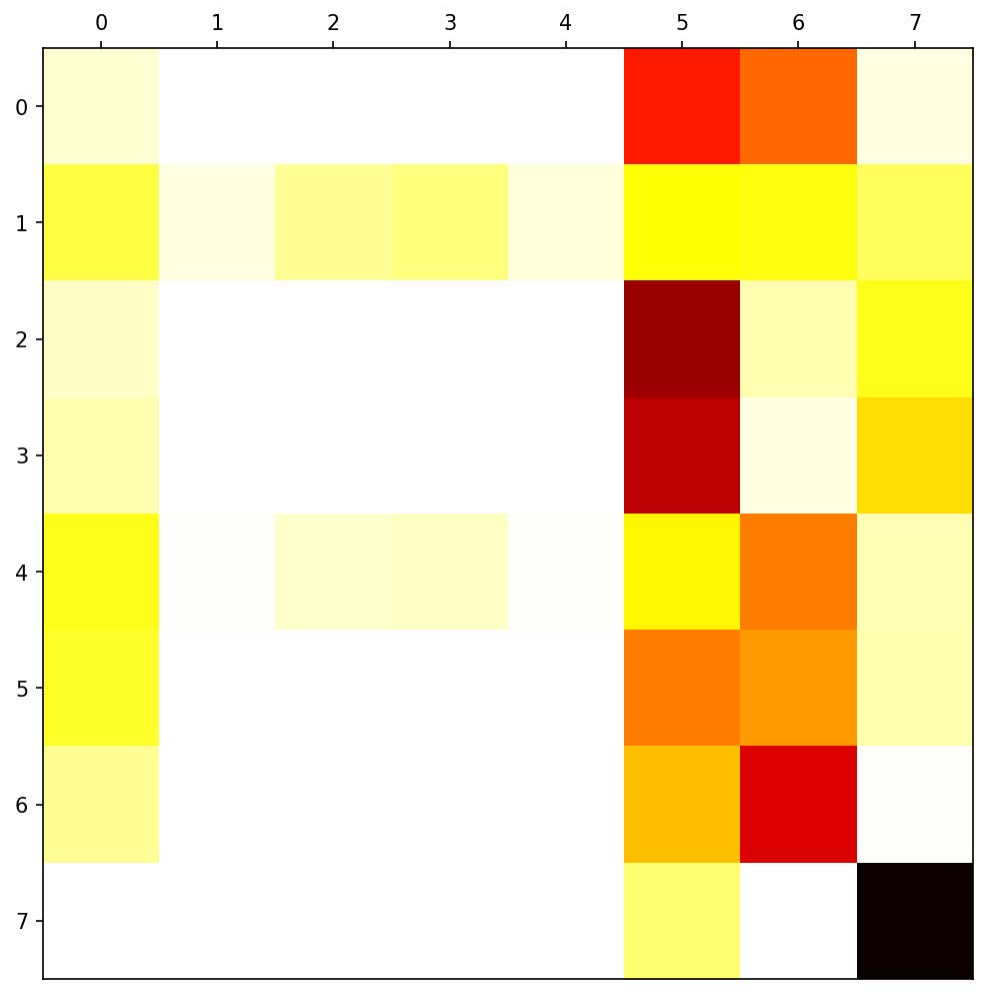}
\end{minipage}
}
\subfigure{
\begin{minipage}[t]{0.1\linewidth}
\centering
\includegraphics[width=2cm, height=2cm]{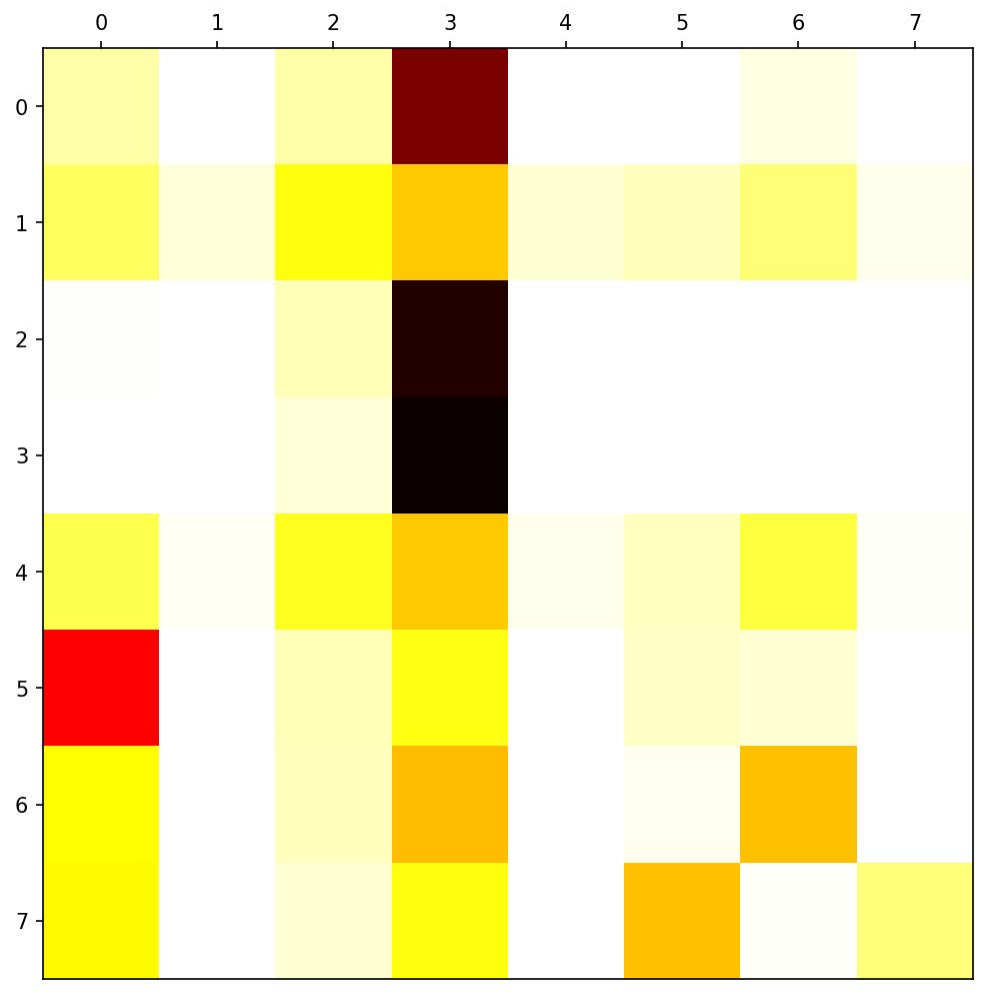}
\end{minipage}
}
\subfigure{
\begin{minipage}[t]{0.11\linewidth}
\centering
\includegraphics[width=2cm, height=2cm]{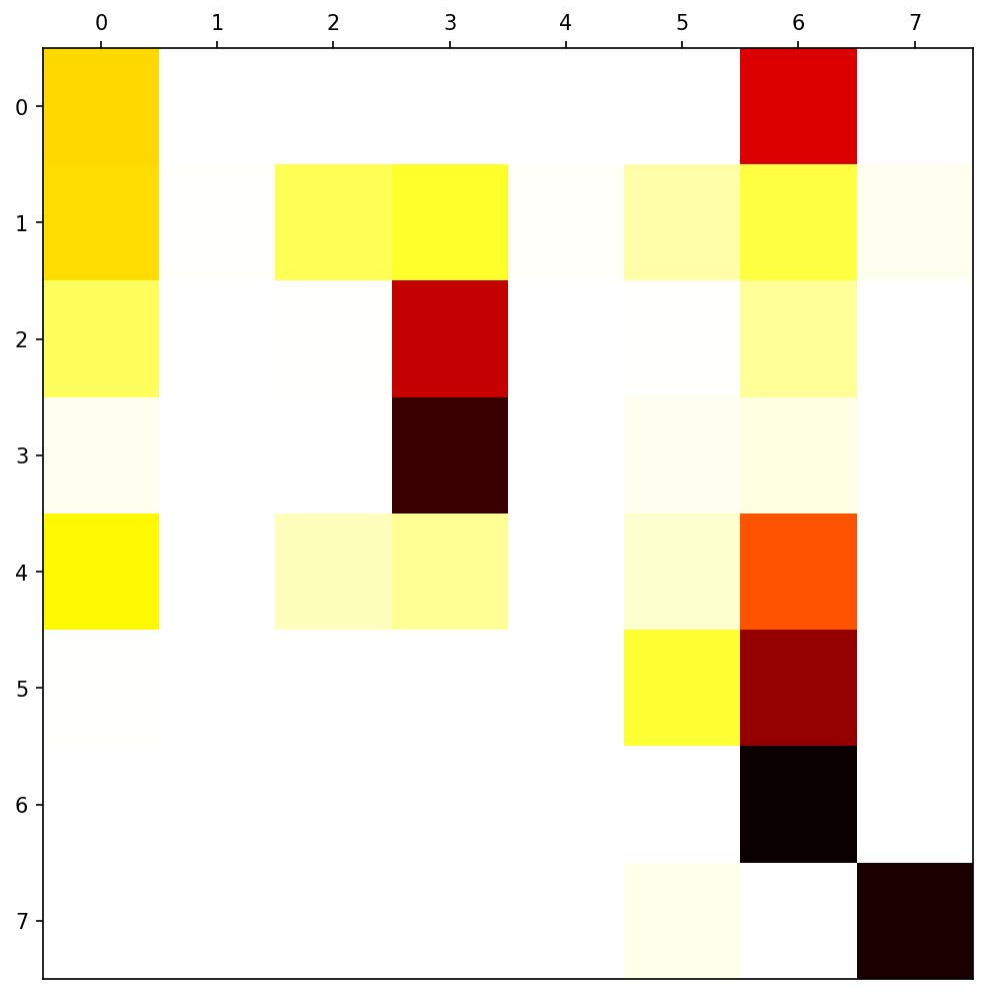}
\end{minipage}
}

\quad
\vspace{-0.5cm}

\subfigure{
\begin{minipage}[t]{0.1\linewidth}
\centering
\includegraphics[width=2cm, height=2cm]{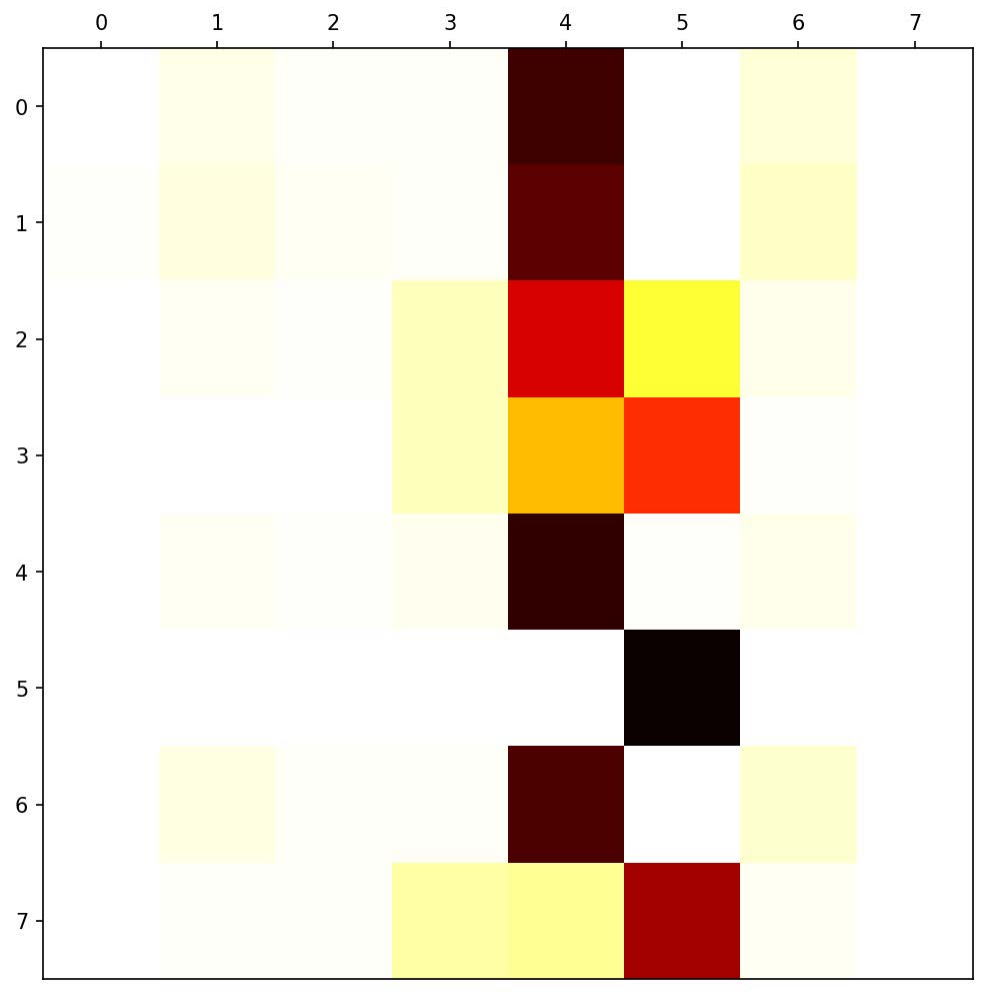}
\end{minipage}
}
\subfigure{
\begin{minipage}[t]{0.1\linewidth}
\centering
\includegraphics[width=2cm, height=2cm]{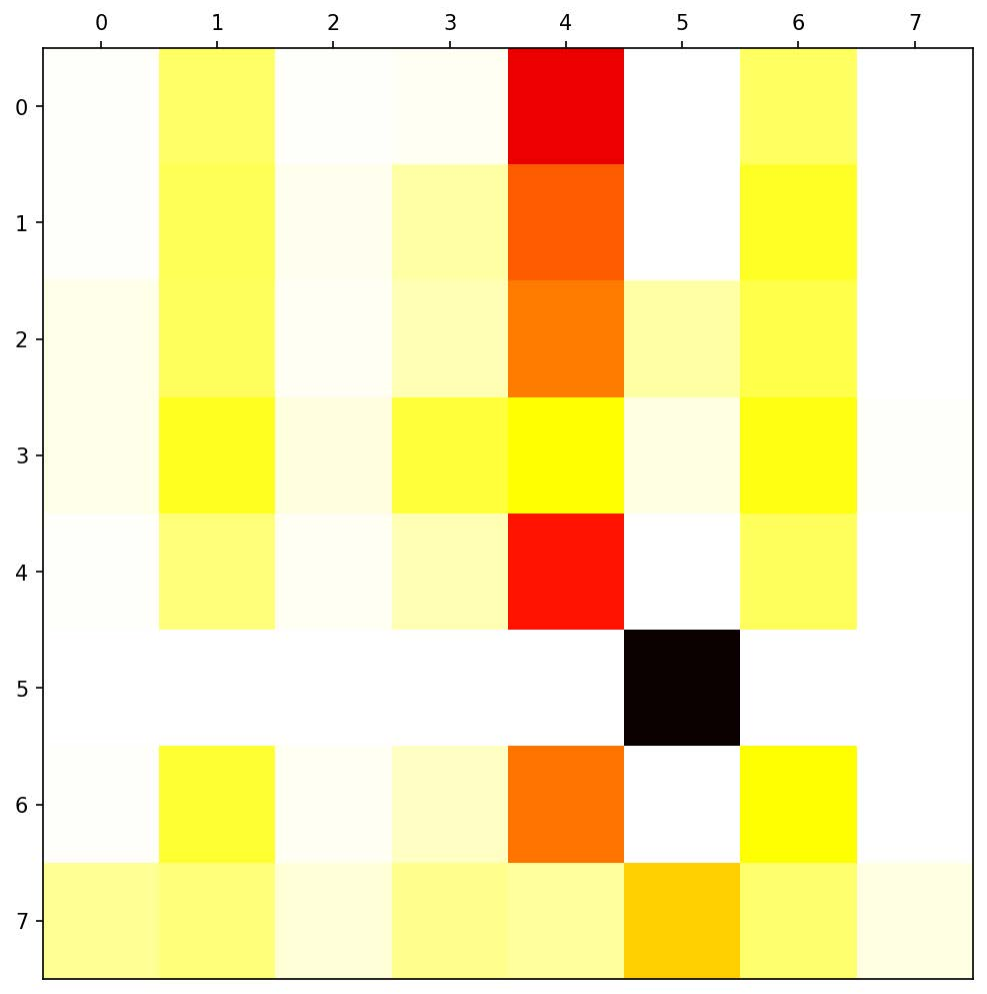}
\end{minipage}
}
\subfigure{
\begin{minipage}[t]{0.1\linewidth}
\centering
\includegraphics[width=2cm, height=2cm]{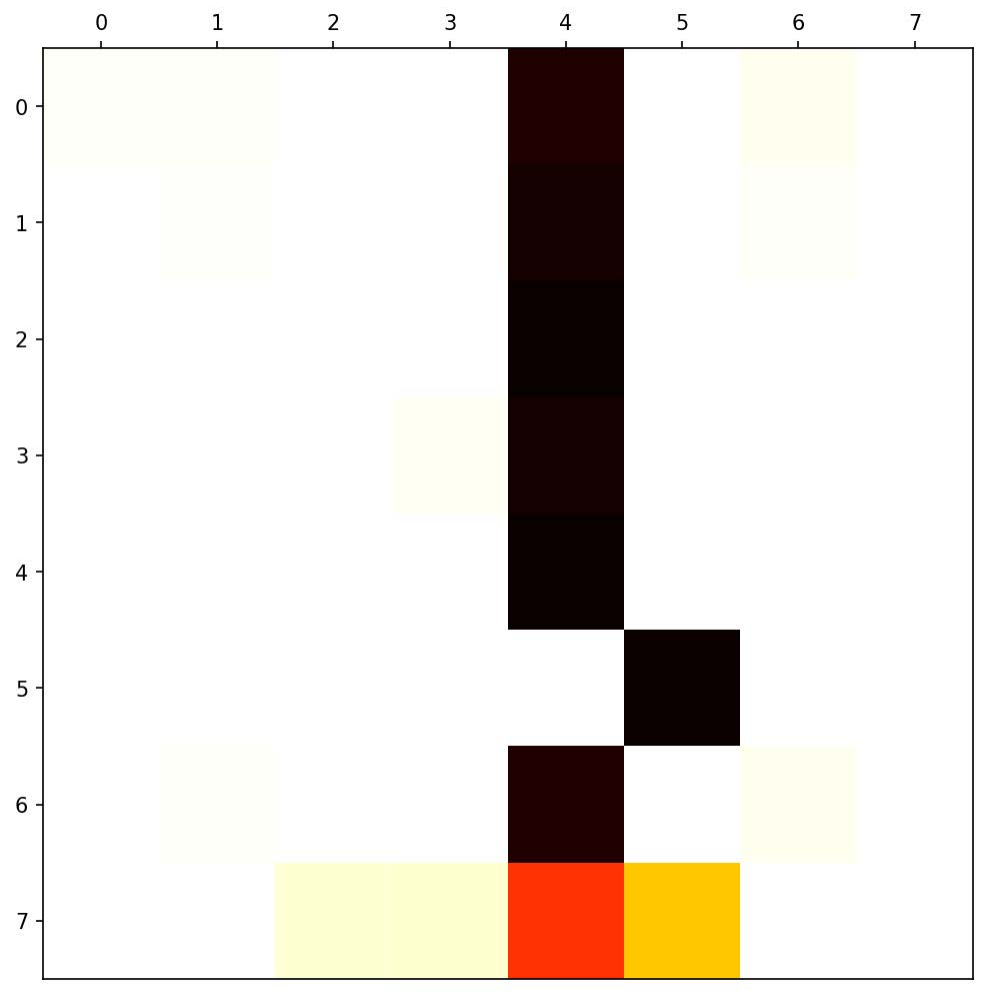}
\end{minipage}
}
\subfigure{
\begin{minipage}[t]{0.1\linewidth}
\centering
\includegraphics[width=2cm, height=2cm]{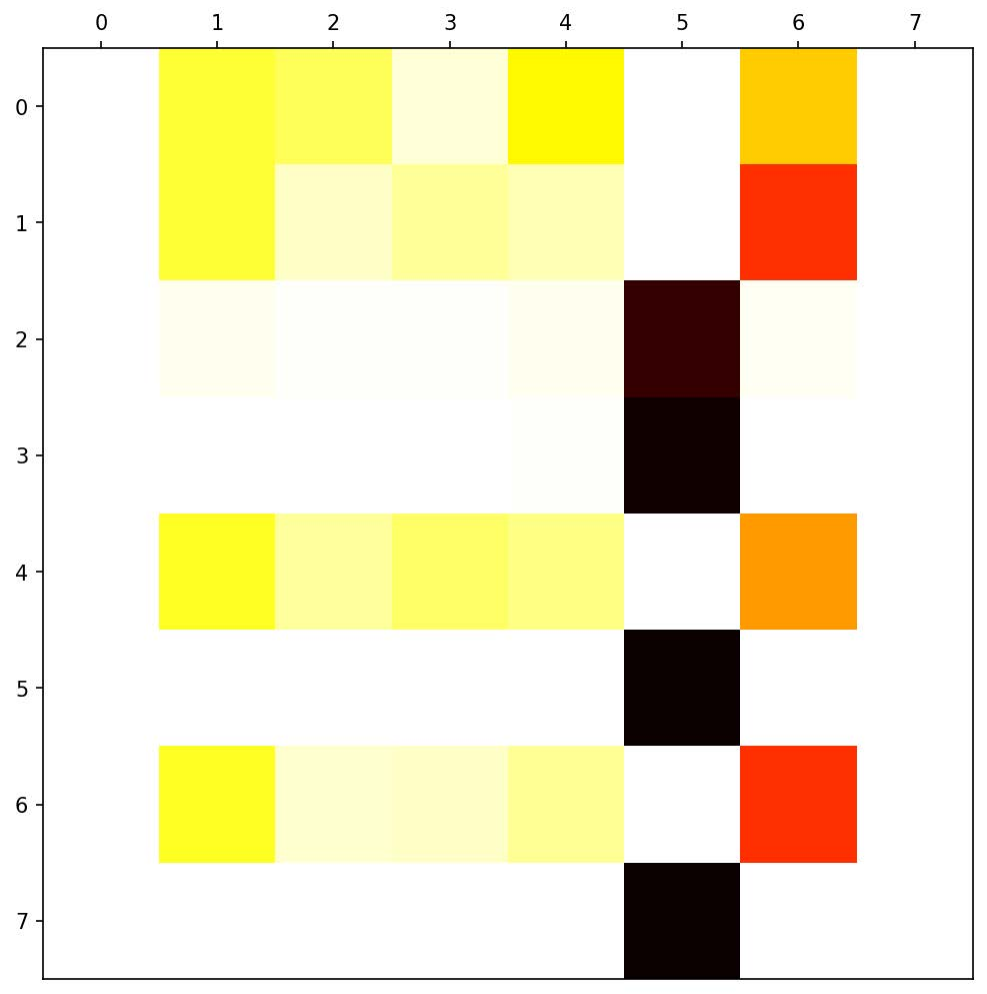}
\end{minipage}
}
\subfigure{
\begin{minipage}[t]{0.1\linewidth}
\centering
\includegraphics[width=2cm, height=2cm]{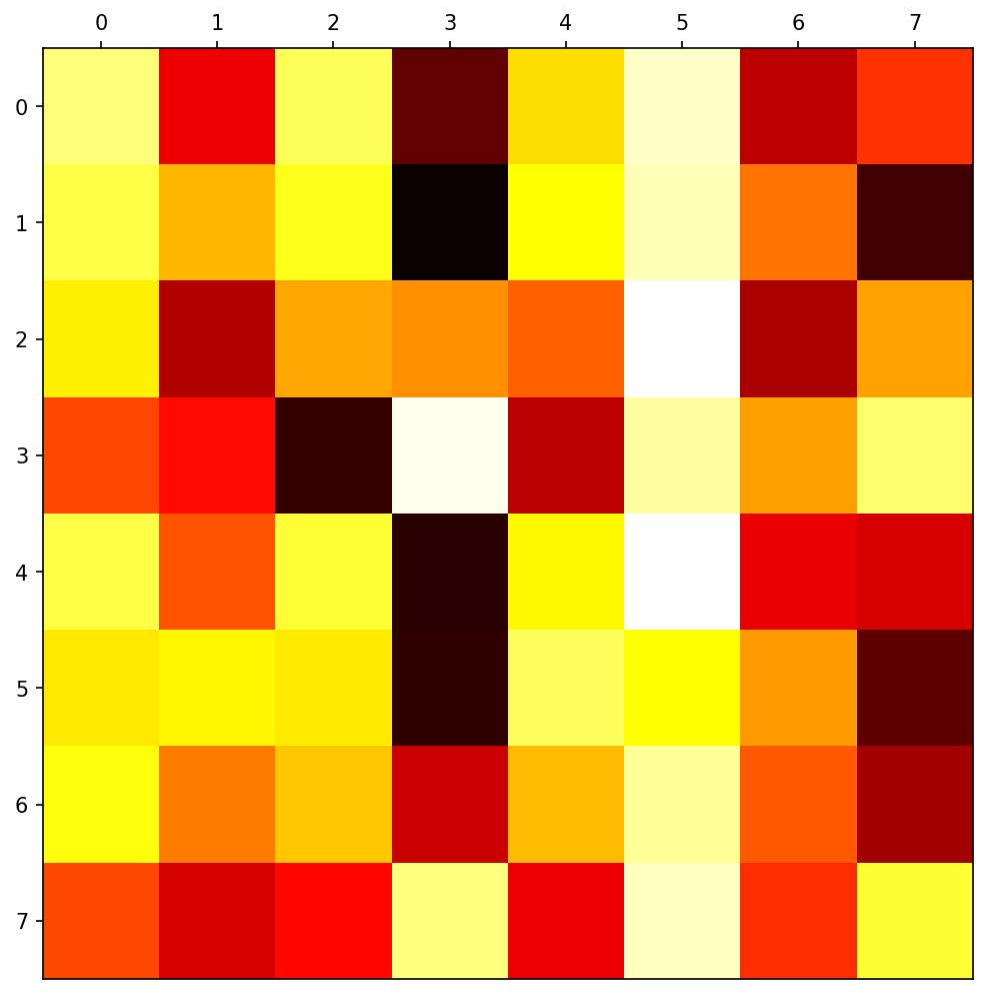}
\end{minipage}
}
\subfigure{
\begin{minipage}[t]{0.1\linewidth}
\centering
\includegraphics[width=2cm, height=2cm]{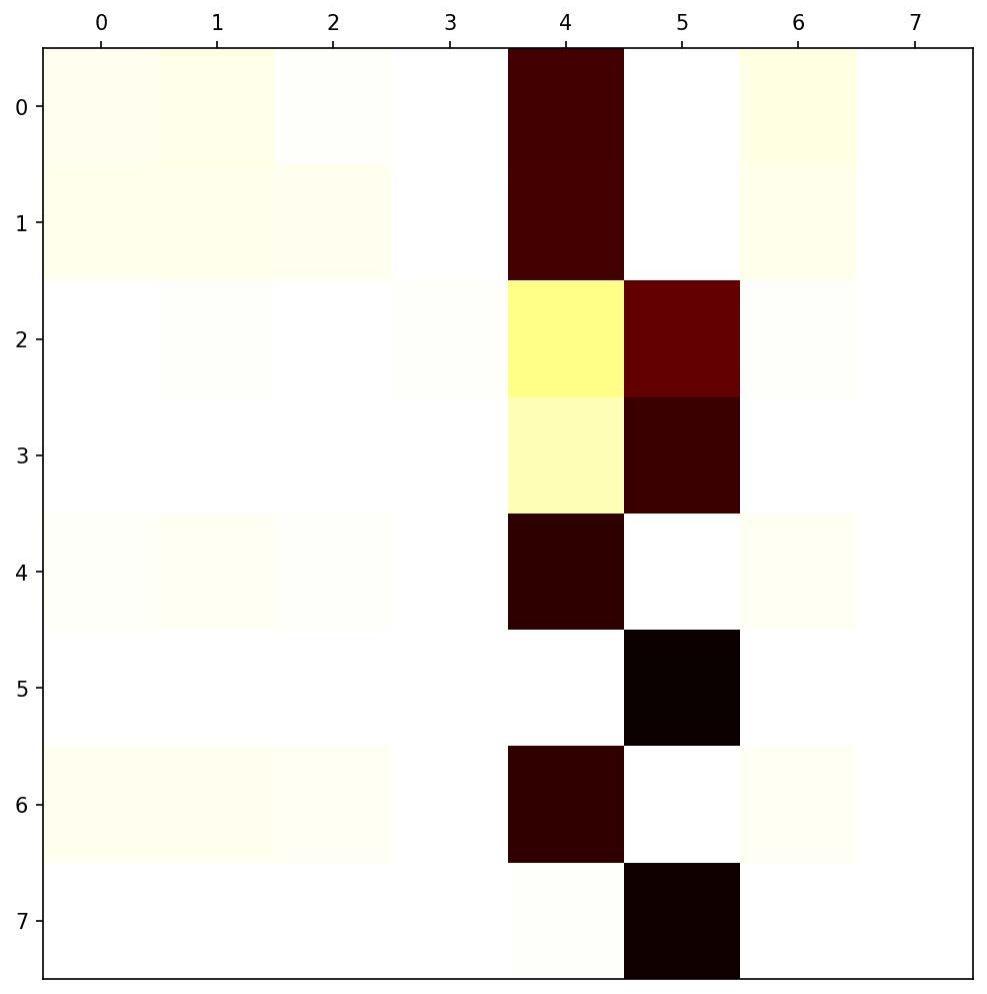}
\end{minipage}
}
\subfigure{
\begin{minipage}[t]{0.1\linewidth}
\centering
\includegraphics[width=2cm, height=2cm]{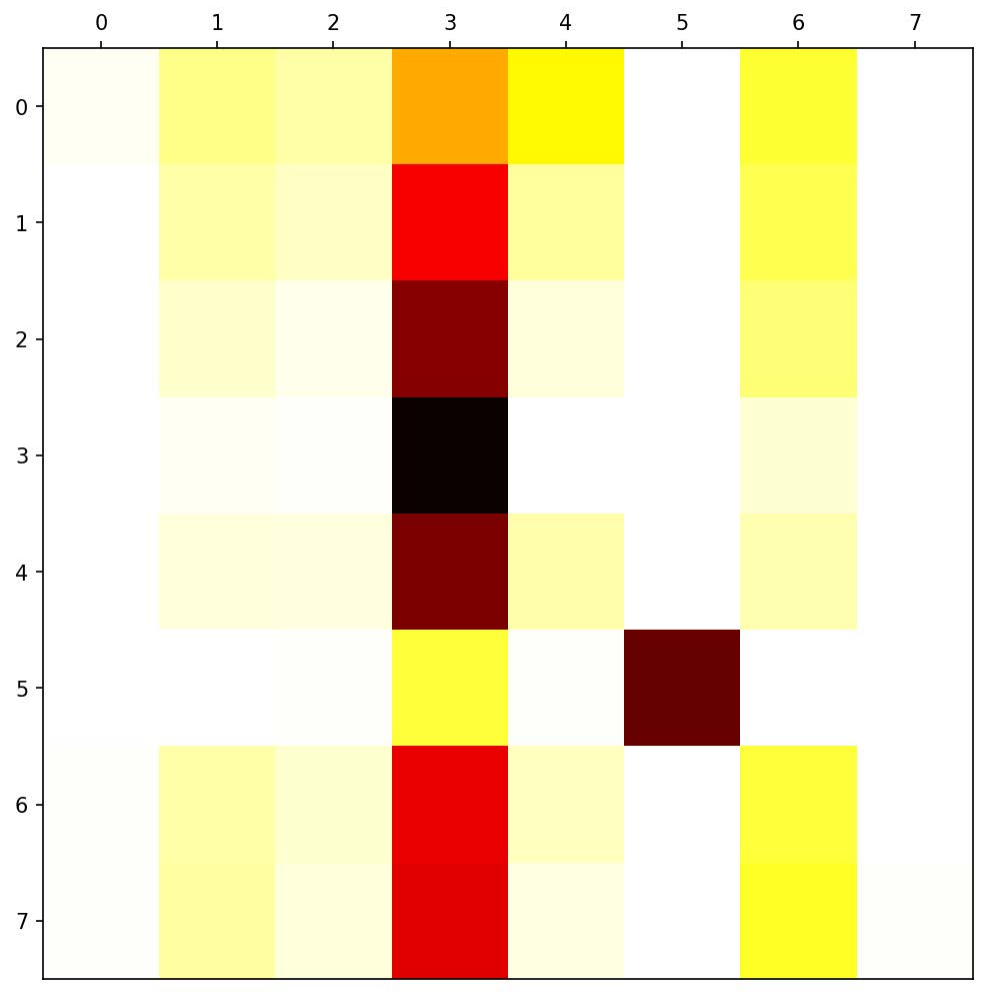}
\end{minipage}
}
\subfigure{
\begin{minipage}[t]{0.11\linewidth}
\centering
\includegraphics[width=2cm, height=2cm]{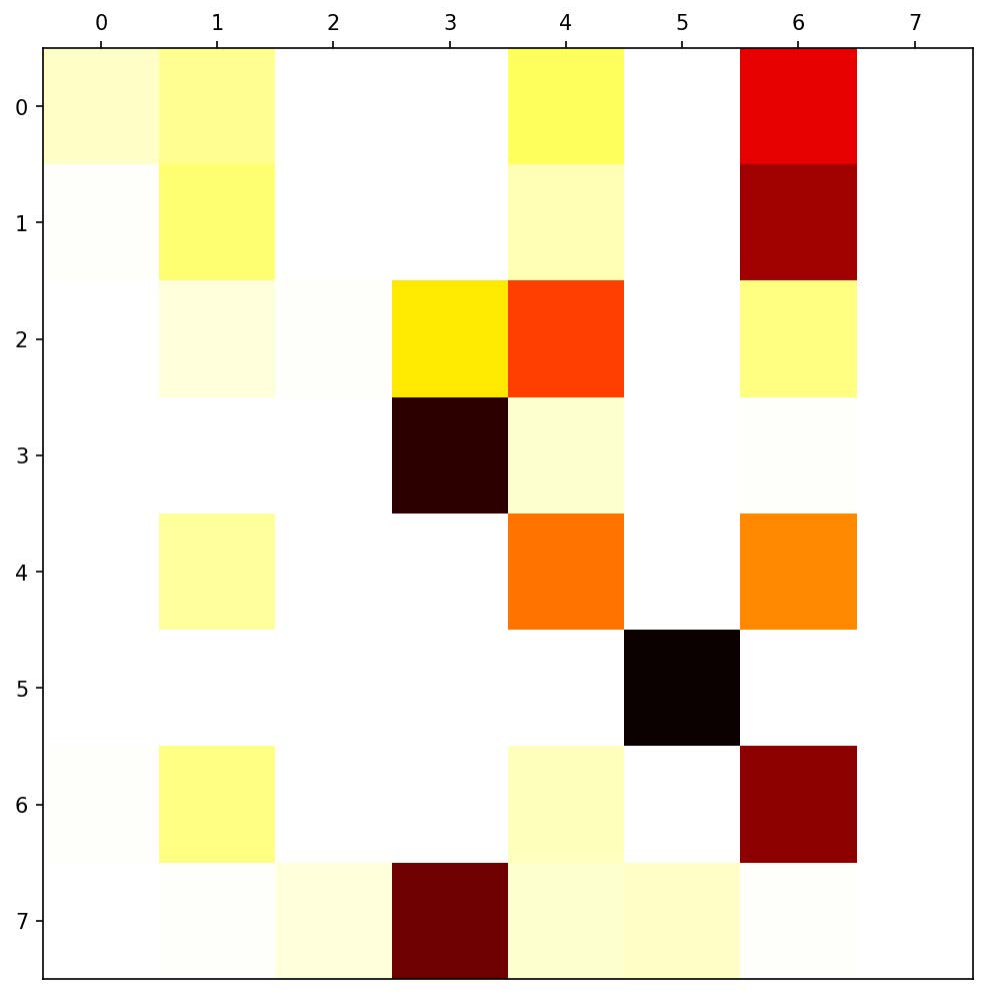}
\end{minipage}
}

\quad
\vspace{-0.5cm}

\subfigure{
\begin{minipage}[t]{0.1\linewidth}
\centering
\includegraphics[width=2cm, height=2cm]{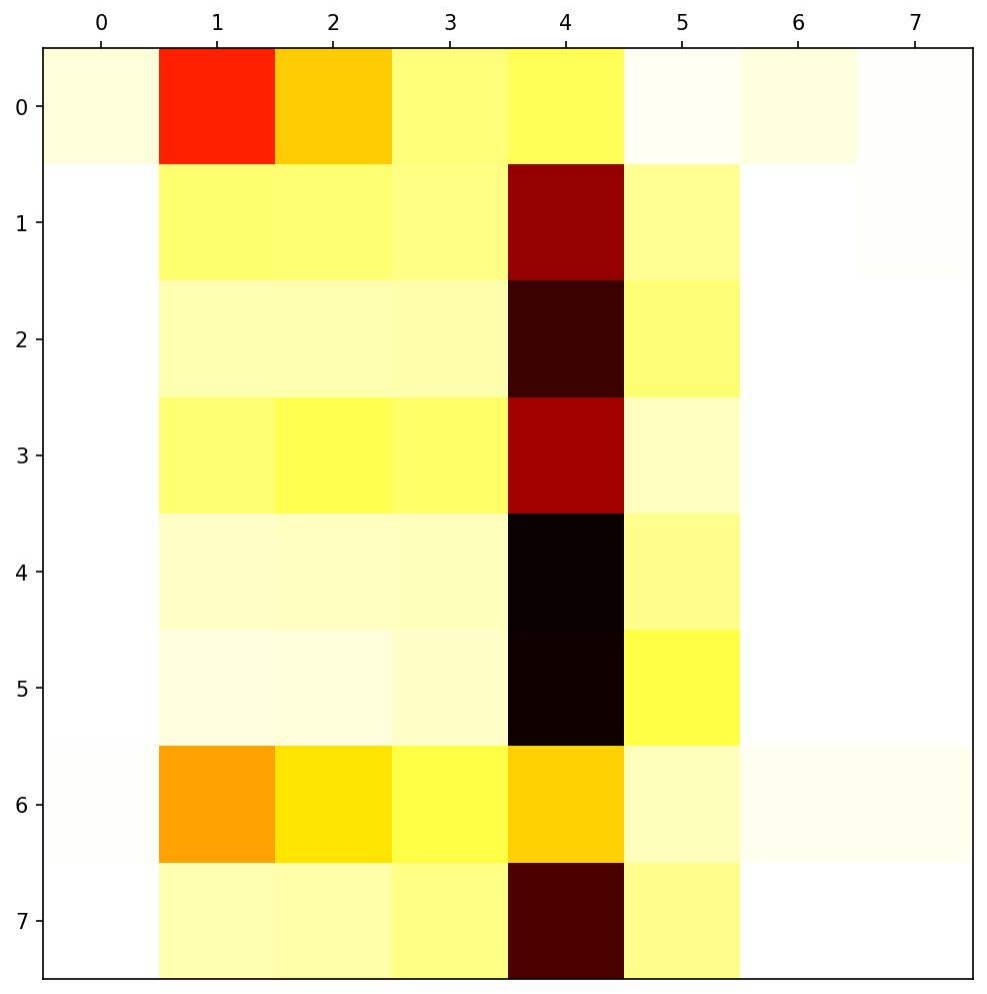}
\end{minipage}
}
\subfigure{
\begin{minipage}[t]{0.1\linewidth}
\centering
\includegraphics[width=2cm, height=2cm]{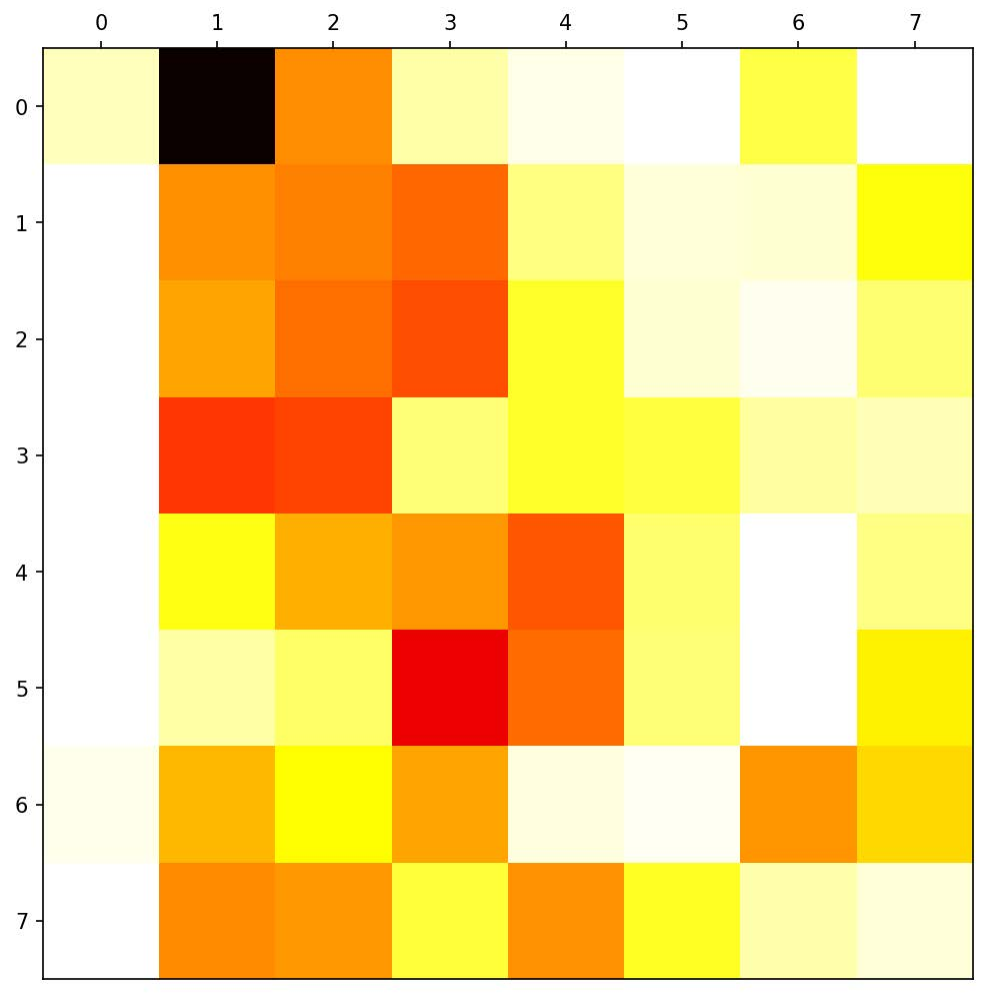}
\end{minipage}
}
\subfigure{
\begin{minipage}[t]{0.1\linewidth}
\centering
\includegraphics[width=2cm, height=2cm]{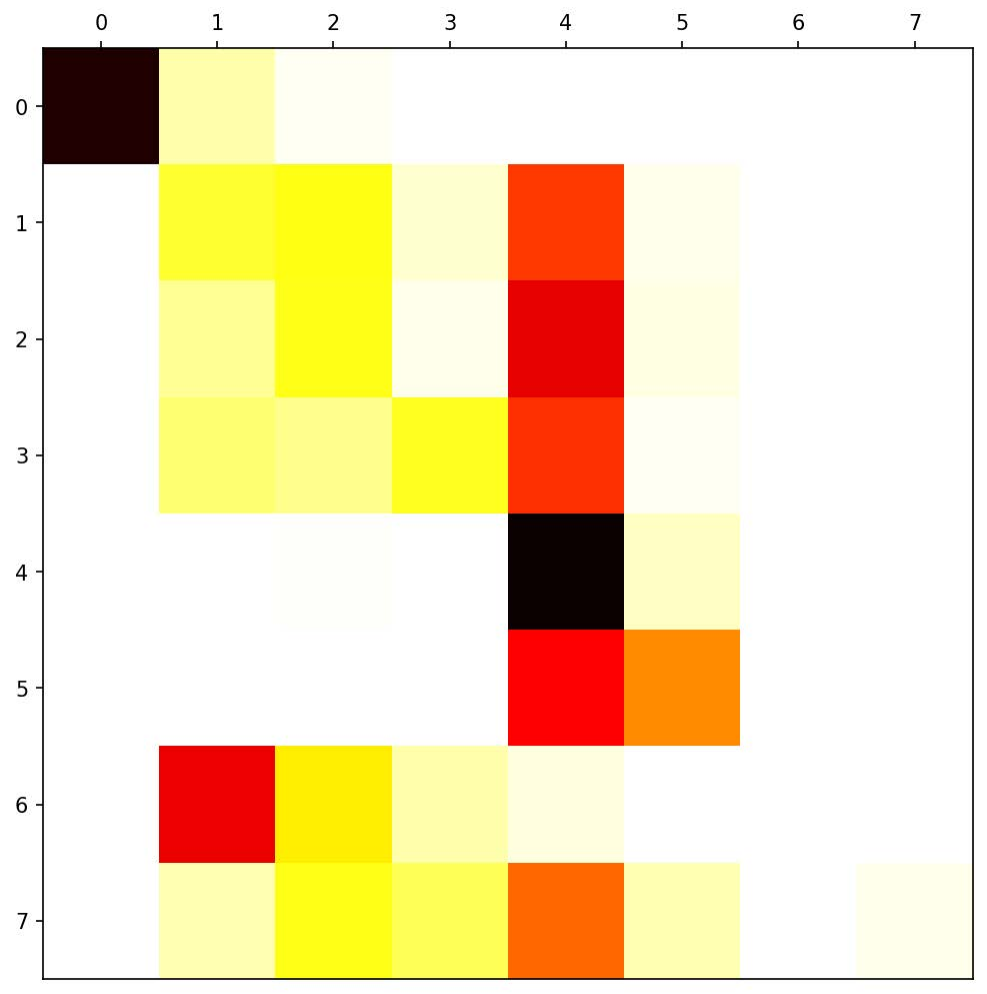}
\end{minipage}
}
\subfigure{
\begin{minipage}[t]{0.1\linewidth}
\centering
\includegraphics[width=2cm, height=2cm]{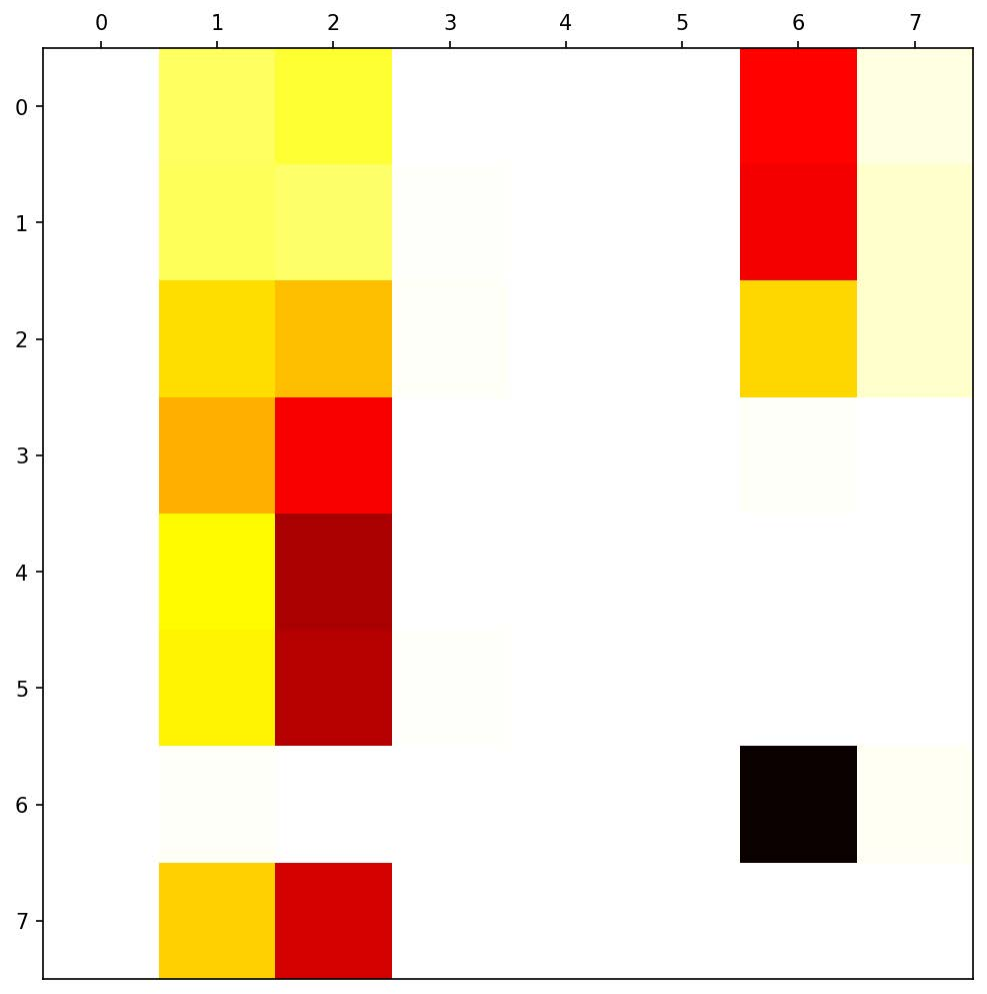}
\end{minipage}
}
\subfigure{
\begin{minipage}[t]{0.1\linewidth}
\centering
\includegraphics[width=2cm, height=2cm]{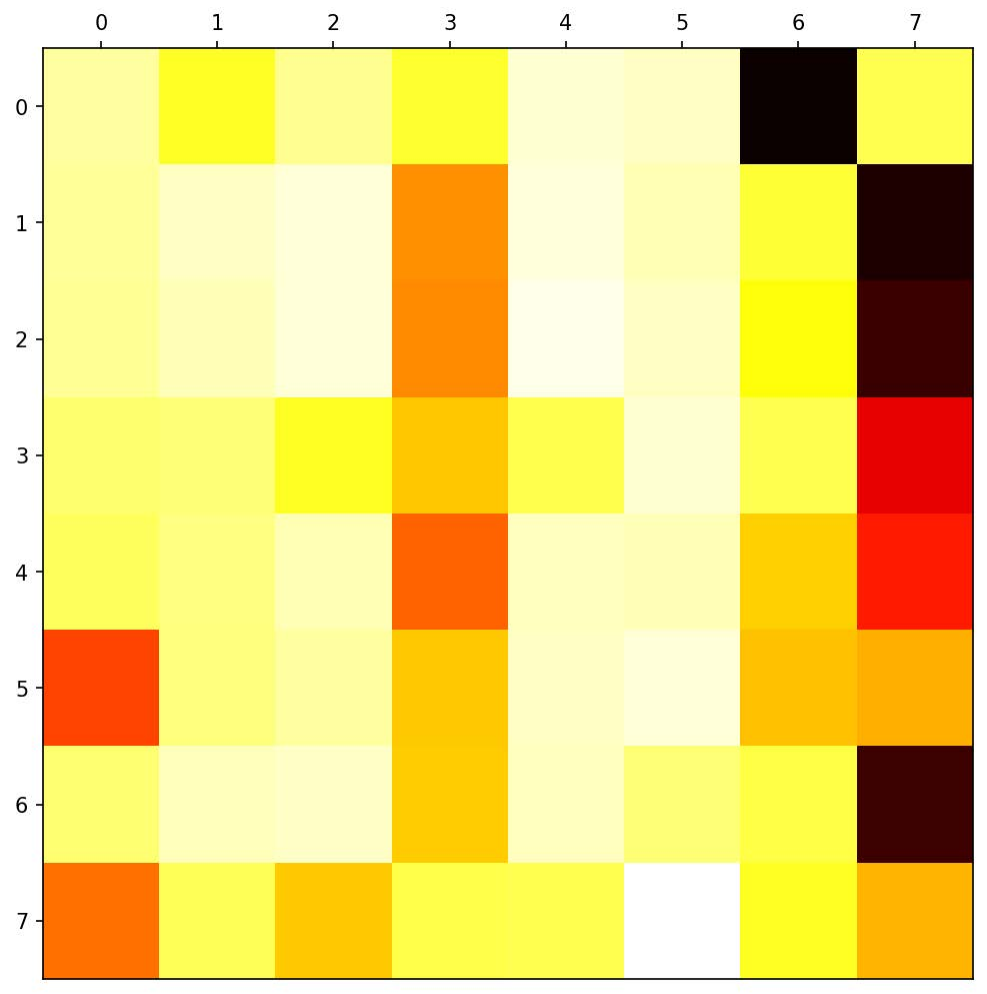}
\end{minipage}
}
\subfigure{
\begin{minipage}[t]{0.1\linewidth}
\centering
\includegraphics[width=2cm, height=2cm]{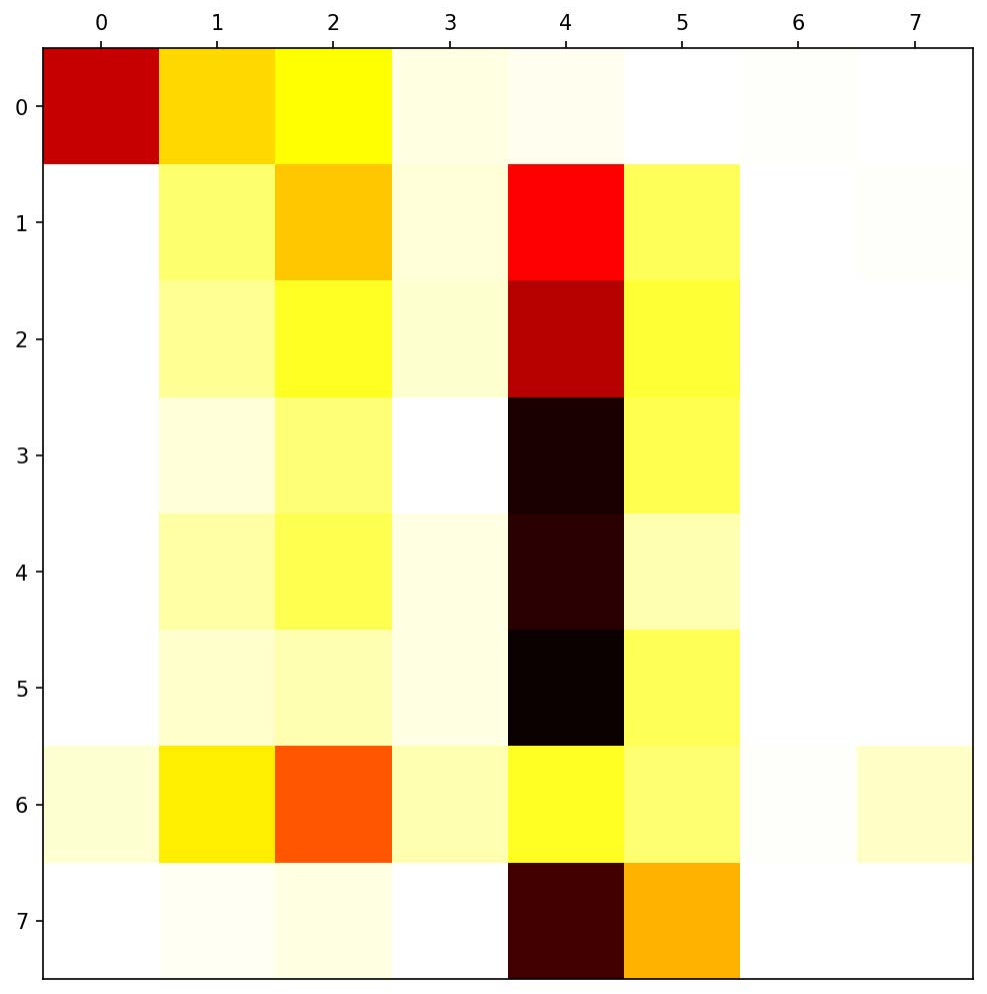}
\end{minipage}
}
\subfigure{
\begin{minipage}[t]{0.1\linewidth}
\centering
\includegraphics[width=2cm, height=2cm]{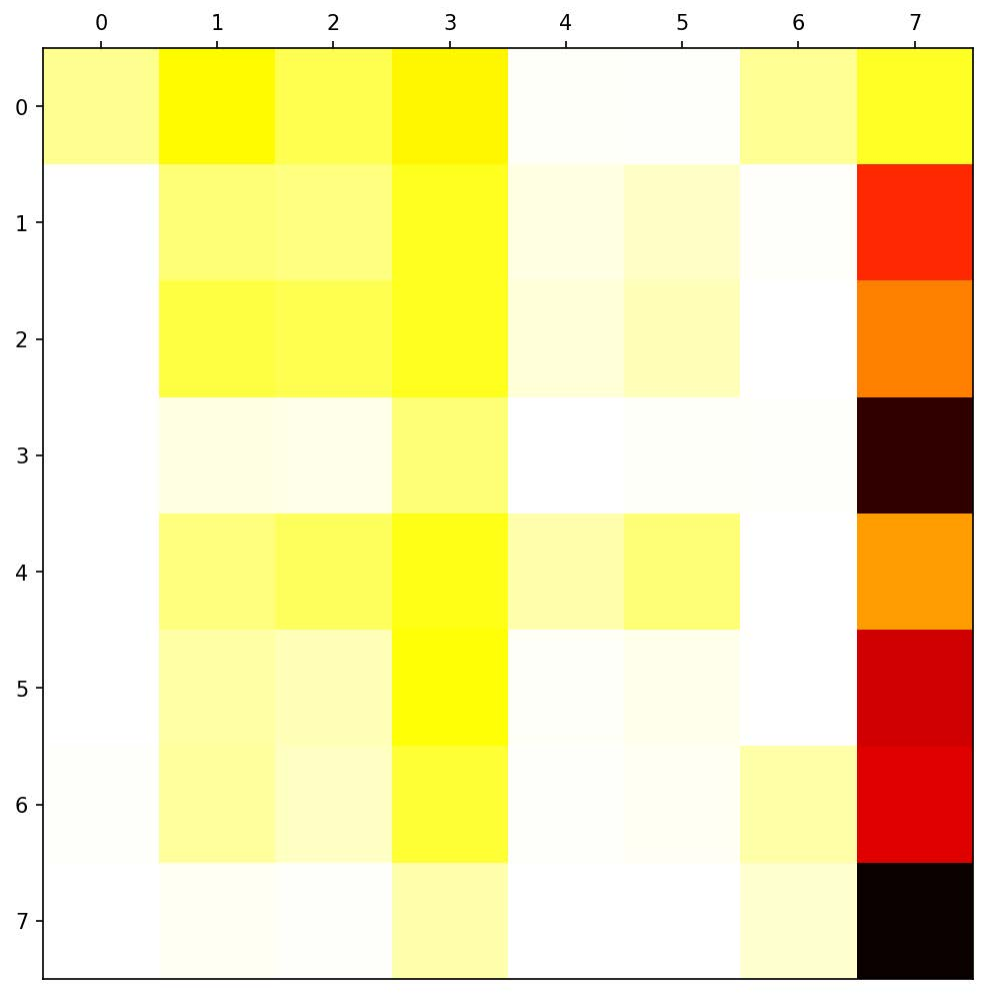}
\end{minipage}
}
\subfigure{
\begin{minipage}[t]{0.11\linewidth}
\centering
\includegraphics[width=2cm, height=2cm]{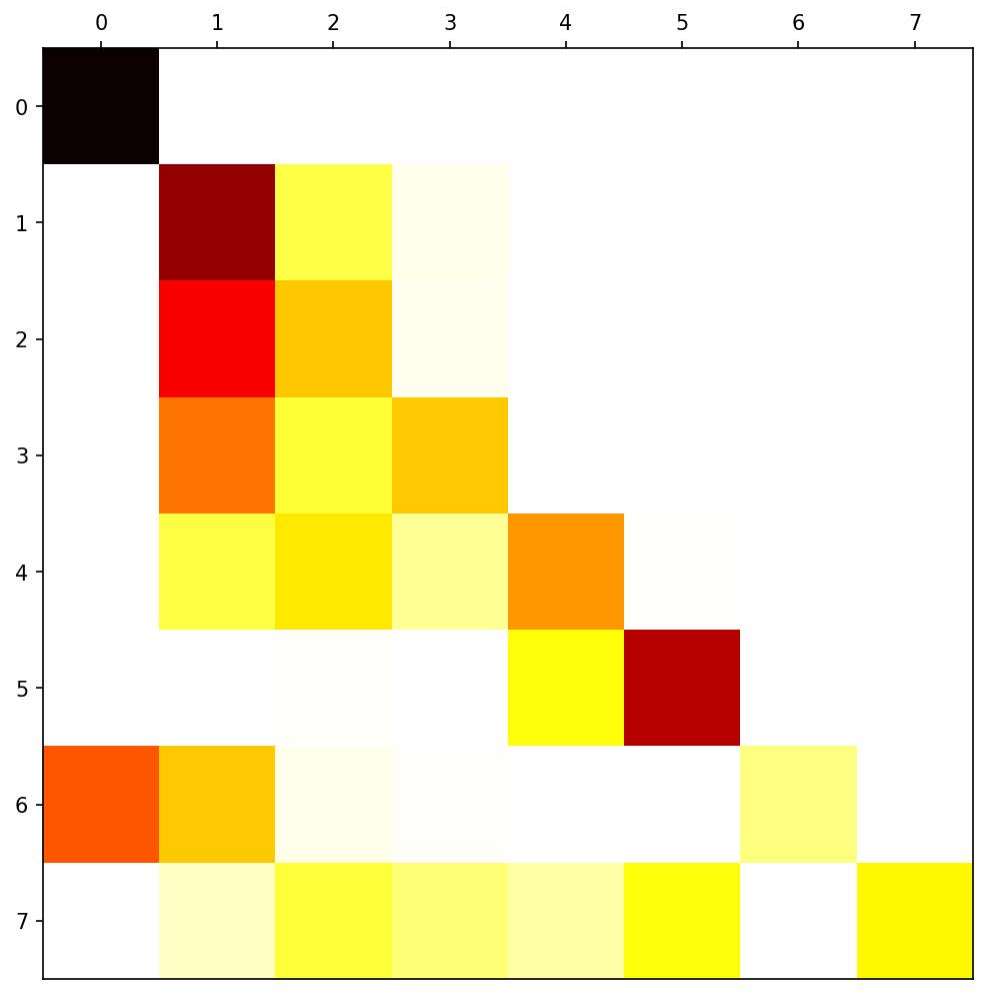}
\end{minipage}
}

\quad

\vspace{-0.5cm}
\setcounter{subfigure}{0}
\subfigure[head 1]{
\begin{minipage}[t]{0.1\linewidth}
\centering
\includegraphics[width=2cm, height=2cm]{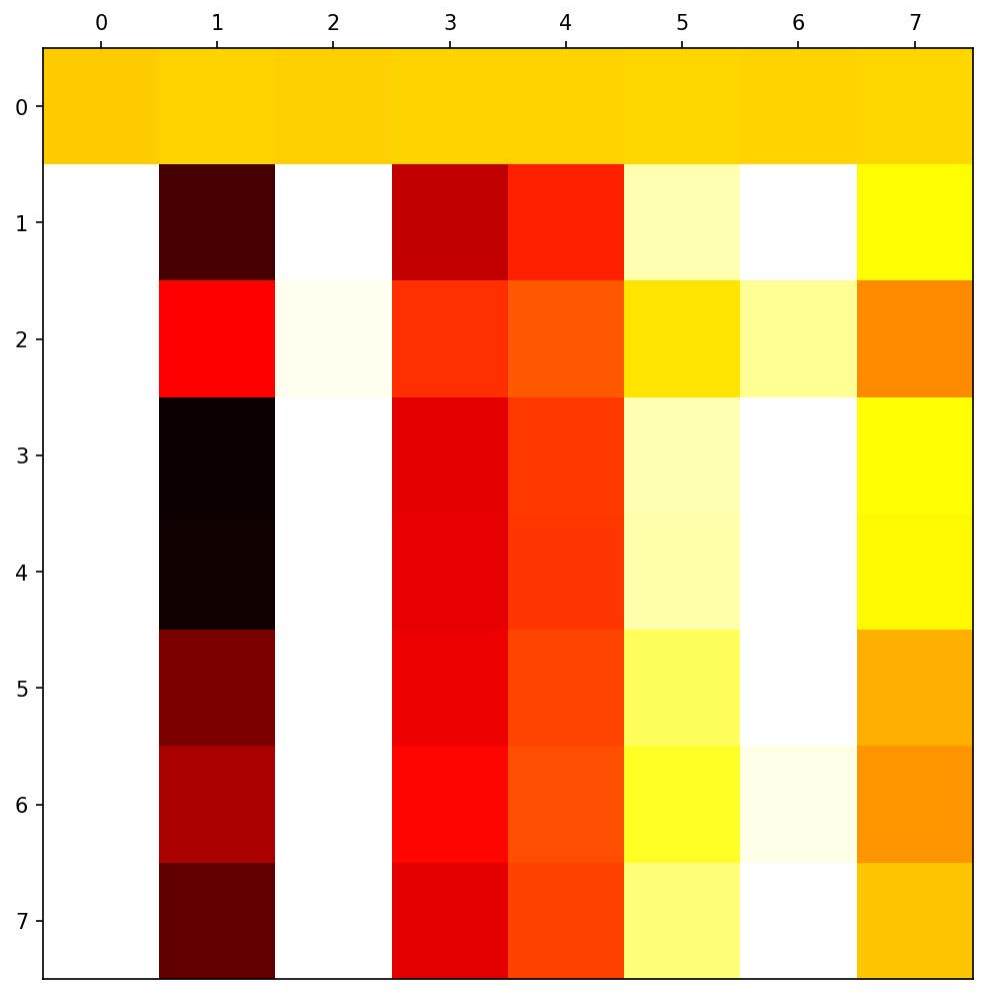}
\end{minipage}
}
\subfigure[head 2]{
\begin{minipage}[t]{0.1\linewidth}
\centering
\includegraphics[width=2cm, height=2cm]{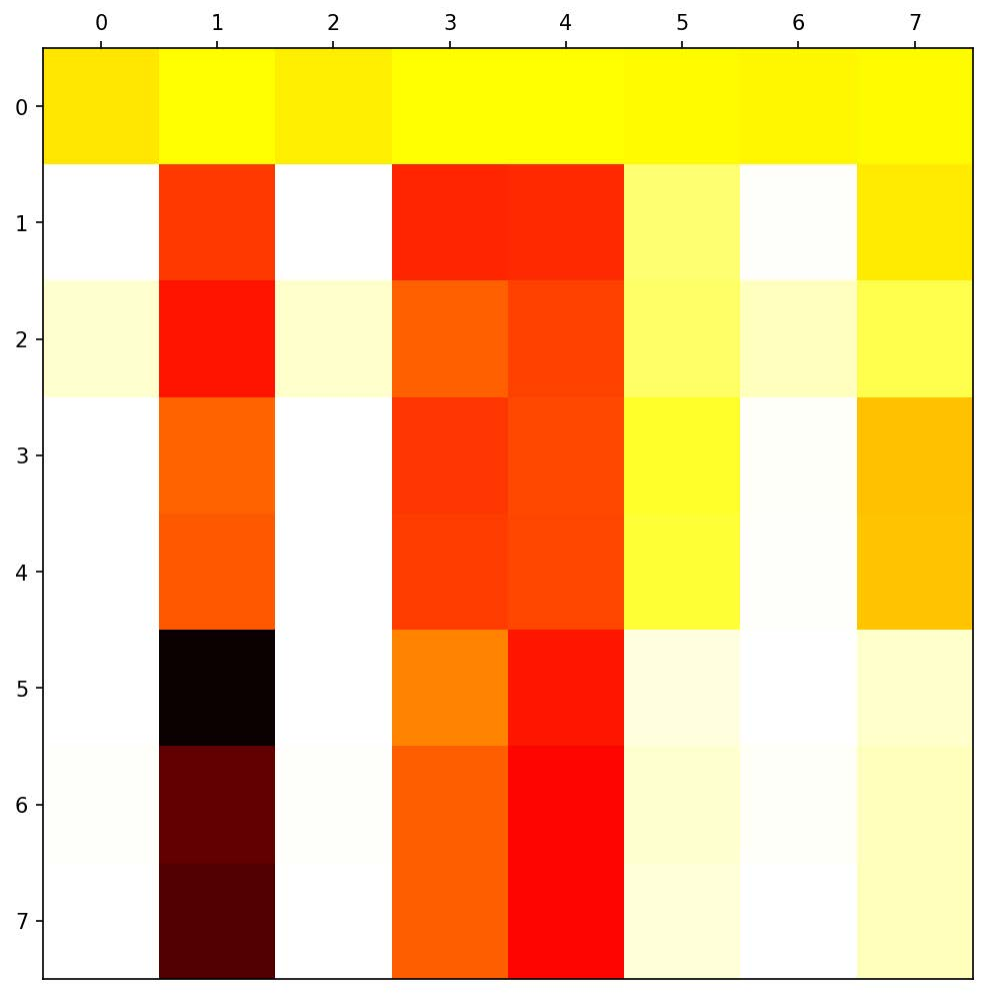}
\end{minipage}
}
\subfigure[head 3]{
\begin{minipage}[t]{0.1\linewidth}
\centering
\includegraphics[width=2cm, height=2cm]{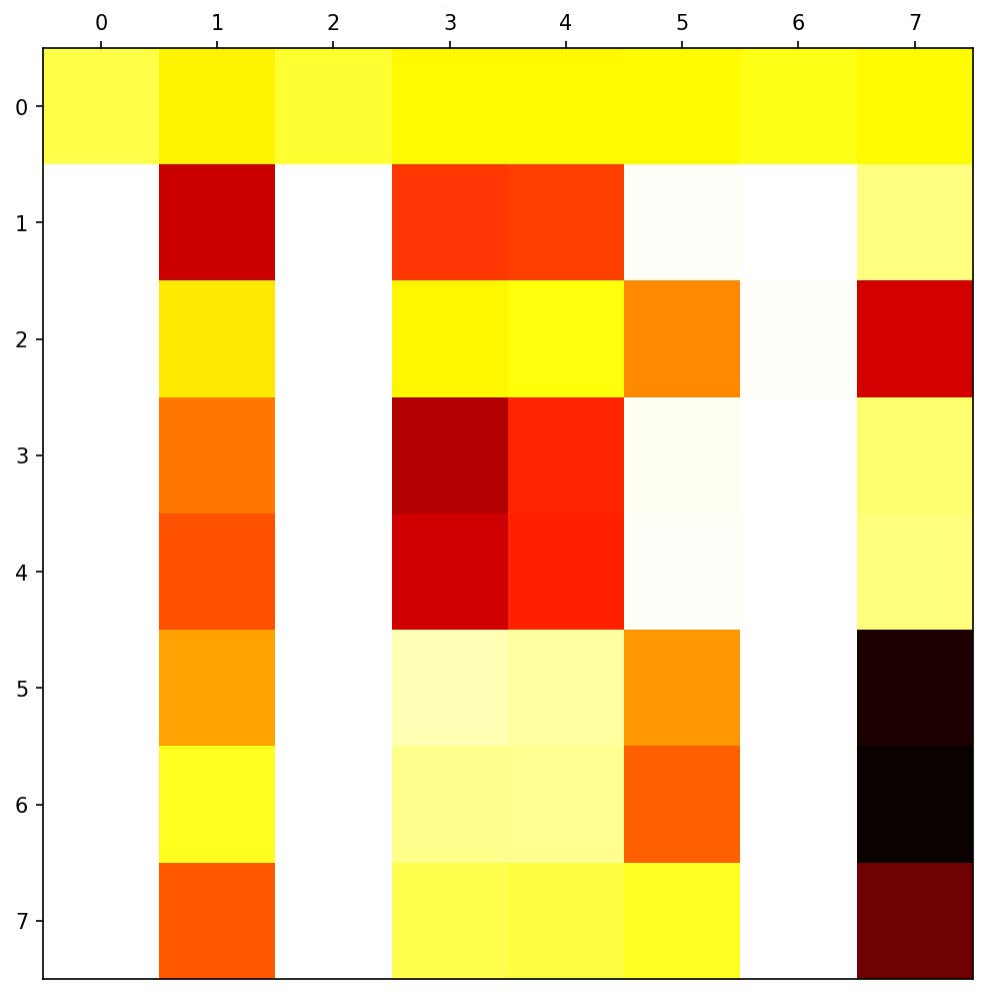}
\end{minipage}
}
\subfigure[head 4]{
\begin{minipage}[t]{0.1\linewidth}
\centering
\includegraphics[width=2cm, height=2cm]{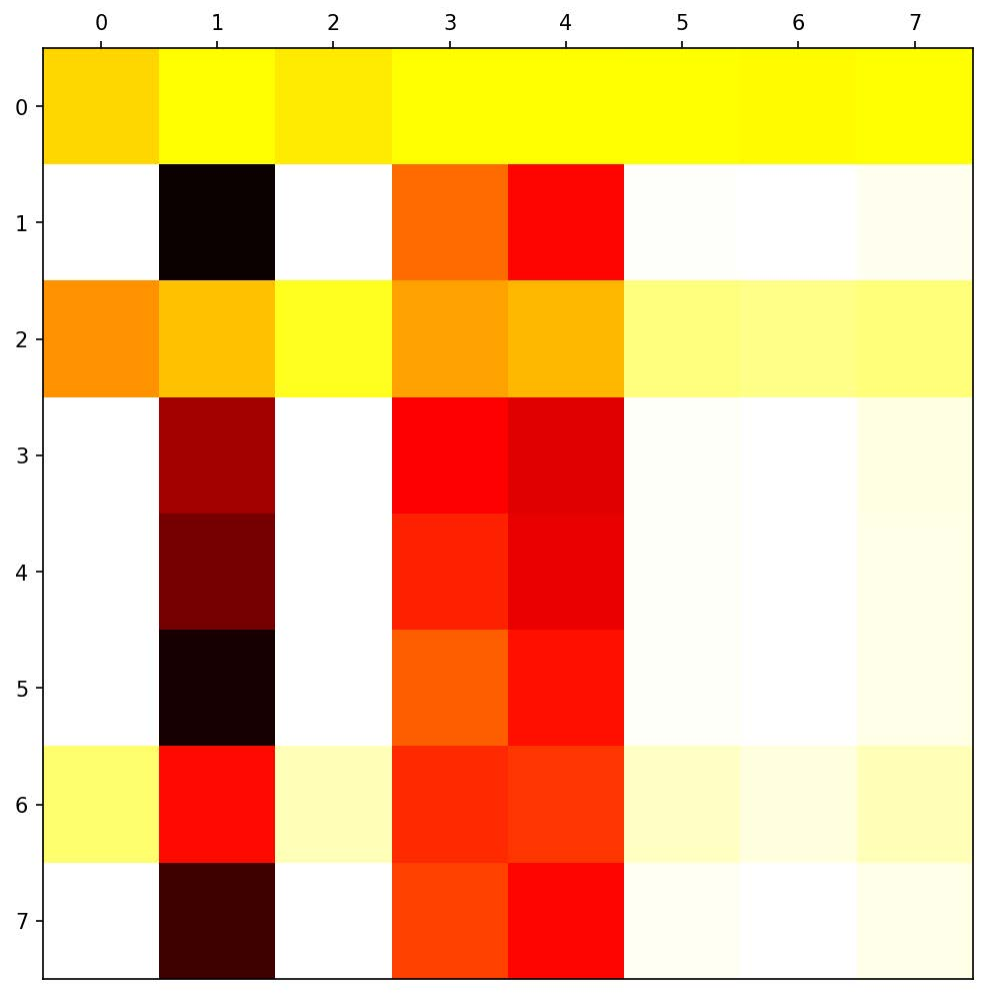}
\end{minipage}
}
\subfigure[head 5]{
\begin{minipage}[t]{0.1\linewidth}
\centering
\includegraphics[width=2cm, height=2cm]{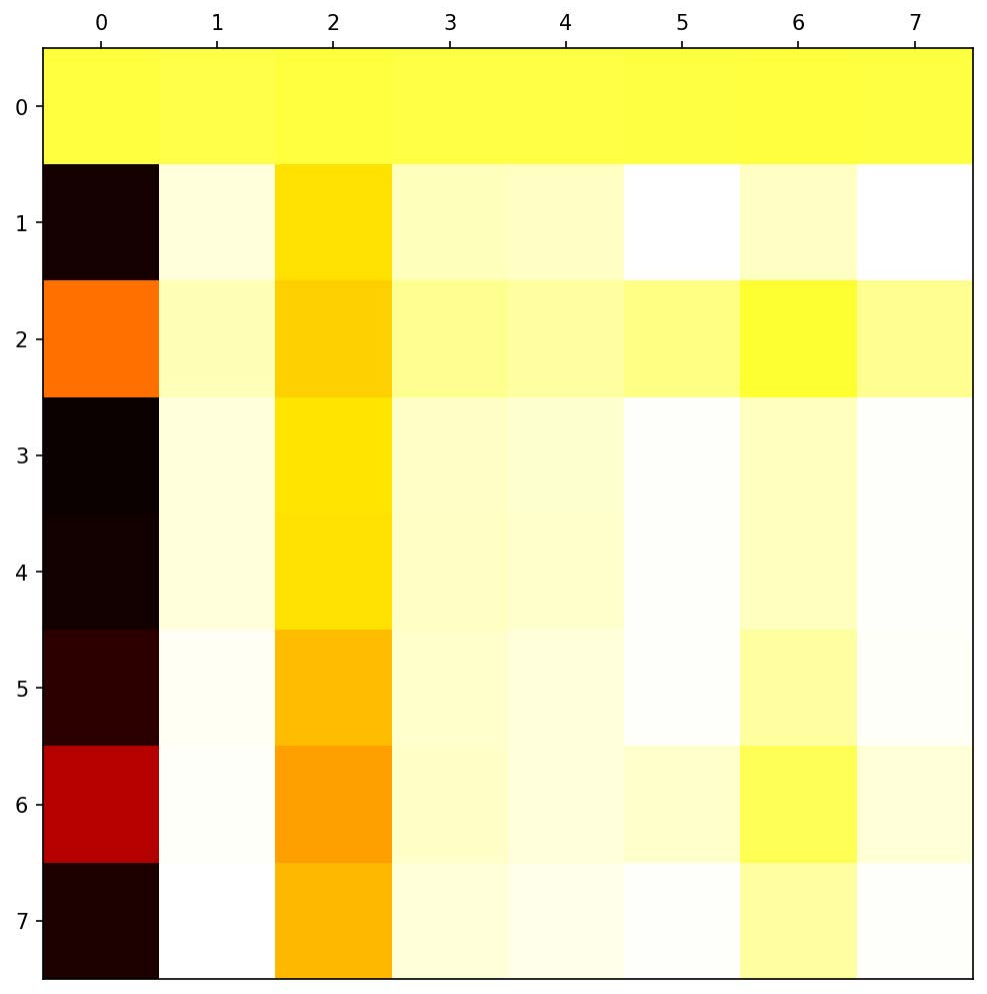}
\end{minipage}
}
\subfigure[head 6]{
\begin{minipage}[t]{0.1\linewidth}
\centering
\includegraphics[width=2cm, height=2cm]{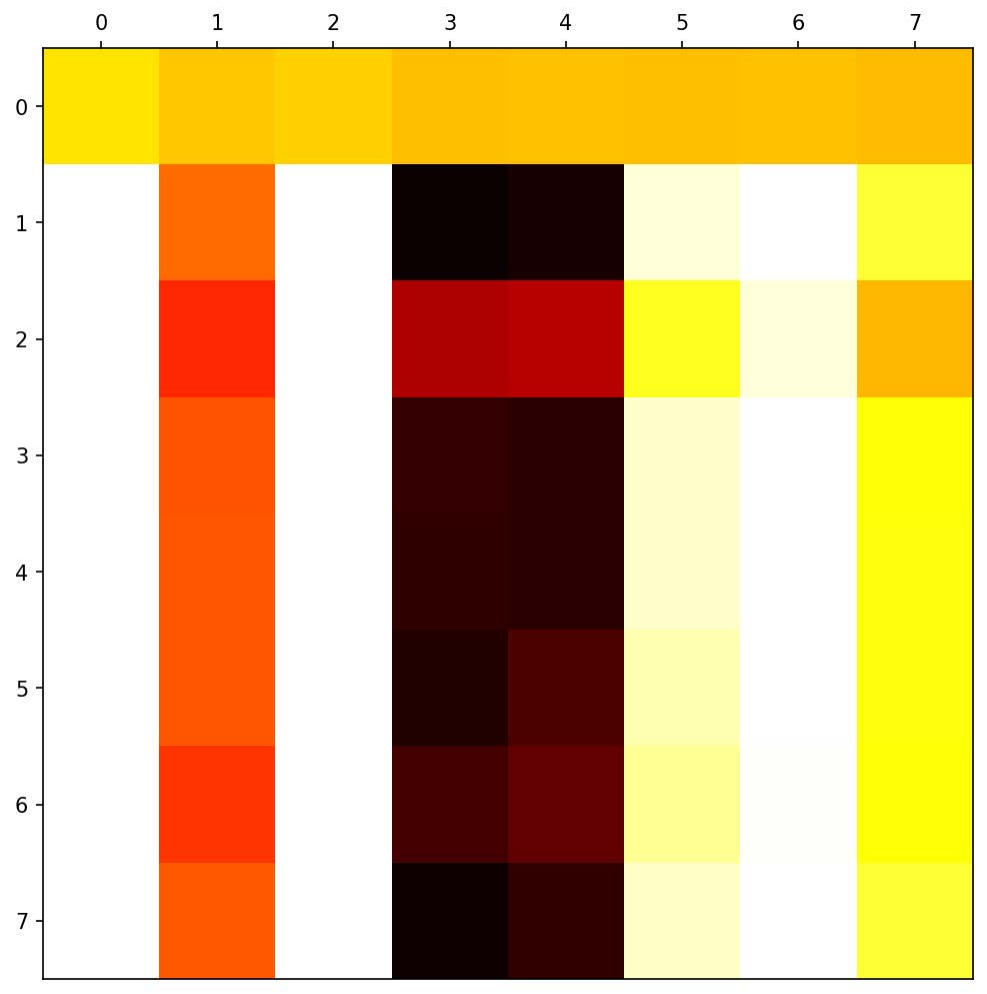}
\end{minipage}
}
\subfigure[head 7]{
\begin{minipage}[t]{0.1\linewidth}
\centering
\includegraphics[width=2cm, height=2cm]{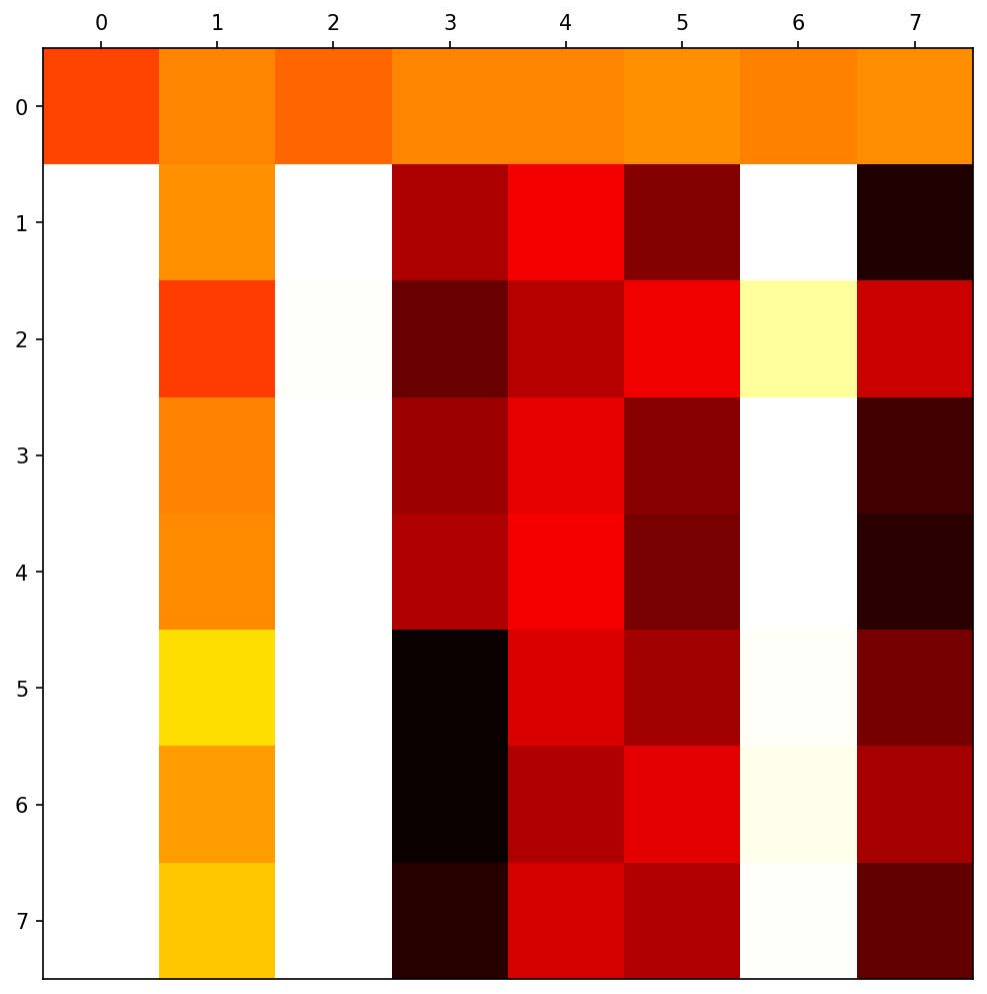}
\end{minipage}
}
\subfigure[head 7]{
\begin{minipage}[t]{0.11\linewidth}
\centering
\includegraphics[width=2cm, height=2cm]{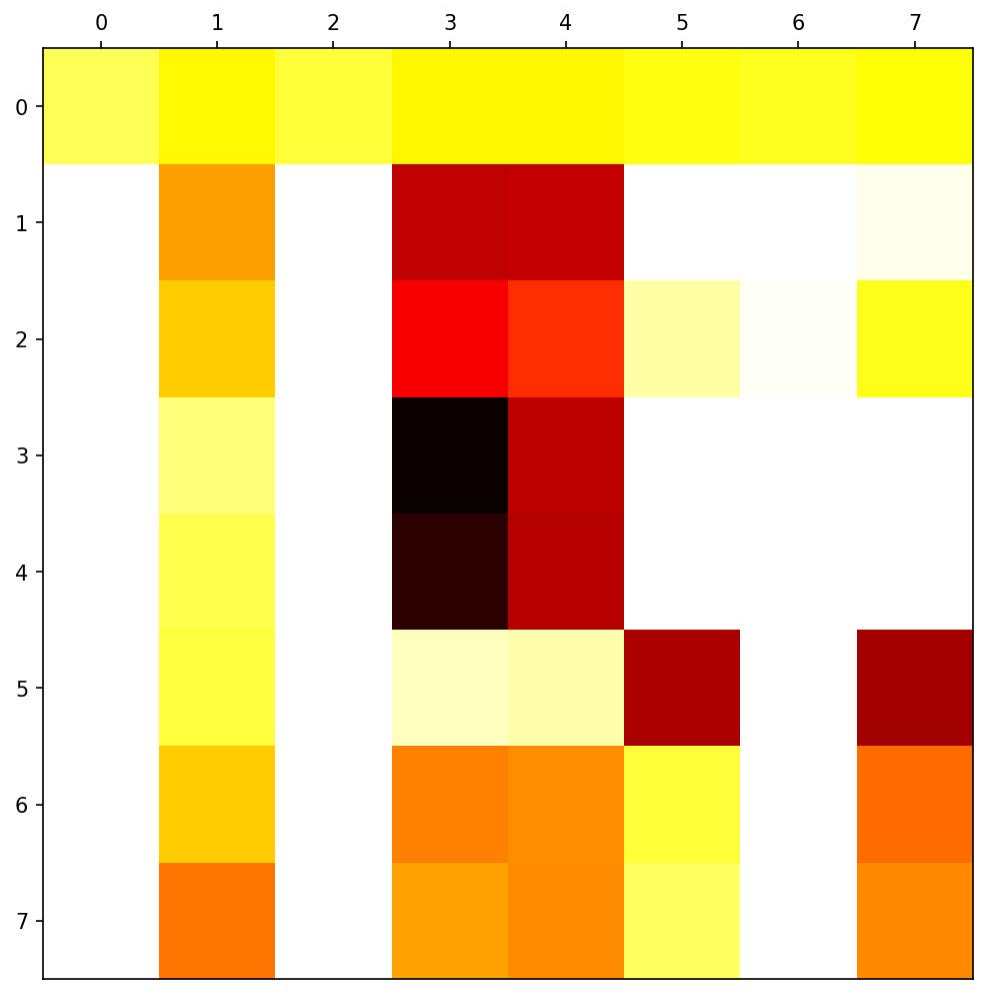}
\end{minipage}
}
\caption{\textbf{Inter-agent attention}: The entire figure represents different time steps along the rows and different attention heads along the columns. In each subfigure, the vertical axis represents the current agent, while the horizontal axis represents the agents being attended to by the current agent.}
\label{example of MAIPP problem}
\end{figure*}

The provided visualization illustrates the attention patterns of eight distinct heads within a multi-agent pathfinding scenario. In this context, darker shades signify higher attention values the given agent assigns to particular positions. The visualization sheds light on several crucial aspects that offer valuable insights into the agent's decision-making process and the role of multi-head attention.

Firstly, it is evident that the eight heads exhibit diverse attention distributions across the environment. This observation implies that each head focuses on different regions, indicating a multi-faceted approach to information processing. By employing multiple heads, the model can effectively focus on diverse spatial configurations simultaneously, enabling it to capture intricate spatial relationships and make well-informed decisions.

Secondly, the changes in attention towards other agents as the agent's position varies highlight the adaptability of multi-head attention. While some heads may focus more on the agent's immediate vicinity, others are likely to give closer attention to areas that potentially influence the agent's trajectory. This adaptive behavior enables the agent to be situationally aware, fostering cooperation or avoidance strategies based on the spatial context.

Additionally, the utilization of multi-head attention in multi-agent pathfinding enables the model to learn a diverse set of attention mechanisms. Some heads may consistently allocate great attention to specific regions along the agent's path, indicating their importance for navigation. On the other hand, certain heads might distribute attention more uniformly across all nodes, demonstrating their role in providing the agent with a broader understanding of the global information. This combination of localized and global attention allows the model to handle complex and dynamic environments more effectively.

In conclusion, the visualization and analysis of multi-head attention in the context of multi-agent pathfinding provide crucial insights into the agent's spatial reasoning and adaptive behavior. By leveraging multiple heads, the model achieves a nuanced understanding of the environment, facilitating efficient navigation and cooperation with other agents. 
This approach represents a promising direction in advancing multi-agent systems, contributing to improved performance and adaptability in real-world scenarios.

\subsection{Simulations in Different Types of Map}

    \begin{figure*}
    \centering
    \subfigure[warehouse world]{\label{warehouse-world}
    \includegraphics[width=0.31\linewidth]{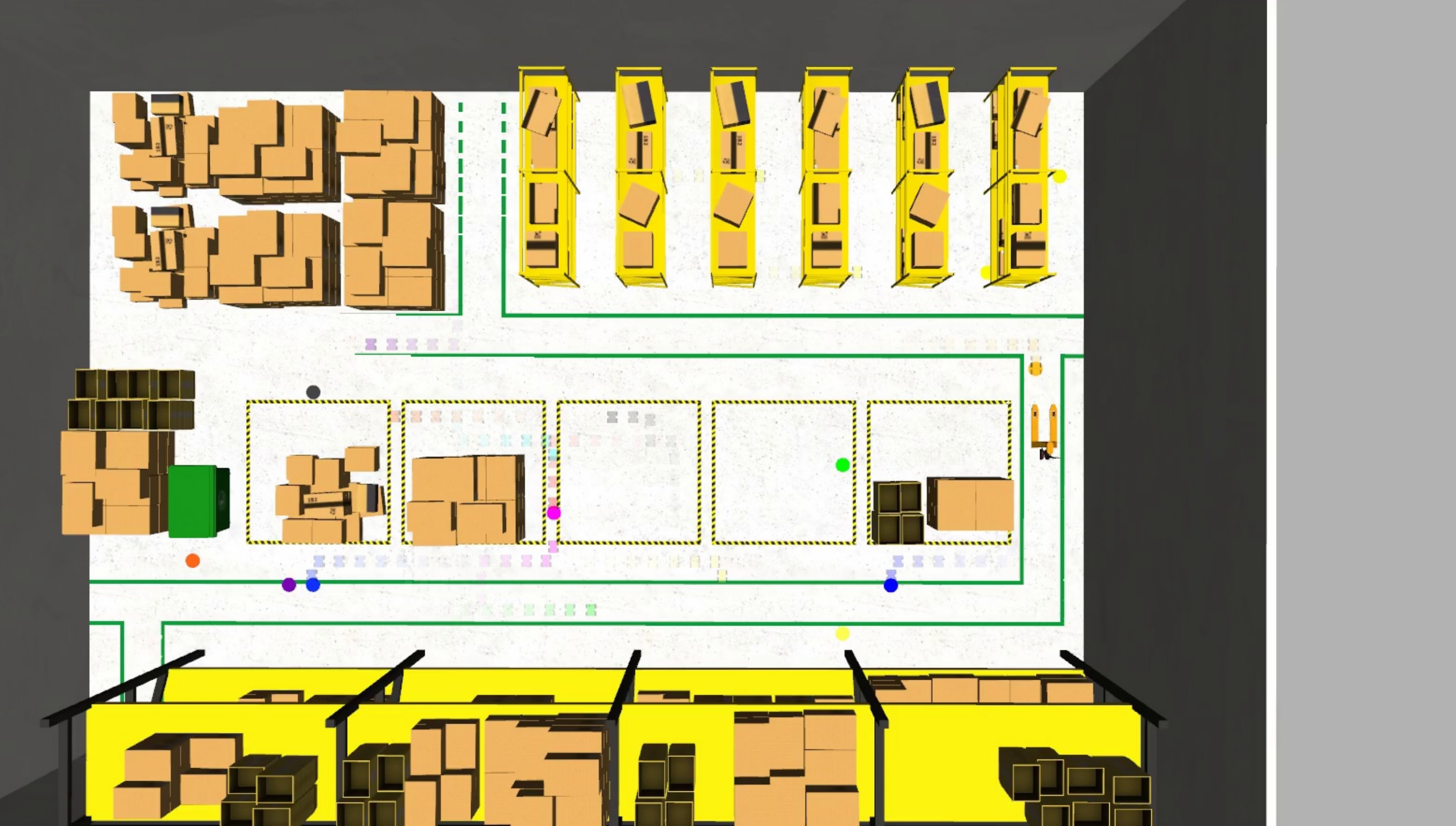}}
    \hspace{0.01\linewidth}
    \subfigure[warehouse world]{\label{bookstore-world}
    \includegraphics[width=0.31\linewidth]{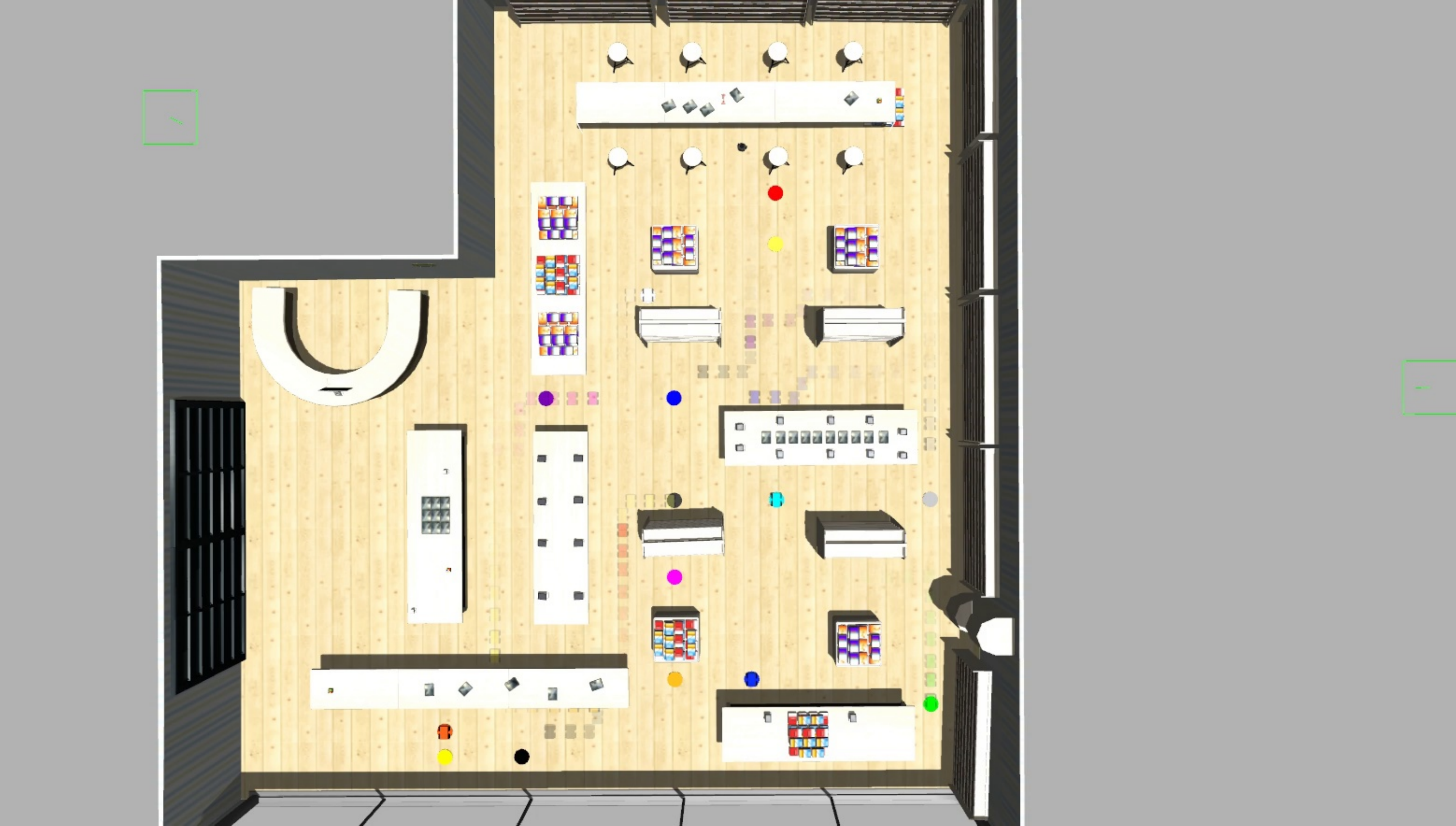}}
    \hspace{0.01\linewidth}
    \subfigure[warehouse world]{\label{hospital-world}
    \includegraphics[width=0.31\linewidth]{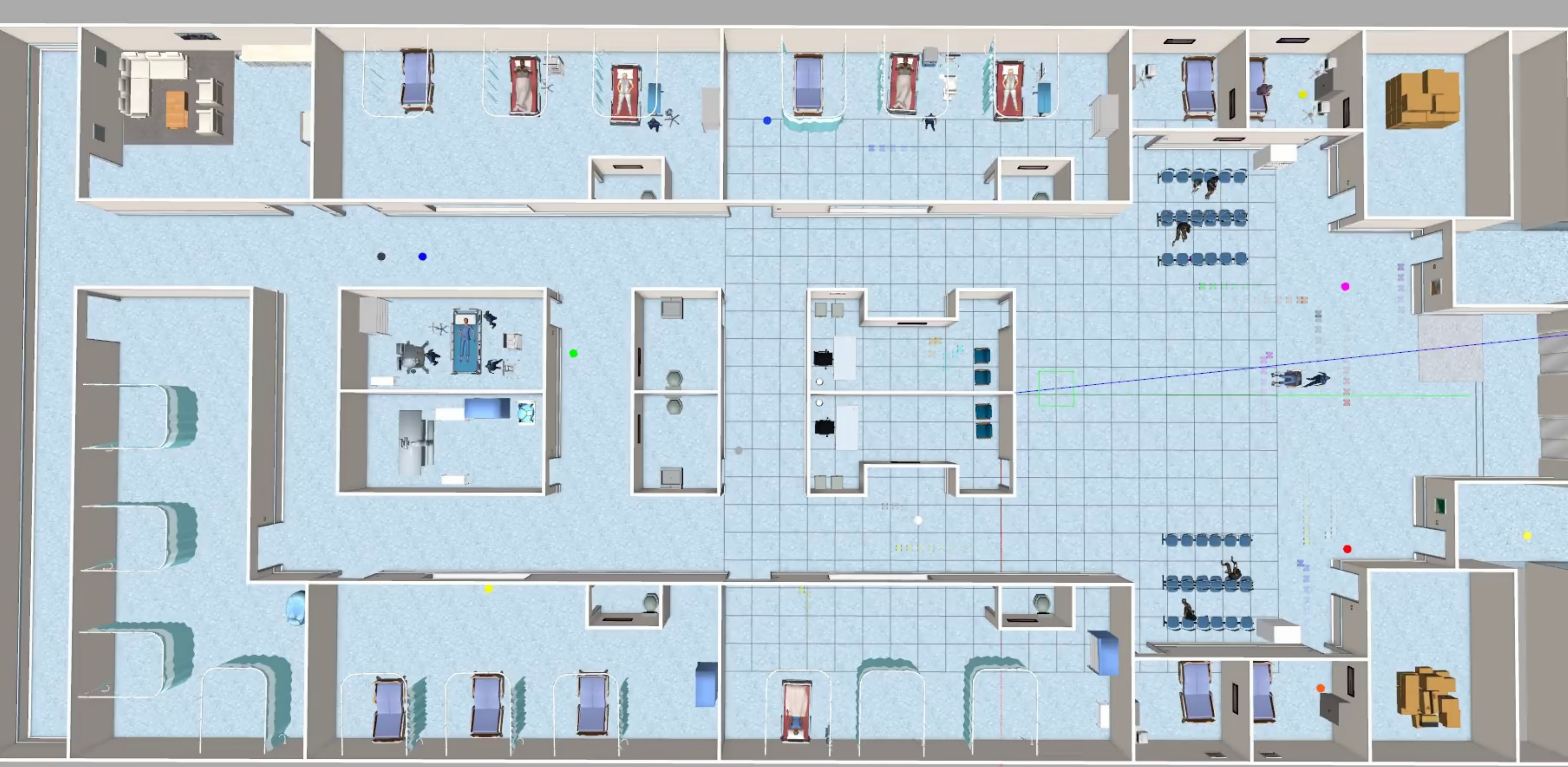}}
    \caption{
    These images capture snapshots of the robots' movement across a series of steps. Circles represent the goal locations, while the rectangles indicate the positions of the robots. A darker shade indicates a more recent step in their motion. 
    }
    \label{aws_test}
    \end{figure*}

    \begin{figure*}
    \centering
    \subfigure[base environment]{\label{no-robot}
    \includegraphics[width=0.22\linewidth]{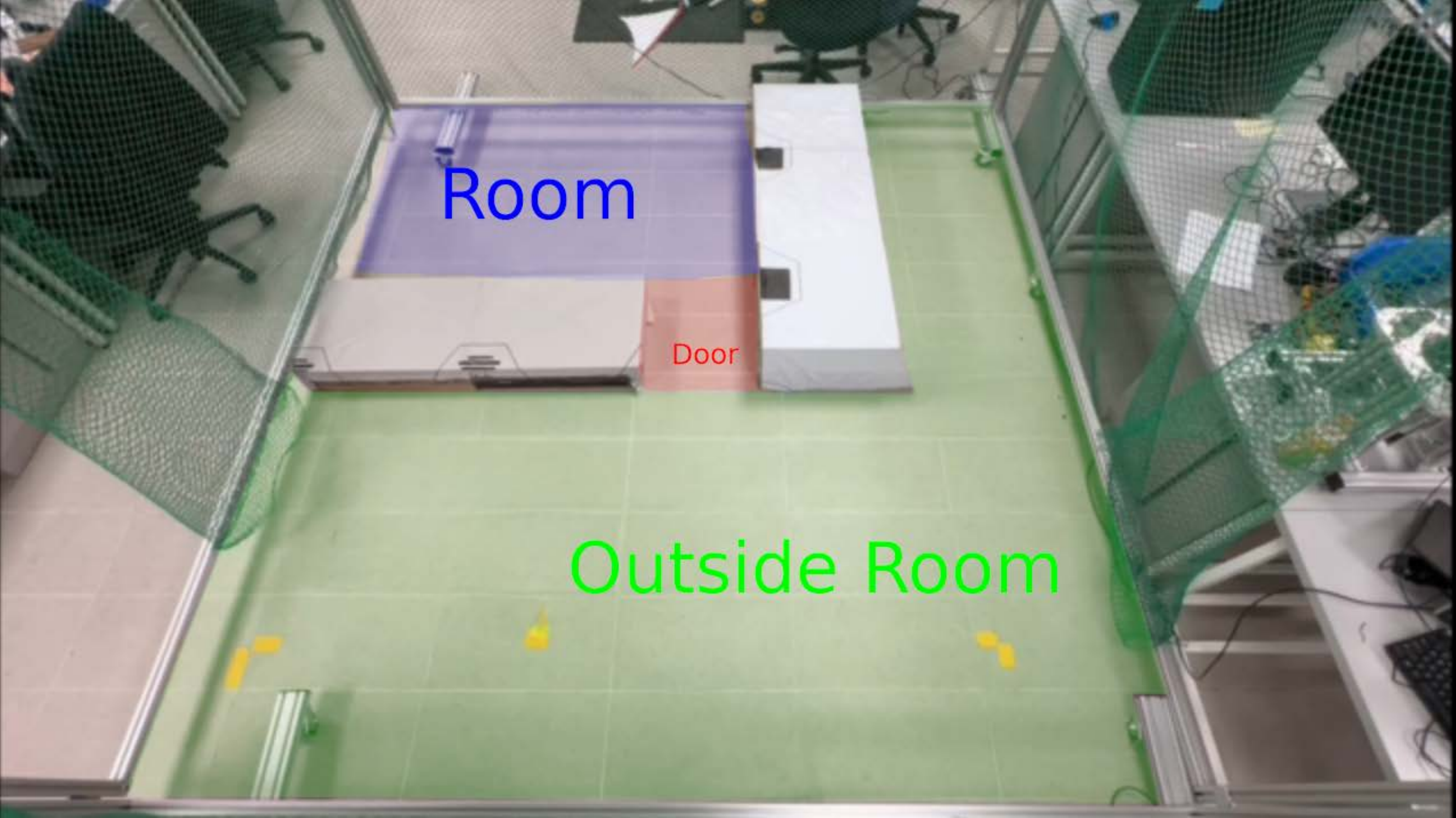}}
    \hspace{0.01\linewidth}
    \subfigure[outside-1 path]{\label{one-robot}
    \includegraphics[width=0.22\linewidth]{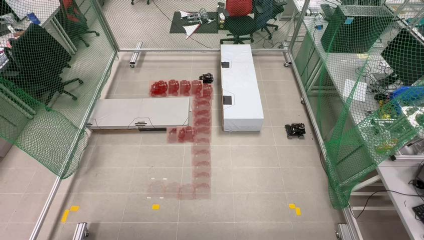}}
    \hspace{0.01\linewidth}
    \subfigure[outside-2 path]{\label{two-robot}
    \includegraphics[width=0.22\linewidth]{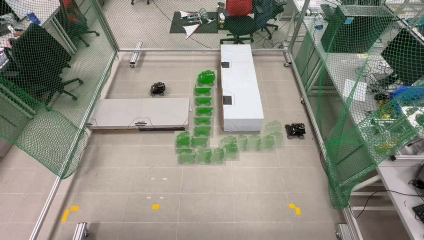}}
    \hspace{0.01\linewidth}
    \subfigure[inside-1 path]{\label{three-robot}
    \includegraphics[width=0.22\linewidth]{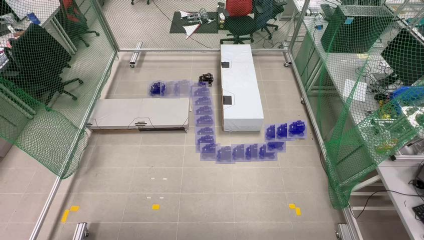}}
    \caption{
     Figure \ref{no-robot} shows various regions of the environment, and \ref{one-robot}, \ref{two-robot}, \ref{three-robot} show the paths taken by the various robots to reach their goals. A darker shade indicates closer proximity to the goal. \ref{one-robot} and \ref{two-robot} are robots that start outside the room and reach their goal inside the room, whereas \ref{three-robot} is the robot that starts inside the room and reaches its goal outside.
    }
    \label{map_analysis}
    \end{figure*}

    To illustrate the ability of ALPHA to be deployed in the real world, we tested it in three standard aws-robotics simulation environments (as shown in Fig \ref{aws_test}). 
    \subsubsection{Setup}
    We chose Gazebo as our simulator due to its high compatibility with ROS. We selected fifteen robots for the simulations, assigning each one a pre-set goal to achieve.
    Considering our assumption that robots can move freely in any cardinal direction, we designed robot models sized at $0.5m\times 0.4m$, equipped with mecanum wheels and individual PID controllers.
    Furthermore, to meet our assumption of a known environment, we initially mapped the surroundings using the SLAM Karto method. We carefully choose the resolution for the occupancy map so each cell of the occupancy grid map can accommodate one robot but at the same time such that we don't lose important environmental information like doorways or compact alleys. The assumption of one robot per grid cell is essential because if a robot's body spans more than one grid cell, it could potentially collide with other robots occupying neighboring cells.
    After multiple iterations, the resolution of the maps was chosen to be 0.5m per occupancy grid block.
    
    \subsubsection{Environment Selection}
    Considering the two primary scenarios that receive significant attention within the MAPF community – structured warehouse-like and random environments, we opted for the \textit{AWS RoboMaker Small Warehouse World\footnote[1]{https://github.com/aws-robotics/aws-robomaker-small-warehouse-world}} and \textit{AWS RoboMaker Bookstore World\footnote[2]{https://github.com/aws-robotics/aws-robomaker-bookstore-world}} as representations of these settings.
     Additionally, to depict highly structured room-like scenarios, we selected the \textit{AWS RoboMaker Hospital World\footnote[3]{https://github.com/aws-robotics/aws-robomaker-hospital-world}}.

    \subsubsection{Results}
     ALPHA guided the robots in the following manner: It received the initial and final positions of each robot, along with the environment map, and used it to generate a path for each robot.  The robots in the environment then used the generated path as immediate waypoints and followed them to their goals.
     Although the maps are quite different from the grid world used in our training, the robots are still able to coordinate and reach the pre-defined goal.
     Notably, the impact of the long-horizon planner becomes evident in the hospital world. In this highly structured environment, robots often encounter rooms that appear to be positioned in the direction of their goals but do not ultimately lead to those goals. In such instances, robots refrain from entering rooms and opt for paths that might appear uncertain or longer when relying solely on FOV information but are, in fact, more optimal.

\subsection{Experimental Validation}

    \begin{figure*}
    \centering
    \subfigure[room-based map]{\label{room-based map}
    \includegraphics[width=0.22\linewidth]{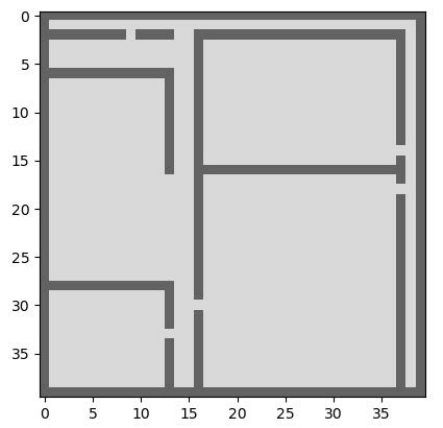}}
    \hspace{0.01\linewidth}
    \subfigure[skeletonized map]{\label{skeletonized map}
    \includegraphics[width=0.22\linewidth]{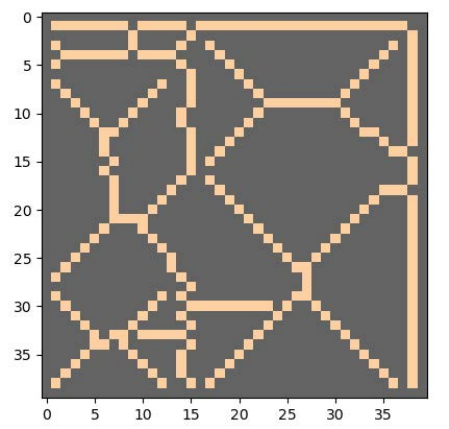}}
    \hspace{0.01\linewidth}
    \subfigure[important nodes]{\label{neighborhood analysis}
    \includegraphics[width=0.22\linewidth]{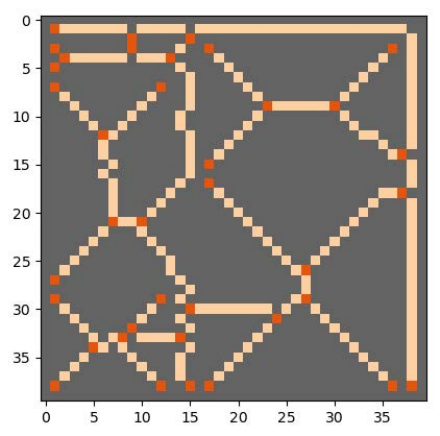}}
    \hspace{0.01\linewidth}
    \subfigure[final graph]{\label{nodes on map}
    \includegraphics[width=0.22\linewidth]{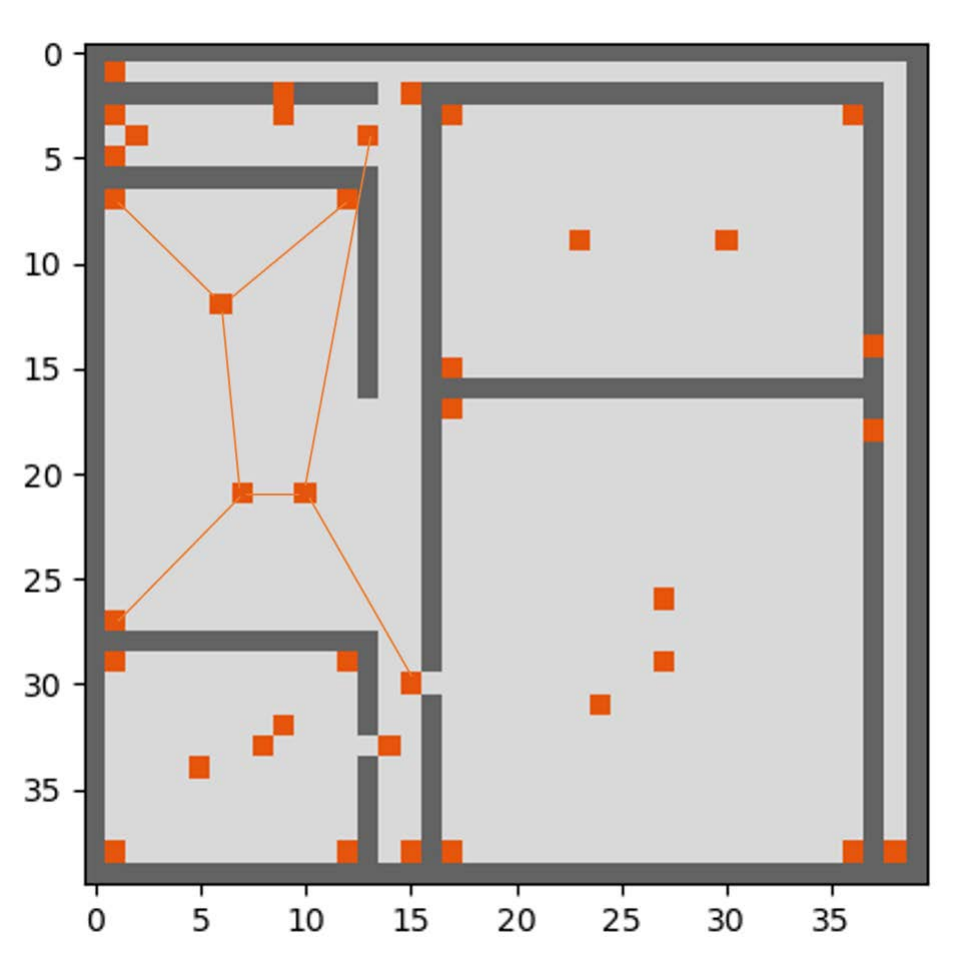}}
    \caption{In the original map (shown as Fig.~\ref{room-based map}), dark gray cell indicates static obstacle; light gray cell indicates free cell.
    In the skeletonized map, most of the free cells are padded, leaving only one pixel width of the skeleton represented in light yellow. 
    Then, by performing neighborhood analysis, special points are extracted as "nodes" in the refined map (as shown in Fig.~\ref{neighborhood analysis}), which are represented by orange.
    Finally, these nodes are mapped to the original map, through which the features of the room-based map can be sparsely represented.
    }
    \label{map_analysis}
    \end{figure*}

    \begin{figure}
    \centering
    \subfigure[]{\label{small_pre}
    \includegraphics[width=0.7\linewidth]{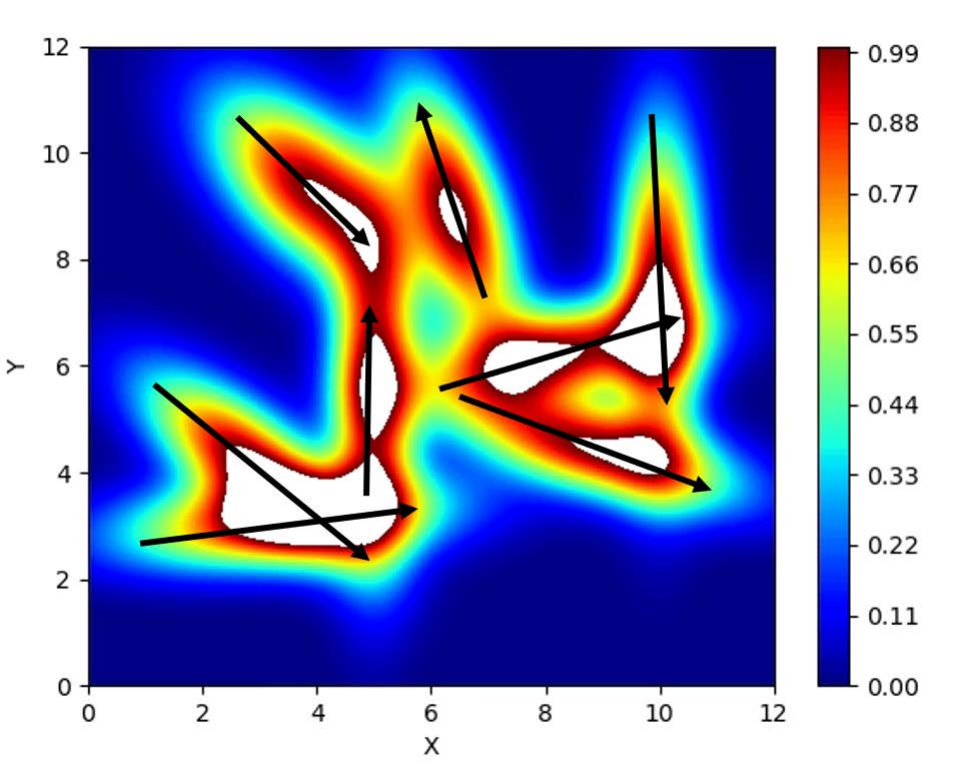}}
    \hspace{0.01\linewidth}
    \subfigure[]{\label{big}
    \includegraphics[width=0.7\linewidth]{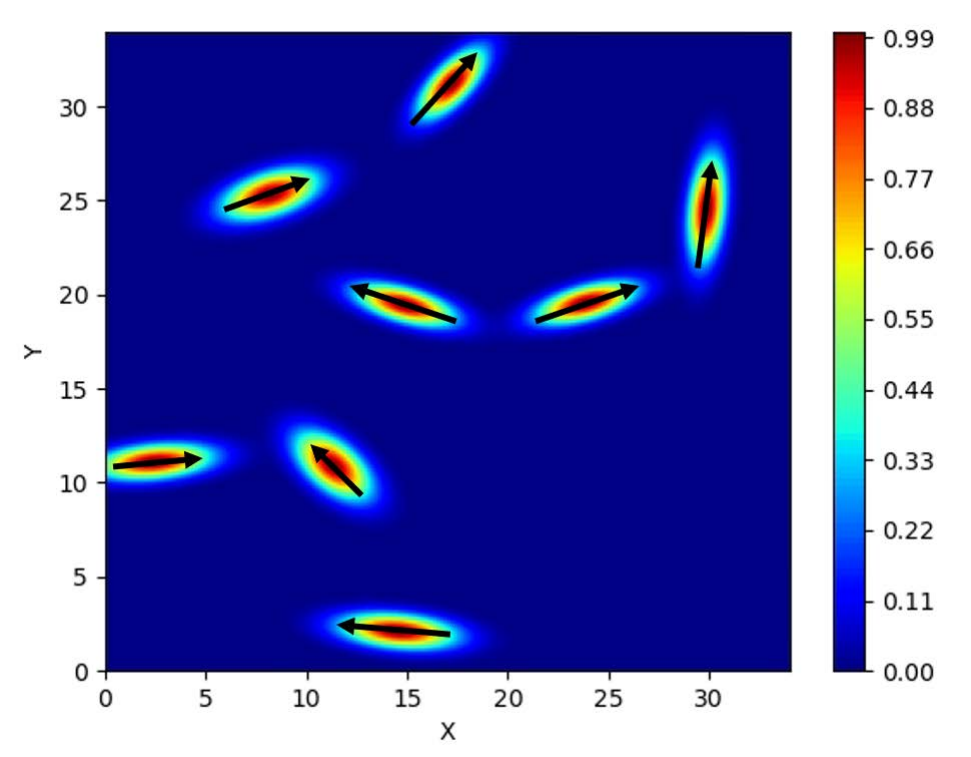}}
    \caption{
        (a) and (b) represent the two-dimensional Gaussian distributions of the intentions of 8 agents predicted under the same frontier (=5 in practice) but in maps of different sizes.
        Warmer colors indicate a higher probability of agents appearing in that position, while colder colors indicate a lower probability.
        The black arrows indicate the possible directions of the agents.
    }
    \label{agent_prediction}
    \end{figure}
    
    \subsubsection{Setup}
       We configured an environment measuring $3.25m \times 3.25m$ containing two long, wall-like obstacles arranged to create a room with a single entrance(as shown in Fig \ref{no-robot}).
       We used three \textit{Sparklife Omnibots} robots with dimensions of approximately $0.25m \times 0.25m$ equipped with mecanum wheels for omnidirectional movement.
       For the precise positioning of the robots, we relied on the \textit{OptiTrack Motion Capture System}.
       Similar to simulations, we used SLAM karto to map the environment for the robots. For this experiment, we chose the resolution to be 0.25m per occupancy grid block, as any more than this was unable to detect the doorway as free space, and any less would lead to one robot occupying more than one grid block, hence leading to collisions between robots.
       We set the starting positions of two robots outside the room and that of the third robot inside.
       We configured the goal positions for the two robots outside the room to be located inside the room while the robot inside the room had its goal positioned outside. This setup increased the number of robots passing through the room's door, turning it into a bottleneck and helping us evaluate how well the robots collaborated.
    \subsubsection{Results}
        The robots displayed excellent collaboration, as evident in Fig \ref{one-robot}, \ref{two-robot}, \ref{three-robot}, which depict the path of the robots: $outside-one$, $outside-two$, and $inside-one$, respectively. 
        At the start, both $outside-one$ and $inside-one$ robots approached the door simultaneously, but $outside-one$ reversed and moved aside, allowing $inside-one$ to pass through. This interaction is evident as the leftward deviation in Fig~\ref{one-robot}. During this exchange, precisely when $outside-one$ yielded the passage to $inside-one,$ $outside-two$ was in the expected path of $inside-one.$ Consequently, $outside-two$ adjusted its position to facilitate $inside-one's$ passage, as indicated by the downward curve in Fig~\ref{two-robot}. 
        One can also refer to \ref{three-robot} to verify that it suffered no deviation from its path at any point.
        Subsequently, after $inside-one$ had a clear path, $outside-two$ queued behind $outside-one$ to pass through the door. This sequence of actions is reflected in the leftward motion depicted in Fig~\ref{two-robot}. Eventually, $outside-two$ positioned itself next to $outside-one,$ awaiting $outside-one$ to cross the door and then followed it toward the goal.

\subsection{Additional Details}

    \subsubsection{Graph Extraction}

    As mentioned in the main text, we need to distill a graph that can be used to represent global world from a highly structured room-like map.
    The extraction of graph is divided into three steps, namely refining the map, extracting nodes, and constructing the graph.
    In this subsection, we will visualize these steps as shown in Fig. \ref{map_analysis}.
    The method used to thin the map in Fig. \ref{map_analysis} is Medial Axis Transform which will generate more branches than the Zhang-Suen algorithm.
    We ultimately chose this method because we wanted to capture as much detail as possible in the global map (as shown in Fig. \ref{room-based map}), and the results are shown in Fig. \ref{skeletonized map}.
    Subsequently, based on neighborhood analysis, that is, analyzing whether a pixel of the thinned image belongs to a branch point or an end point.
    If it is one of the two, then this pixel will be defined as a "node" with special meaning, and the resulting nodes are shown in Fig. \ref{neighborhood analysis}.
    Finally, we map the generated nodes back to the original map (as shown in Fig. {nodes on map}), and calculate other features besides coordinates to form the final augmented map.

    \subsubsection{Agent Intentions}

    In this section, we use a 2D Gaussian heatmap to visualize the agent’s intent.
    Using the means and variances of the A* path points of the agent's a few future steps, a 2D heat map as shown in Fig 12 is generated based on the Gaussian process.
    On this basis, we add a vector pointing from the agent's current position to the path point calculated by A* a few steps later, which is used to represent the agent's intentions in the short term.
    In summary, we visualize the agent's intent consisting of the agent's current location, locations it may visit in the short term, and the agent's orientation.

\end{document}